\documentclass[10pt,twocolumn,letterpaper]{article}

\usepackage[pagenumbers]{cvpr} % To force page numbers, e.g. for an arXiv version
\makeatletter
\@namedef{ver@everyshi.sty}{}
\makeatother

\usepackage{graphicx}
\usepackage{amsmath}
\usepackage{amssymb}
\usepackage{booktabs}

\usepackage{cite}
\usepackage{csvsimple}
\usepackage{amsmath}
\usepackage{booktabs}

\usepackage{pgfplots}
\usepackage{subcaption}
\usetikzlibrary{matrix}
\usepgfplotslibrary{groupplots}
\pgfplotsset{compat = newest}

\usepackage{array}
\usepackage{colortbl}
\usepackage{multirow}
\usepackage{xfp}
\usepackage{hhline}
\usepackage{siunitx}
\usepackage{pifont}
\usepackage[pagebackref,breaklinks,colorlinks]{hyperref}

\usepackage[capitalize]{cleveref}
\crefname{section}{Sec.}{Secs.}
\Crefname{section}{Section}{Sections}
\Crefname{table}{Table}{Tables}
\crefname{table}{Tab.}{Tabs.}

\usetikzlibrary{patterns}

\def\cvprPaperID{3} % *** Enter the CVPR Paper ID here
\def\confName{CVPR}
\def\confYear{2023}
\usepackage{layouts}

\begin{document}


\author{Timo Kaiser, Christoph Reinders, Bodo Rosenhahn\\
Institute for Information Processing (tnt)\\
L3S - Leibniz Universität Hannover, Germany\\
%Institut für Informationsverarbeitung and L3S Research Center\\% Leibniz University Hannover\\
%Hannover, Germany\\
{\tt\small \{kaiser, reinders, rosenhahn\}@tnt.uni-hannover.de}
% \and
% Christoph Reinders
}

\title{Compensation Learning in Semantic Segmentation}
\twocolumn[{%
\renewcommand\twocolumn[1][]{#1}%
\maketitle
\begin{center}
    \centering
    \captionsetup{type=figure}
    \includegraphics[width=1\textwidth]{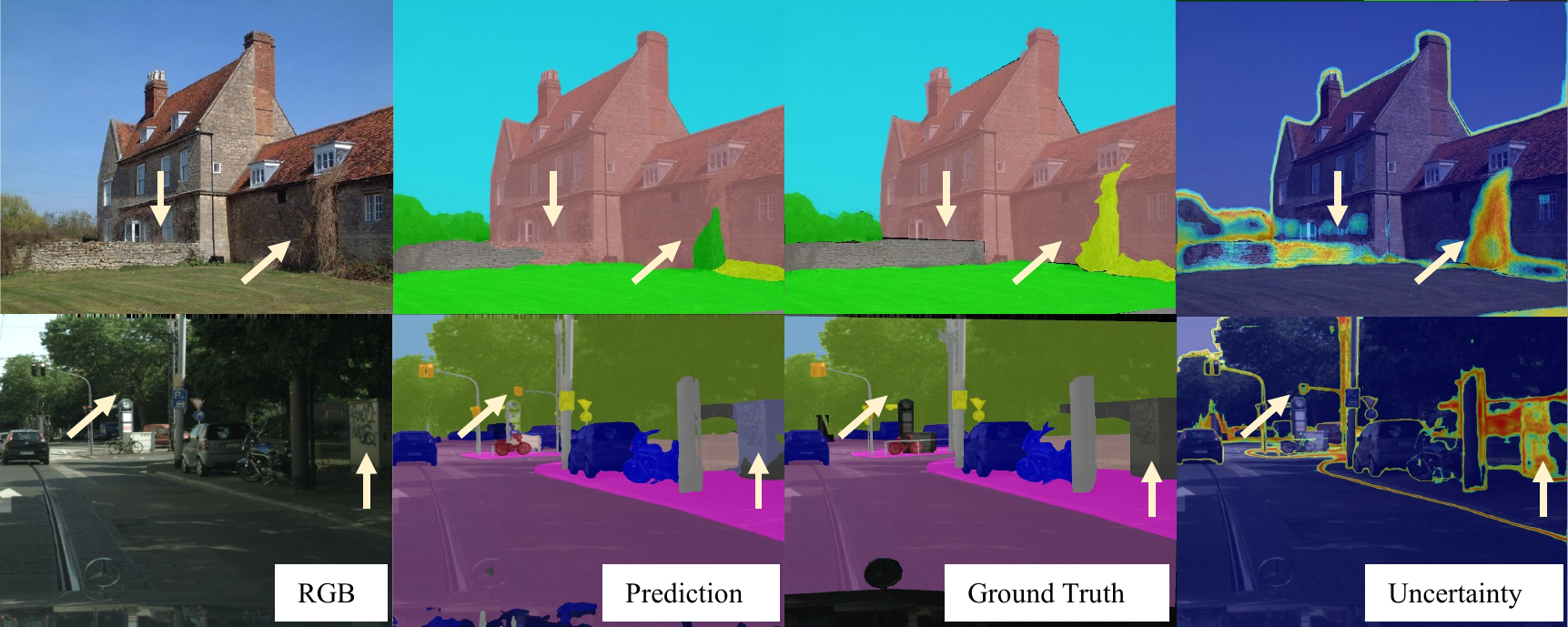}
    \vspace{-17pt}
    \captionof{figure}{Our local compensation provides uncertainty estimation and localizes wrong model predictions and ground truth label noise in ADE20k (top) and Cityscapes (bottom). Arrows highlight regions of interest with errors and colors indicate different classes.}
    \vspace{-5pt}
    \label{fig:teaser}
\end{center}%
}]

%%%%%%%%% ABSTRACT
% \begin{abstract}
%   The ABSTRACT is to be in fully justified italicized text, at the top of the left-hand column, below the author and affiliation information.
%   Use the word ``Abstract'' as the title, in 12-point Times, boldface type, centered relative to the column, initially capitalized.
%   The abstract is to be in 10-point, single-spaced type.
%   Leave two blank lines after the Abstract, then begin the main text.
%   Look at previous CVPR abstracts to get a feel for style and length.
% \end{abstract}
\begin{abstract}
Label noise and ambiguities between similar classes are challenging problems in developing new models and annotating new data for semantic segmentation.
In this paper, we propose Compensation Learning in Semantic Segmentation, a framework to identify and compensate ambiguities as well as label noise. More specifically, we add a ground truth depending and globally learned bias to the classification logits and introduce a novel uncertainty branch for neural networks to induce the compensation bias only to relevant regions. 
Our method is employed into state-of-the-art segmentation frameworks and several experiments demonstrate that our proposed compensation learns inter-class relations that allow global identification of challenging ambiguities as well as the exact localization of subsequent label noise. Additionally, it enlarges robustness against label noise during training and allows target-oriented manipulation during inference. We evaluate the proposed method on %the widely used datasets 
Cityscapes, KITTI-STEP, ADE20k, and COCO-stuff10k. 
% \textcolor{blue}{\textit{Code will be made publicly available.}}
\end{abstract}

%%%%%%%%% BODY TEXT

\newcommand{\todo}[1]{\iftrue\textcolor{blue}{#1}\fi}

\section{Introduction}
Semantic segmentation is a well-known and challenging task in computer vision~\cite{9356353, Kirillov_2019_CVPR, borse2021inverseform}. Thanks to the large investment of time and resources, the research community published a large number of elaborately curated datasets to train and evaluate methods for semantic segmentation~\cite{Weber2021NEURIPSDATA, Cordts2016Cityscapes, 10.1007/978-3-319-10602-1_48, Neuhold_2017_ICCV, 10.1007/978-3-319-46475-6_7,zhou2017scene}. Nevertheless, the industry needs an increasing amount of accurately annotated data and spends billion dollars to curate them~\cite{LabellingCosts}. Unfortunately, the annotation task stays challenging for humans even with advanced semi-automated annotation frameworks~\cite{10.1145/3240508.3241916, Castrejon_2017_CVPR, Voigtlaender19CVPR_MOTS}, because ambiguous image elements often can be assigned to multiple classes. Thus, annotated data is often noisy, with the consequence that the optimization of stochastic methods like neural networks is corrupted and the evaluation is distorted. Even the ground truth of widely used research benchmarks, which form the basis of this and many other papers, are subject to noise, as lamented by~\cite{Liu_2022_CVPR}. Semi-automated annotation without incorporating label noise is therefore a serious problem in semantic segmentation.

While tackling the impact of noisy labels is a well known research \mbox{topic~\cite{6685834, song2020learning,NEURIPS2020_e0ab531e,bressan2022semantic}}, avoiding noisy labels during labeling is shallow investigated. Because modern semi-automated annotation frameworks estimate an initial guess with a pre-trained segmentation framework~\cite{10.1145/3240508.3241916, Castrejon_2017_CVPR, Voigtlaender19CVPR_MOTS, Caelles_2017_CVPR}, an obvious way to improve the annotation framework is to improve the segmentation framework. To remove the residual error in the estimate, the human curator is still asked to inspect and correct the entire image. To reduce this effort, uncertainty estimation can help to guide the curator to find the most likely error regions. 
Current approaches like \textit{Bayesian Neural Networks}~\cite{mukhoti2018evaluating,Wang_2022_CVPRc}, that estimate and incorporate uncertainty in semantic segmentation aim to make the training more robust against label noise, but mainly detect boundaries between neighboring segments~\cite{bressan2022semantic,Wang_2022_CVPRc,Atigh_2022_CVPR}. 

Instead of using uncertainty estimation to make training more robust against noise, we aim to utilize robust training methods and uncertainty estimation to avoid new noise during data annotation.
Therefore, we present a novel method transferring compensation learning to semantic segmentation to compensate noise and ambiguities with end-to-end trainable compensation weights. Compensation learning, which adds ground truth depending bias to model predictions, has been introduced by Yao \textit{et al.}~\cite{yao2021compensation} for image classification. It allows the lowering of the influence of similar classes in order to reduce the impact of ambiguities and noise. We induce symmetry to make compensation learning stable during training and introduce an adaptive uncertainty branch that estimates the local importance of compensation.

Experiments on the widely used segmentation datasets Cityscapes~\cite{Cordts2016Cityscapes}, KITTI-STEP~\cite{Weber2021NEURIPSDATA}, ADE20k~\cite{zhou2017scene}, and COCO-stuff10k~\cite{10.1007/978-3-319-10602-1_48} show that our method 
learns interpretable inter-class compensations and is able to estimate prediction uncertainties. We present how compensation identifies challenging class pairs and the uncertainty localizes prediction errors very accurately.
Besides the interpretable guidance for data annotation, our method increases the robustness of training semantic segmentation methods with noisy labels and additionally introduces a useful method to improve the segmentation accuracy of certain classes. Moreover, we analyze and visualize inter-class ambiguities for the datasets.

\textbf{In summary}, our work contributes a novel framework\footnote{Code available at \href{https://github.com/tnt-LUH/compensation\_learning}{https://github.com/tnt-LUH/compensation\_learning}} to improve semi-automated annotation that
\begin{itemize}
    \itemsep0em 
    \item learns human-interpretable compensation weights of global inter-class ambiguities.
    \item introduces a novel uncertainty branch to adapt the compensation locally. The branch provides local guidance to image regions with high risk of errors. 
    \item improves robustness against noise during training.
    \item allows application-oriented manipulation of segmentation accuracy during inference.
\end{itemize}

\section{Related Work}
\label{sec:related_work}
Approaches to improve semi-automated annotation frameworks~\cite{10.1145/3240508.3241916, Castrejon_2017_CVPR, Voigtlaender19CVPR_MOTS, Caelles_2017_CVPR} are stronger \textit{Semantic Segmentation} methods to predict pseudo-labels, \textit{Robust Learning} algorithms to reduce memorization of noise during fine-tuning, and \textit{Uncertainty Estimation} as guidance to find prediction errors for human refinement. Related work for the mentioned topics and the state-of-the-art in \textit{Robust Semantic Segmentation} are presented in this section.  

\vspace{-12pt}
\paragraph{Semantic Segmentation.}
The predominant approach for semantic segmentation methods is using convolutional neural networks with encoder/decoder~\cite{Long_2015_CVPR} or feature pyramid~\cite{8100143} architectures. Extending the architecture with atrous convolutions~\cite{chen2017rethinking} improves the accuracy and ended with the introduction of \textit{DeepLabv3+}~\cite{deeplabv3plus2018}, which is widely used in science~\cite{Liu_2019_CVPR,9052469,cheng2020panoptic}. 
Latest state-of-the-art methods like
\textit{SegFormer}~\cite{NEURIPS2021_64f1f27b} apply transformers~\cite{pmlr-v80-parmar18a} to the architecture or use different approaches like Markov Random Fields~\cite{Xu_2020_ACCV}, binary space partitioning~\cite{GriOst2021a}, or class-agnostic clustering of associated pixels~\cite{NEURIPS2021_55a7cf9c}.

In general, the improvement of the above methods is accompanied by the introduction of improved backbones, such as~\cite{He_2016_CVPR,Howard_2019_ICCV,Liu_2021_ICCV,He_2022_CVPR}.

\vspace{-12pt}
\paragraph{Robust Learning.}
Common methods handling
label noise can be divided into label correction, loss correction methods, and meta-learning~\cite{Wang_2019_ICCV, SUN2022108467, song2020learning}. 

The goal of label correction is to identify and modify corrupted data annotations. Thereby, approaches vary from matching pseudo-labels to dynamic prototypes~\cite{Zhang_2021_CVPR}, estimating the non-affiliation to classes~\cite{kim2019nlnl}, or using bootstrapping, which maximizes the entropy between sample features and model predictions~\cite{reed2014training, pmlr-v97-arazo19a, NEURIPS2020_e0ab531e}. Bootstrapping is also used to approximate ensemble predictions~\cite{lu2022selc}. % by moving average to get more reliable soft-labels. 
Others directly optimize ground truth labels ~\cite{https://doi.org/10.48550/arxiv.2202.14026,Wang_2022_CVPRa,tanaka2018joint}. 

In loss correction approaches, the loss objective is adapted for each training sample. Whereas focal loss increases the impact of hard samples~\cite{Lin_2017_ICCV}, others down-weight uncertain samples. Weights are obtained for example via mutual agreement of model ensembles \cite{Wei_2020_CVPR} or \textit{peer-predictions}\cite{pmlr-v119-liu20e,zhu2021second} and the lowest-k weighted samples are rejected~\cite{NEURIPS2018_ad554d8c}. Instead of weighting, other methods adapt the loss objective, \eg, by combing loss functions~\cite{pmlr-v119-ma20c}, bounding losses~\cite{zhang2018generalized,ijcai2020-305}, or adding contrastive learning methods~\cite{Yi_2022_CVPR}.
Assuming statistically consistent noise, a transition matrix models the probability of label flips between certain classes~\cite{JMLR:v18:16-315}. A known transition matrix helps to determine the clean prediction and it can be integrated into neural networks\cite{Goldberger2017TrainingDN, sukhbaatar2014training, patrini2017making, NEURIPS2019_9308b0d6, yao2020dual}. Some approaches learn transitions in an end-to-end manner~\cite{sukhbaatar2014training, patrini2017making} or define it by human supervision~\cite{NEURIPS2018_aee92f16}. 
Instead of modelling probabilities, Yao \textit{et al.}~\cite{yao2021compensation} propose learning of ground truth depending bias. 
The influence of conditional noise can be compensated by adding bias to unconditional probability logits.

Meta-learning~\cite{vanschoren2019meta} with clear meta-data is used to predict additional information, \eg.,~the expected noise per training sample to weight the loss~\cite{shu2019meta} of prior predictions~\cite{SUN2022108467}. Meta-learning is also used to estimate the aforementioned transition matrix~\cite{Wang_2020_CVPR} or to augment new data by mixing meta- and noisy data~\cite{kaiser2022blind, Li2020DivideMix} or corrupting the meta-data~\cite{chimeramix}.  

\vspace{-12pt}
%\subsection{Uncertainty Estimation}
\paragraph{Uncertainty Estimation.}
Uncertainty Estimation of neural networks is mandatory to evaluate automated decisions such as the creation of pseudo-labels during annotation. According to~\cite{gawlikowski2021survey}, approaches can be divided into \textit{single deterministic methods} that predicts the uncertainty in one forward step~\cite{oberdiek2018classification,malinin2018predictive}, \textit{Bayesian methods} that utilize stochastic sampling~\cite{graves2011practical,barber1998ensemble}, \textit{ensembles} to evaluate multiple predictions~\cite{guo2018margin,valdenegro2019deep,lindqvist2020general}, and \textit{test-time augmentation}~\cite{lyzhov2020greedy}.  

\vspace{-12pt}
\paragraph{Robust Semantic Segmentation.}
Many of the aforementioned methods are not applicable in semantic segmentation, either conceptually or in terms of complexity, or are applicable but not investigated further.
Current state-of-the-art methods for robust learning in semantic segmentation detect label noise in an iterative process on the training set. Liu \textit{et al.}~\cite{Liu_2022_CVPRa} detect the memorization effect for every pixel and correct them with multi-scaled predictions. Since the model needs to be retrained from scratch after noise detection, this method cannot be reasonable applied in online semi-automated annotation.
Wang \textit{et al.}~\cite{Wang_2022_CVPR} propose a semi-supervised framework using contrastive predictive coding loss~\cite{oord2018representation}, but it does not identify label noise.
Related to robust learning, uncertainty is incorporated in semantic segmentation methods. Atigh \textit{et al.}~\cite{Atigh_2022_CVPR} provide an uncertainty estimation by embedding semantic segmentation into hyperbolic space. Others estimate uncertainty with \textit{Bayesian Neural Networks}~\cite{mukhoti2018evaluating,Wang_2022_CVPRc}, model ensembles~\cite{Holder_2021_ICCV}, or explicitly define uncertainty at region borders~\cite{bressan2022semantic} or for entire segments via aggregated dispersion measures~\cite{9206659}. Although the latter go in the right direction, they do not explicitly consider ambiguities like our method.

\begin{figure*}[ht]
    \centering
    \includegraphics[width=1\textwidth]{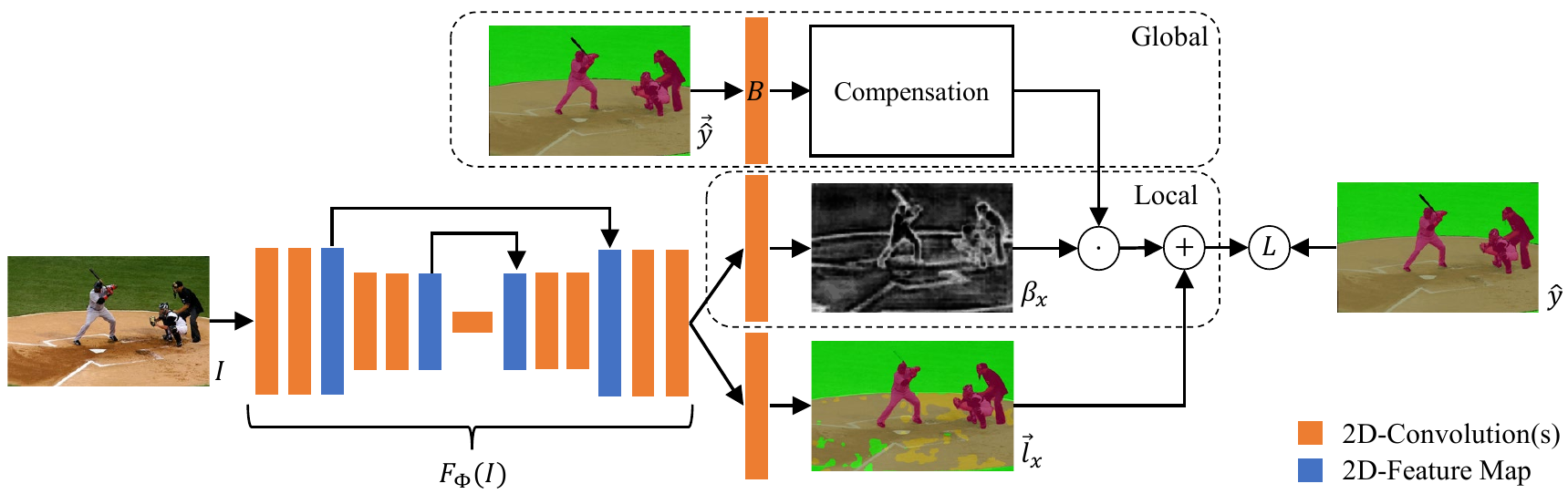}
    \vspace{-18pt}
    \caption{The global compensation $B$ and local uncertainty $\beta_x$ detects noisy image regions and makes convolutional segmentation frameworks $F_\Phi(I)$ robust against noise and ambiguities. It adds a ground truth $\hat{y}$ depending compensation to logits $\vec{l}_x$ to optimize $L$.}
    \label{fig:framework}
    \vspace{-15pt}
\end{figure*}

\section{Method}
\label{sec:method}
In this section, we present our proposed method of compensation learning in semantic segmentation that introduces global and local guidance for human label correction into existing segmentation frameworks.
Furthermore, we introduce symmetry constraints that improve training and show how compensation can be used to manipulate inference of segmentation networks. The overall framework is presented in Fig.~\ref{fig:framework}. 

\subsection{Preliminaries}
The goal of semi-automated annotation tools is to accurately solve the semantic segmentation task with the least human curation effort. 
Semantic segmentation is a multi-class classification problem, in which each pixel~$x$ of an image $I$ should be assigned to the true class \mbox{label $y\in C$} from a set of classes $C=\{1,\ldots, K \}$. Modern annotation tools estimate an initial guess of the unknown label, which is then manually inspected and corrected to $y$ by a human curator. The initial guess is nowadays estimated by neural networks $F_\Phi$ with optimized weights $\Phi$ (\eg, by \cite{cheng2020panoptic, deeplabv3plus2018, mohan2021efficientps, NEURIPS2021_64f1f27b}). During estimation, the \mbox{probability $P(Y=i|x, \Phi)$} that represents the likelihood of pixel $x$ belonging to class $i$ is estimated for every $i\in C$. 
First, the classifier $F_\Phi$ predicts independent logits $\vec{l}_x=\{l_{x,1}, \ldots, l_{x,K} \}$, which are then transformed into conditional probabilities using the softmax function $\mathbb{S}$~\cite{NIPS1989_0336dcba} 
\begin{equation}
\label{equ:softmax_wo_transition}
    P(Y=i|x, \Phi) = \mathbb{S}( \vec{l}_x )_{i} = \frac{e^{l_{x,i}}}{\sum_{n=1}^K e^{l_{x,n}}},    \qquad
    i \in C.
\end{equation}
Finally, the pixel~$x$ gets assigned to class~$i$ with the highest probability.

Suitable weights $\Phi$ need to be obtained during a preceding optimization process with already annotated image data. The general approach is to minimize the \textit{cross-entropy} loss
\begin{equation}
\label{equ:loss}
    L_{\text{CE}}= -\frac{1}{|I|} \sum_{\substack{(x,\hat{y})\\x\in I}} \log\big( P(Y=\hat{y}|x, \Phi) \big), \quad \hat{y} \in C,
\end{equation}
in which $\hat{y}$ denotes a given ground truth label for pixel $x$. Optionally, $\Phi$ can be fine-tuned on new annotated image data to decrease the domain gap and increase the segmentation accuracy for new estimations~\cite{Weber2021NEURIPSDATA}.

\subsection{Global Compensation Learning}
\label{sec:global_compensation}
Unfortunately, optimizing the \textit{cross-entropy} loss is prone to overfit on noisy or ambiguous pixels\cite{wei2022learning,zhang2018generalized}. The segmentation accuracy of the classifier $F_\Phi$ degrades during pre-training or induces confirmation bias during optional fine-tuning~\cite{confirmation_bias}. For noisy labels, the ground truth label $\hat{y}$ differs from the true label $y$. Most label flips $\hat{y}\neq y$ are based on ambiguities between $\hat{y}$ and $y$. For example,  a curator might simply recognize a \textit{bus} as a \textit{truck} when it is in the distance. Thus, ambiguous visual appearance and label flips should be seen as conditional dependent.

An ambiguous pixel $x$ that could equally be assigned to classes $i$ or $j$ usually has only one ground truth label, \eg,~$\hat{y}=j$. According to the visual similarity, a well trained classifier will generate similar logits~$l_{x,i}\approx l_{x,\hat{y}}$, which cause approximately equal \mbox{probabilities~$P(Y=i|x,\Phi)\approx P(Y=\hat{y}|x,\Phi)$.} This leads to a large loss in~$L_\text{CE}$, even if the probability for the ground truth label~$P(Y=\hat{y}|x,\Phi)$ has the highest probability and solves the classification task correctly. For the aforementioned example, the logit $l_{x,i}$ should therefore not strongly influence the probability~$P(Y=\hat{y}|x)$. Technically, this intuition can be implemented by decreasing~$l_{x,i}$ for pixels with the label~$\hat{y}=j$. 

To address ambiguities and label noise, we adapt compensation learning~\cite{yao2021compensation} and add a trainable conditional bias to the logits $\vec{l}_x$ during training. The softmax formulation from Eq.~\eqref{equ:softmax_wo_transition} is extended to
\begin{equation}
\label{equ:softmax_w_compensation}
    P(Y=i|x,  \hat{y}, \Phi, B) = \frac{e^{l_{x,i} +B_{i\hat{y}}}}{\sum_{n=1}^K e^{l_{x,n}+B_{n\hat{y}}}},    \qquad
    i,\hat{y} \in C
\end{equation}
where $B\in \mathbb{R}^{K\times K}$ denotes a compensation matrix. Each element $B_{i\hat{y}}$ enables decreasing or increasing the impact of logit $l_{x,i}$ for pixels with the ground truth label $\hat{y}$. 
Instead of manually defining $B$ and to address two-dimensional images, we integrate $B$ into the neural segmentation framework $F_\Phi$ with a single two-dimensional convolutional layer and optimize it alongside $\Phi$.
During training, compensation can reduce overfitting by minimizing Eq.~\eqref{equ:loss} for ambiguous pixel regions and provide insights about inter-class correlations.
The following sections describe how we improve compensation learning to boost semantic segmentation,  and how we deduce a novel uncertainty estimation and inference approach.   

\subsection{Local Compensation Learning}
\label{sec:local_compensation_learning}
The compensation matrix manipulates optimization globally so that an element $B_{i\hat{y}}$ affects the \mbox{probabilities $P(Y=i|x,\hat{y},\Phi)$} for all pixels with ground truth label $\hat{y}$. This is not reasonable for unambiguous pixels because it lowers the impact of good training samples in the optimization process. Thus, we introduce a novel uncertainty branch to estimate a local importance factor $\beta_x\in[0,1]$ for each pixel $x$, and change the global concept of compensation Eq.~\eqref{equ:softmax_w_compensation} to
\begin{equation}
\label{equ:softmax_w_compensation_local}
    P(Y=i|x,  \hat{y}, \Phi, B) = \frac{e^{l_{x,i} +\beta_x B_{i\hat{y}}}}{\sum_{n=1}^K e^{l_{x,n}+\beta_x B_{n\hat{y}}}},    \quad
    i,\hat{y} \in C.
\end{equation}
To predict the local importance $\beta_x$ from high-level features, a lightweight regression head is added parallel to the classification head of the base segmentation framework. We employ two pointwise convolutional layers with 64 and 1 output channels, followed by a batchnorm and a sigmoid activation, respectively. The convolutional weights are added to $\Phi$ to be trained alongside the original segmentation framework. Our novel framework to observe conditional probabilities $\vec{p}_x$ for pixel $x$ can be expressed with the corresponding logits $\vec{l}_x$, the softmax $\mathbb{S}$, and the one-hot ground truth vector $\vec{\hat{y}}$ as 
\begin{equation}
\label{equ:compensation_integration}
    \forall x \in I: \quad \vec{p}_x= \mathbb{S}\big( \vec{l}_x+\beta_x \cdot B \vec{\hat{y}} \big), \qquad
    \hat{y} \in C.
\end{equation}
The general architecture is visualized in Fig.~\ref{fig:framework}.

As ambiguities and label errors accompany, we propose compensation as local guidance to detect prediction errors during the annotation process. Inspired by the cost intensive uncertainty estimation of \textit{Bayesian Neural Networks}~\cite{mukhoti2018evaluating}, we introduce a lightweight approach to determine the uncertainty only with the uncompensated logits $\vec{l}_x$ and the global and local compensation $B$ and $\beta_x$. Since the true annotation $y=\hat{y}$ is most likely in the top-$k$ uncompensated predictions, we calculate the local variance for pixel $x$ with
\begin{equation}
    \sigma^2_x = \frac{1}{k}\sum_{c\in C^k} \big( \underbrace{P(Y=o|x,\hat{y}=c, \Phi)}_{\text{Eq. \eqref{equ:softmax_w_compensation_local}}} - \underbrace{P(Y=o|x,\Phi)}_{\text{Eq. \eqref{equ:softmax_wo_transition}}} \big)^2, % \in [0, 1].
\end{equation}
where $C^k$ denotes the subset of the top-$k$ classes and $o$ is the class with the highest uncompensated probability. % $o=\argmax_{o\in C} P(Y=o|x,\Phi)$. 
The variance approaches zero, if $\beta_x$ is small or the top-$k$ classes do not compensate each other. In contrast, the variance increases for compensating classes in $C^k$, which indicates underlying ambiguities.
To refine $\sigma^2_x$, we incorporate the model uncertainty $u_x$ and estimate the likelihood of prediction errors $e_x$ of pixel $x$ with
\begin{equation}
\label{equ:uncertainty}
    \forall x\in I: e_x = \bigg[\sigma^2_x\cdot \underbrace{\big(1 - \max_{i\in C} P(Y=i|x, \Phi)\big)}_{u_x}\bigg]^\varphi \in [0, 1]
\end{equation}
and add an exponent $\varphi$ that allows to amplify or diminish the uncertainty for visualization purposes. In our experiments, $\varphi$ is adapted manually for visualization purposes only and we set $k$ to $5$ (see supplementary material, Sec.~D).

\subsection{Constrained and Penalized Compensation}
The described compensation matrix $B$ entails two risks: 
Adding a compensation matrix allows a mode collapse and can amplify the negative impact of imbalances in the training data. 

Setting $\beta_x\approx 1$, $B_{i,i}\gg 0$ and $B_{i,j\neq i}\ll 0$ minimizes the main objective Eq.~\eqref{equ:loss} without need of reasonable $\Phi$. To prevent the mode collapse, we penalize $B$ with local lasso regression~\cite{10.2307/2346178} and extend the loss from Eq.~\eqref{equ:loss} to  
\begin{equation}
\label{equ:loss_full}
    L = -\frac{1}{|I|} \sum_{\substack{(x,\hat{y})\\x\in I}} \log\bigg(P(Y=\hat{y}|x, \Phi, B)\bigg)+ \frac{\alpha}{K}\sum_i^C \beta_x|B_{i\hat{y}}|
\end{equation}
and weight the loss by $\alpha$ to adjust the penalty of $B$.
We also constrain the diagonal entries of $B$ to be zero: 
\begin{equation}
\label{equ:constraint1}
    \forall i \in C :\quad     B_{ii} = 0 \ ,
\end{equation}
as class $i$ cannot have reasonable correlations to itself.

To enlarge the robustness against imbalances like the proportion of road pixels and sidewalk pixels, we add an optional symmetry constraint 
\begin{equation}
\label{equ:constraint2}
    \forall i,j \in C :\quad     B_{ij} = B_{ji}
\end{equation}
to stabilize the training. 
As drawback, this symmetry suppresses potential insights about the distribution of label errors from the global perspective as described in Sec.~\ref{sec:global_compensation}.   

\subsection{Compensated Inference}
\label{sec:compensated_inference}
The described compensation framework relies on the ground truth label $\hat{y}$ and needs to be modified for the inference of unseen images for the semi-automated annotation task. A simple option is to remove $B$ by setting Eq.~\eqref{equ:softmax_w_compensation_local} back to Eq.~\eqref{equ:softmax_wo_transition}. 
For applications with prioritized classes, we instead propose to relax and estimate the ground truth one-hot vector $\vec{\hat{y}}$ (see Eq.~\eqref{equ:compensation_integration}) with the uncompensated prediction of Eq.~\eqref{equ:softmax_wo_transition}. While incorporating the model prediction, the usage of learned compensation $B$ is not reasonable because it could induce confirmation bias for wrong predictions. % from Eq.~\eqref{equ:softmax_wo_transition}.
Instead, the compensation matrix $B$ can be defined manually to induce intended bias. Application-oriented compensations can prioritize important or vulnerable classes during prediction of the initial guess in the annotation process. To prioritize a class $i$ in general, $B_{ii}$ needs to be increased to a large positive value. To prioritize $i$ against a specific class $j$, $B_{ij}$ needs to be decreased to a large negative value, respectively. Our compensated inference can be applied a posteriori without the need of extra training and is therefore a non-bayesian alternative to~\cite{Chan_2019_CVPR_Workshops}. 

\label{sec:interpretable_compensation}

\begin{table*}[t]
    \centering
    \newcommand{\pst}[1]{{\cellcolor{red!\fpeval{100*(#1+0.18)/3.0}}}\scalebox{0.75}{#1}}
    \newcommand{\ngt}[1]{{\cellcolor{blue!\fpeval{100*(#1+0.18)/3.0}}}\scalebox{0.75}{-#1}}
    \newcommand{\zero}[0]{\scalebox{0.6}{0}}
    \newcolumntype{"}{@{\hskip\tabcolsep\vrule width 1pt \hspace{0.1pt}|}}
    \newcolumntype{x}[0]{>{\centering\arraybackslash\hspace{0pt}}p{0.024\textwidth}}
    \caption{The learned compensation values $B_{ij}$ for 11 classes in KITTI-STEP and Cityscapes provided by our method. A negative value~$B_{ij}$ lowers the impact of class $i$ for pixels with the ground truth annotation $j$. Note that values are rounded after the first digit.} 
    \vspace{-7pt}
    \resizebox{\textwidth}{!}{
    \begin{tabular}{c|x|x|x|x|x|x|x|x|x|x|x||x|x|x|x|x|x|x|x|x|x|x}
    \toprule
    
\multicolumn{1}{c|}{\multirow{5}{*}{$B_{ij}$}} & \multicolumn{11}{c||}{KITTI-STEP} & \multicolumn{11}{c}{Cityscapes} \\ \cline{2-23}
    
& \rotatebox{90}{\textit{road}} 
& \rotatebox{90}{\textit{sidewalk}} 
& \rotatebox{90}{\textit{building}} 
& \rotatebox{90}{\textit{wall}} 
& \rotatebox{90}{\textit{fence}} 
& \rotatebox{90}{\textit{pole}} 
& \rotatebox{90}{\textit{tr. light}}
& \rotatebox{90}{\textit{tr. sign}}
& \rotatebox{90}{\textit{vegetation}} 
& \rotatebox{90}{\textit{terrain}} 
& \rotatebox{90}{\textit{sky}} 
& \rotatebox{90}{\textit{road}} 
& \rotatebox{90}{\textit{sidewalk}} 
& \rotatebox{90}{\textit{building}} 
& \rotatebox{90}{\textit{wall}} 
& \rotatebox{90}{\textit{fence}} 
& \rotatebox{90}{\textit{pole}}
& \rotatebox{90}{\textit{tr. light}}
& \rotatebox{90}{\textit{tr. sign}}
& \rotatebox{90}{\textit{vegetation\ }}  
& \rotatebox{90}{\textit{terrain}} 
& \rotatebox{90}{\textit{sky}} \\\midrule % \hline\hline

\textit{road} &\zero &\ngt{1.7} &\zero &\zero &\ngt{0.2} &\ngt{0.2} &\zero &\ngt{0.1} &\ngt{0.3} & \ngt{1.1} &\zero                
&\zero &\ngt{2.1} &\ngt{0.1} &\ngt{0.1} &\ngt{0.1} &\ngt{0.1} &\zero &\zero &\ngt{0.2} &\ngt{0.6} &\zero\\\cline{1-1} 

\textit{sidewalk} &\ngt{1.6} &\zero &\ngt{0.4} &\ngt{0.1} &\ngt{0.1} &\ngt{0.2} &\zero &\zero &\ngt{0.4} &\ngt{0.8} & \zero                 
&\ngt{2.0} &\zero &\ngt{0.8} &\ngt{0.3} &\ngt{0.3} &\ngt{0.5} &\zero &\zero &\ngt{0.4} &\ngt{1.0} &\zero\\\cline{1-1} 

\textit{building} &\zero &\ngt{0.4} &\zero &\ngt{0.1} &\ngt{0.3} &\ngt{0.8} &\ngt{0.1} &\ngt{0.2} &\ngt{1.7} & \ngt{0.1} &\ngt{0.5}                
&\ngt{0.1} &\ngt{0.8} &\zero &\ngt{0.9} &\ngt{0.9} &\ngt{1.9} &\ngt{0.4} &\ngt{0.8} &\ngt{2.5} &\ngt{0.2} &\ngt{0.8}\\\cline{1-1} 

\textit{wall} &\zero &\zero &\ngt{0.1} &\zero &\ngt{0.2} &\zero &\zero &\zero &\ngt{0.1} &\zero & \zero               
&\zero &\ngt{0.3} &\ngt{0.8} &\zero &\ngt{0.6} &\ngt{0.1} &\zero &\zero &\ngt{0.5} &\ngt{0.1} &\zero\\\cline{1-1}

\textit{fence} &\ngt{0.1} &\ngt{0.1} &\ngt{0.2} &\ngt{0.2} &\zero &\ngt{0.1} &\zero &\zero &\ngt{0.7} & \ngt{0.3} &\zero                
&\zero &\ngt{0.2} &\ngt{0.7} &\ngt{0.6} &\zero &\ngt{0.3} &\zero &\zero &\ngt{0.6} &\ngt{0.2} &\zero\\\cline{1-1}

\textit{pole} &\ngt{0.1} &\ngt{0.2} &\ngt{0.8} &\zero &\ngt{0.1} &\zero &\ngt{0.1} &\ngt{0.2} &\ngt{1.1} & \ngt{0.2} &\ngt{0.3}                
&\zero &\ngt{0.4} &\ngt{1.7} &\ngt{0.1} &\ngt{0.3} &\zero &\ngt{0.2} &\ngt{0.3} &\ngt{1.3} &\ngt{0.2} &\ngt{0.2}\\\cline{1-1}

\textit{tr. light} &\zero &\zero &\zero &\zero &\zero &\ngt{0.1} &\zero &\zero &\ngt{0.1} & \zero &\zero                
&\zero &\zero &\ngt{0.3} &\zero &\zero &\ngt{0.2} &\zero &\zero &\ngt{0.3} &\zero &\zero\\\cline{1-1}

\textit{tr. sign} &\zero &\zero &\ngt{0.2} &\zero &\zero &\ngt{0.2} &\zero &\zero &\ngt{0.4} & \zero &\zero                
&\zero &\zero &\ngt{0.7} &\zero &\zero &\ngt{0.3} &\ngt{0.1} &\zero &\ngt{0.3} &\zero &\zero\\\cline{1-1}

\textit{vegetation} &\ngt{0.3} &\ngt{0.5} &\ngt{1.8} &\ngt{0.2} &\ngt{0.8} &\ngt{1.2} &\ngt{0.2} &\ngt{0.5} &\zero & \ngt{1.3} &\ngt{1.2}                
&\ngt{0.1} &\ngt{0.5} &\ngt{2.4} &\ngt{0.6} &\ngt{0.7} &\ngt{1.4} &\ngt{0.3} &\ngt{0.4} &\zero &\ngt{1.1} &\ngt{0.9}\\\cline{1-1}    

\textit{terrain} &\ngt{1.0} &\ngt{0.7} &\zero &\zero &\ngt{0.3} &\ngt{0.2} &\zero &\zero &\ngt{1.2} & \zero &\zero                
&\ngt{0.4} &\ngt{1.0} &\ngt{0.1} &\ngt{0.1} &\ngt{0.2} &\ngt{0.2} &\zero &\zero &\ngt{1.0} &\zero &\zero\\\cline{1-1}

\textit{sky} &\zero &\zero &\ngt{0.4} &\zero &\zero &\ngt{0.3} &\zero &\ngt{0.1} &\ngt{1.1} & \zero &\zero                
&\zero &\zero &\ngt{0.7} &\zero &\zero &\ngt{0.2} &\zero &\zero &\ngt{0.8} &\zero &\zero\\\midrule

\end{tabular}}
\vspace{-10pt}
\label{tab:qualitative_results}
\end{table*}

\begin{figure*}[t]
    \centering
    \includegraphics[width=1\textwidth]{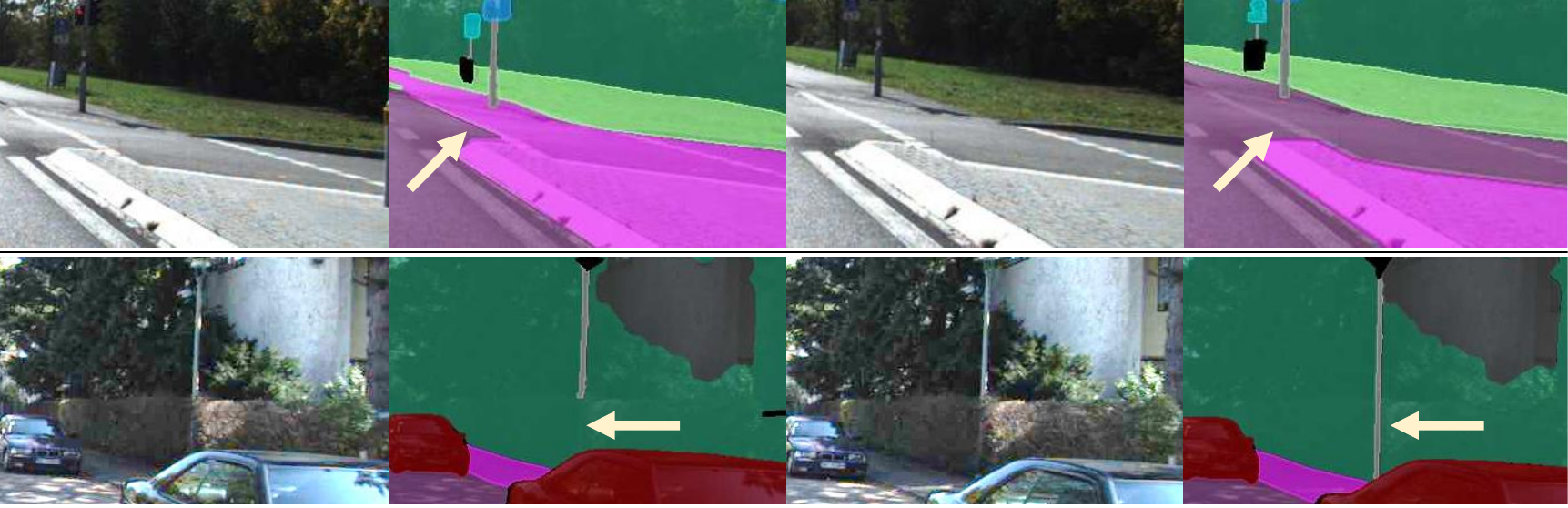}
    \vspace{-20pt}
    \caption{
    Most challenging class pairs indicated by learned compensation weights. The first and third column show subsequent images from KITTI-STEP video sequences and their ground truth. The upper row shows a flip between \textit{road} and \textit{sidewalk}, and the lower between \textit{pole} and \textit{vegetation}. Both ambiguous pairs are challenging and cause systematical errors.}
    \label{fig:samples}
    \vspace{-10pt}
\end{figure*}

\section{Experiments}
\label{sec:experiments}
In this section, we present several experiments to evaluate the proposed method. First, the experimental setup and metrics used for evaluation are introduced. Then, we study how our method can be used to identify challenging inter-class ambiguities globally and prediction errors locally. Also the impact of our method on robustness against conditional label noise is evaluated and experiments are presented, which demonstrate the application-orientated usage of compensated inference. 

\subsection{Experimental Setup}
We evaluate our method on the four publicly available semantic segmentation datasets Cityscapes \cite{Cordts2016Cityscapes}, KITTI-STEP~\cite{Weber2021NEURIPSDATA}, ADE20k~\cite{zhou2017scene}, and COCO-stuff10k~\cite{10.1007/978-3-319-10602-1_48}. These widely used datasets establish benchmarks for state-of-the-art segmentation methods with small and large amount of classes. Furthermore, the datasets Cityscapes and KITTI-STEP enable comparability for the interpretation task of ambiguities, because they share the equal set of classes $C$. Both datasets provide segmentation data in an autonomous driving setting, where the images are annotated per-pixel with 19 semantic classes, whereas ADE20k is annotated with 151 and COCO with 170 classes. We evaluate on the validation sets to allow extensive investigations.

To analyze the impact of compensation, we integrate our method into the well-known semantic segmentation framework \textit{DeepLabv3+}~\cite{deeplabv3plus2018} and the state-of-the-art framework \textit{SegFormer}~\cite{NEURIPS2021_64f1f27b}. 
The loss balancing $\alpha$ is set to $0.01$ and $1$, the learning rate to $0.01$ and $0.00006$, and we train for \num{80000} and \num{160000} epochs, respectively. 

Ours and reported reference methods are employed on top of the baseline frameworks and implemented in the widely used framework \textit{MMSegmentation}\cite{mmseg2020} to improve the reproducability. More details on the experimental setup and implementations of later mentioned reference methods can be found in the supplementary material, Sec.~A.

To evaluate our method, we use the widely-used {mIoU} metric~\cite{thoma2016survey} that evaluates the assignment of class labels and balances underrepresented classes. We also use the class accuracy ($\text{Acc}_c$) and aggregated accuracy ($\text{Acc}_a$)  
to verify segmentation results on pixel level. Mathematical details of the metrics are elaborated in the supplementary material, Sec.~B.  

% % \subsection{Metrics}
% We evaluate with the widely-used {mIoU} metric~\cite{thoma2016survey}, which is also referenced as \textit{Semantic Quality}~\cite{Long_2015_CVPR, Weber2021NEURIPSDATA}. With the set of pixels $X_c$ that are assigned to class $c$ and ground truth labels $\hat{Y}_c$ annotated as $c$, the mIoU metric is defined as 
% \begin{equation}
% \label{equ:miou}
% \text{mIoU}=\frac{1}{|C|}\sum_{c\in C}\frac{|X_c\cap \hat{Y}_c|}{|X_c\cup \hat{Y}_c|}\ .
% \end{equation}
% The {mIoU} metric evaluates the assignment of class labels and balances underrepresented classes. We also use the class accuracy ($\text{Acc}_c$) and aggregated accuracy ($\text{Acc}_a$)  
% \begin{equation}
% \label{equ:acc_i}
% \text{Acc}_c=\frac{|X_c\cap \hat{Y}_c|}{|\hat{Y}_c|}\quad \text{and} \quad \text{Acc}_a=\frac{\sum_{c\in C}|X_c\cap \hat{Y}_c|}{\sum_{c\in C}|\hat{Y}_c|}
% \end{equation}
% to verify segmentation results on pixel level. 

\subsection{Global Compensation: Interpretable Data}
\label{sec:experiments_global}
In this section, we analyze the learned compensation weights $B$ after training with Eq.~\eqref{equ:compensation_integration} but without symmetry constraint (Eq.~\eqref{equ:constraint2}). Tab.~\ref{tab:qualitative_results} shows a sub-selection of $B$ trained on the low class datasets KITTI-STEP and Cityscapes with \textit{DeepLabv3+}. 
In direct comparison, both datasets share the strongest compensated class pairs, \eg,~\textit{road}-\textit{sidewalk} or \textit{pole}-\textit{building}.
The outstanding class pairs $ij$ with large compensation weights~$B_{ij}$ can be considered more difficult for the curator and prone to label noise.  
Our method identifies intuitive, like \textit{road}-\textit{sidewalk}, and also not intuitive ambiguities, such as \textit{pole}-\textit{building} and \textit{pole}-\textit{vegetation}. These ambiguities can be verified by multiple samples of label flips in the data. 
Examples for the most ambiguous class pairs are shown in Fig.~\ref{fig:samples}.
Compared to the distribution dependent confusion matrix, the compensation matrix indicates ambiguities independently based on their impact during optimization.  
The full compensation matrices for \textit{SegFormer} and \textit{DeepLabv3} and confusion matrices can be found in the supplementary material, Sec.~G and~H. We noticed, that compensation stronger influences the transformer based \textit{SegFormer}, because values in $B$ are much larger compared to \textit{DeepLabv3+}. 

In summary, the proposed method learns and provides human-interpretable insights about inter-class ambiguities in the model optimization. 
With this information, human curators are able to improve dataset quality by focusing on systematical errors caused by ambiguities. 

\subsection{Local Compensation: Label Noise Detection}
\label{sec:local_compensation}
\begin{figure*}[t]
\centering
\begin{tikzpicture}
    \begin{groupplot}[
        group style={group size= 4 by 1},
        height=0.22\textwidth,
        width=0.27\textwidth,
        xmin = 0, xmax = 0.5,
        xtick distance = 0.25,
        minor tick num = 1,
        grid = both,
        xlabel = {$r_\text{Area}$},
        major grid style = {lightgray},
        minor grid style = {lightgray!25},
        scaled ticks=false, 
        %log ticks with fixed point,
        tick label style={/pgf/number format/fixed},
    ]
        \nextgroupplot[
                    legend style={at={(0.5,-0.05)}, anchor=north},
                    legend to name=zelda,
                    legend columns=7,
                    title=Cityscapes,
                    xmax = 0.5,
                    xtick distance = 0.25,
                    ymin = 0.96, ymax = 1.0,
                    ytick distance = 0.04,
                    ylabel = {$\text{Acc}_a$},
                    domain = 0:1,
                ] 
                \addplot[
                    thick,
                    red,
                ] file[skip first] {detection_city_val_Combined.txt};\addlegendentry{Ours ($e_x$)};
                \addplot[
                    black,thick,
                ] file[skip first] {detection_city_val_Alpha.txt};\addlegendentry{$\beta_x$};
                \addplot[
                    blue,thick,
                ] file[skip first] {detection_city_val_Bayes.txt};\addlegendentry{BNN};
                \addplot[
                    green,
                    thick,
                ]{0.0+x};\addlegendentry{HIS};
                \addplot[
                    dashed,
                    black,
                ]{0.964+x};\addlegendentry{Oracle};
                \coordinate (top) at (rel axis cs:0,1);
        \nextgroupplot[
                    title=KITTI,
                    ymin = 0.9, ymax = 1.,
                    xmax = 0.5,
                    ytick distance = 0.05,
                ] 
                \addplot[
                    thick,
                    red,
                ] file[skip first] {detection_kitti_val_Combined.txt};
                \addplot[
                    black,thick,
                ] file[skip first] {detection_kitti_val_Alpha.txt};
                \addplot[
                    blue,thick,
                ] file[skip first] {detection_kitti_val_Bayes.txt};
                \addplot[
                    dashed,
                    black,
                ]{0.909+x};
        \nextgroupplot[
                    title=COCO,
                    ymin = 0.6, ymax = 1.,
                    ytick distance = 0.2,
                ] 
                \addplot[
                    thick,
                    red,
                ] file[skip first] {detection_coco10k_val_Combined.txt};
                \addplot[
                    black,thick,
                ] file[skip first] {detection_coco10k_val_Alpha.txt};
                \addplot[
                    green,
                    thick,
                ] file[skip first] {detection_coco10k_val_hyperbolic.txt};
                \addplot[
                    blue,thick,
                ] file[skip first] {detection_coco10k_val_Bayes.txt};
                \addplot[
                    dashed,
                    black,
                ]{0.663+x};
        \nextgroupplot[
                    title=ADE20k,
                    ymin = 0.8, ymax = 1.,
                    ytick distance = 0.1,
                ] 
                \addplot[
                    thick,
                    red,
                ] file[skip first] {detection_ade20k_val_Combined.txt};
                \addplot[
                    black,thick,
                ] file[skip first] {detection_ade20k_val_Alpha.txt};
                \addplot[
                    green,
                    thick,
                ] file[skip first] {detection_ade20k_val_hyperbolic.txt};
                \addplot[
                    blue,thick,
                ] file[skip first] {detection_ade20k_val_Bayes.txt};
                \addplot[
                    dashed,
                    black,
                ]{0.805+x};
                \coordinate (bot) at (rel axis cs:1,0);
    \end{groupplot}
    \path (top)--(bot) coordinate[midway] (group center);
    \node[right=1em,inner sep=0pt] at(3,-1.25) {\pgfplotslegendfromname{zelda}};
\end{tikzpicture}
\vspace{-5pt}
\caption{Analysis of error detection guided by our method used to correct prediction errors of ~\textit{DeepLabv3+}. The ratio of top-$k$ uncertain pixels $r_\text{Area}$ that is corrected by the human curator improves the segmentation accuracy $\text{Acc}_a$. We report the accuracy after correction with our guidance $e_x$, the local compensation weight $\beta_x$, BNN~\cite{mukhoti2018evaluating}, and HIS~\cite{Atigh_2022_CVPR} (if available). Oracle mimes the optimal uncertainty estimation.} 
\label{fig:noise_group}
\vspace{-10pt}
\end{figure*}
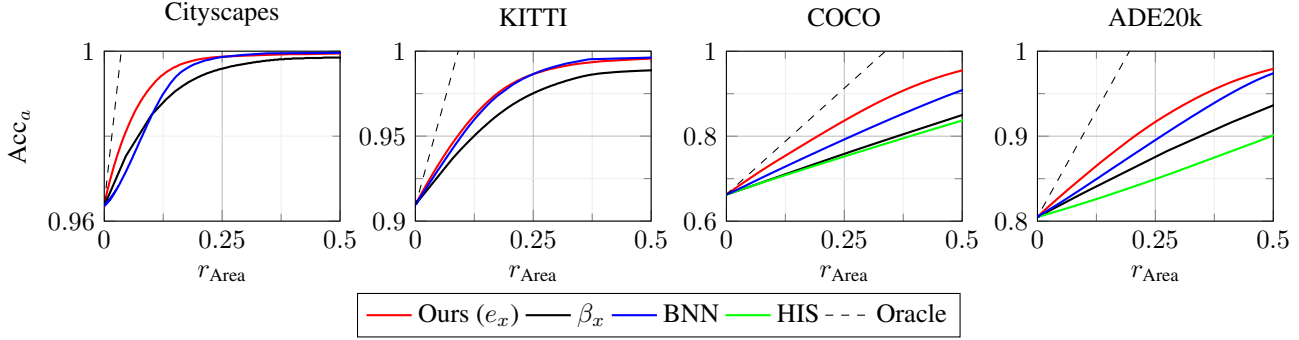
\begin{table}
    \centering
    \caption{Area-under-Curve for the noise detection experiments in Fig.~\ref{fig:noise_group} of our method compared to BNN~\cite{mukhoti2018evaluating} and HIS~\cite{Atigh_2022_CVPR}. Note that HIS evaluation is only available for the large-scale datasets. }
    \vspace{-5pt}
    \resizebox{\columnwidth}{!}{%
    \begin{tabular}{c c | c c c c }
    \toprule
    \multicolumn{2}{c}{AUC$\uparrow$}  & ~Cityscapes & KITTI~ & ~COCO & ADE20K~ %& Mean
    \\\midrule
    \multirow{5}{*}{\rotatebox{90}{{Method~~}}}&{Oracle} & 0.999 & 0.996 & 0.943 & 0.981 \\
    \cmidrule{2-6}
    
    &{$\beta_x$} & 0.995 & 0.980 & 0.845 & 0.926 \\
    &{BNN~\cite{mukhoti2018evaluating}} & 0.996 & 0.986 & 0.881  & 0.944 \\
    &{HIS~\cite{Atigh_2022_CVPR}} & - & - & 0.835 & 0.902 \\
    &\textbf{Ours} ($e_x$) & \textbf{0.997} & \textbf{0.987} & \textbf{0.909} & \textbf{0.951}\\
    \bottomrule
    \end{tabular}}
\label{tab:auc}
\vspace{-15pt}
\end{table}

In the next experiment, we evaluate the ability of local compensation to locate potential prediction errors in the label estimation step during the annotation process. 
To verify the ability of noise detection, we mime the label correction process on the validation datasets by replacing the ratio $r_\text{Area}$ of the most uncertain pixels of the model prediction with ground truth data. We estimate our uncertainty $e_x$ from Eq.~\eqref{equ:uncertainty} and compare it with the na\"ive $\beta_x$, a \textit{Bayesian Neural Network} (BNN) approach~\cite{mukhoti2018evaluating}, and with the hyperbolic segmentation framework (HIS) from Atigh \textit{et al.}~\cite{Atigh_2022_CVPR} for the large-scale datasets.
With these uncertainties, we evaluate the accuracy with increasing~$r_\text{Area}$.
The results are shown in Fig.~\ref{fig:noise_group} and the corresponding area-under-curve metrics are provided in Tab.~\ref{tab:auc}. The refined uncertainty estimation $e_x$ boosts $\beta_x$ and provides trustworthy guidance for label correction. For example, our method indicates approx. $50\%$ of all prediction errors on Cityscapes with high precision by selecting less than $5\%$ of the area. 
Our lightweight approach continuously outperforms the computational expensive BNN as well as the hyperbolic approach HIS. 
Note that the annotator is interested in high accuracy for small $r_\text{Area}$, since the area needs to be inspected manually. 

Image samples and the corresponding $e_x$, model predictions and ground truth data are visualized in Fig.~\ref{fig:teaser}. It shows the fine-grained uncertainty prediction and how $e_x$ can be used to identify prediction errors. Interestingly, we found multiple inconspicuous errors in the ground truth data, which are predicted correctly by our trained models (see the highlighted traffic-light).
More samples and more detailed figures are provided in the supp. material, Sec.~E.

\subsection{Robust Compensation: Training with Noise}
\label{sec:label_noise}
\begin{table}[t]
    \centering
    \caption{Mean intersection over union of our method compared to the baseline frameworks \textit{DeepLabv3+} and \textit{SegFormer} and applied robust learning methods 
    \textit{LogComp}, \textit{s-model}, and \textit{c-model}. Note that \textit{c-model} is not applicable to large-scale experiments. }
    \vspace{-5pt}
    \resizebox{\columnwidth}{!}{%
    \begin{tabular}{c | c c c c | c }
    \toprule
    mIoU $\uparrow$ & Cityscapes & KITTI & COCO & ADE20K & Mean
    \\\midrule
    \multicolumn{1}{l|}{\textit{DeepLabv3+}} & 0.797 & 0.570 & \textbf{0.358} & \textbf{0.431} & 0.539\\
    \multicolumn{1}{l|}{\ \ \ + \textit{LogComp}~\cite{yao2021compensation}} & 0.783 & 0.567 & 0.356 & \textbf{0.431} & 0.534\\
    \multicolumn{1}{l|}{\ \ \ + \textit{s-model}~\cite{Goldberger2017TrainingDN}} & \textbf{0.799} & 0.569 & 0.329 & 0.315 & 0.503\\
    \multicolumn{1}{l|}{\ \ \ + \textit{c-model}~\cite{Goldberger2017TrainingDN}} & 0.219 & 0.137 & - & - & -\\
    \multicolumn{1}{l|}{\ \ \ + \textbf{Ours}} & 0.\textbf{799} & \textbf{0.574} & 0.354 & 0.429 &0.539 \\
    \multicolumn{1}{l|}{\ \ \ + \textbf{Ours (+ Sym)}} & 0.798 & 0.572 & \textbf{0.358} & \textbf{0.431} & \textbf{0.541}\\
    \midrule    
    \multicolumn{1}{l|}{\textit{SegFormer}} & \textbf{0.821} & 0.625 & 0.413 & 0.482 & 0.585\\
    \multicolumn{1}{l|}{\ \ \ + \textit{LogComp}~\cite{yao2021compensation}} & {0.785} & 0.628 & {0.427} & {0.471} & 0.578\\
    \multicolumn{1}{l|}{\ \ \ + \textit{s-model}~\cite{Goldberger2017TrainingDN}} & \textbf{0.821} & 0.570 & 0.430 & 0.482 & 0.576\\
    \multicolumn{1}{l|}{\ \ \ + \textit{c-model}~\cite{Goldberger2017TrainingDN}} & 0.492 & 0.489 & - & - &-\\
    \multicolumn{1}{l|}{\ \ \ + \textbf{Ours}} & 0.816 & 0.649 & 0.428 & 0.459 & 0.596\\
    \multicolumn{1}{l|}{\ \ \ + {\textbf{Ours (+ Sym)}}} & \textbf{0.821} & \textbf{0.658} & \textbf{0.432} & \textbf{0.485} & \textbf{0.599}\\
    \bottomrule
    \end{tabular}}
\label{tab:metrics}
\vspace{-10pt}
\end{table}

\begin{table}[t]
    \centering
    \caption{Average prediction uncertainty with and without compensation learning in the baseline framework \textit{DeepLabv3+} indicating overfitting, \textit{a}.\textit{k}.\textit{a}. memorization effect~\cite{Liu_2022_CVPRa}. Note that we removed all compensation related weights during inference.}
    \vspace{-5pt}
    \resizebox{\columnwidth}{!}{%
    \begin{tabular}{c | c | c c c c}
    \toprule
    $u_x\uparrow$ & Compensation & Cityscapes & KITTI & COCO & ADE20K
    \\\midrule
    \multirow{2}{*}{\rotatebox{90}{Train}} & \ding{55} & 0.027 & 0.028 & 0.076 & 0.100 \\
    & \ding{51} & \textbf{0.048} & \textbf{0.035} & \textbf{0.081} & \textbf{0.122} \\
    \midrule    
    \multirow{2}{*}{\rotatebox{90}{Val}} & \ding{55} & 0.035 & 0.039 & 0.199 & 0.141 \\
    & \ding{51} & \textbf{0.058} & \textbf{0.048} & \textbf{0.204} & \textbf{0.160} \\
    \bottomrule
    \end{tabular}}
\label{tab:uncertainty}
\vspace{-10pt}
\end{table}

\begin{figure}[t]
\centering
\resizebox{\columnwidth}{!}{
\begin{tikzpicture}
\begin{axis}[
    title=Performance with Noise Induction,
    scaled ticks=false, 
    log ticks with fixed point,
    tick label style={/pgf/number format/fixed},
    xmin = 0, xmax = 100,
    ymin = 0.52, ymax = 0.6,
    xtick distance = 20,
    ytick distance = 0.05,
    grid = both,
    minor tick num = 2,
    major grid style = {lightgray},
    minor grid style = {lightgray!25},
    width = 1\columnwidth,
    height = 0.5\linewidth,
    xlabel = {$n$},
    ylabel = {mIoU},
    legend pos = north east,
    legend style = {
      legend columns=2},
]
\addplot[
    thick,
    red,
] file[skip first] {noise_influence_step_ours.txt};
\addplot[
    blue,
    dashed
] file[skip first] {noise_influence_step_transition.txt};
\addplot[
    black,
] file[skip first] {noise_influence_step_none.txt};
\addplot[
    purple,
    dashed
] file[skip first] {noise_influence_step_log_comp.txt};
\legend{Ours, \textit{s-model}, \textit{DeepLabv3+}, \textit{LogComp}}
\end{axis}
\end{tikzpicture}
}
\vspace{-20pt}
\label{fig:accuracy_noise_step}
\caption{Performance degradation of segmentation quality after induction of label flips with noise parameter $n$ showing the impact of our method on the robustness of \textit{DeepLabv3+} compared to \textit{LogComp} and \textit{s-model}.
}
\vspace{-12pt}
\label{fig:accuracy_noise}
\end{figure}
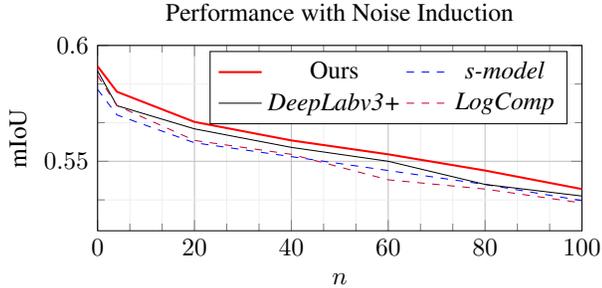

Compensation learning decreases the impact of ambiguities on the loss during training and therefore enlarges the robustness against ambiguity based label noise and overfitting. To evaluate the impact of our method on the robustness, we first evaluate the impact of our method with and without the optional symmetry constraint on the segmentation performance of \textit{SegFormer} and \textit{DeepLabv3+} (see Tab.~\ref{tab:metrics}). To compare the results, we evaluate the simple \textit{s-model} and complex \textit{c-model} transition matrix approach as described in \cite{Goldberger2017TrainingDN}.
While \textit{s-model} and \textit{c-model} show instabilities in either large and/or small scale datasets, our method achieves the same or higher mIoU than the baseline. We also evaluate the na\"ive and unconstrained compensation method \textit{LogComp} as proposed by Yao \textit{et al.}~\cite{yao2021compensation} and show that our contributions are mandatory to apply compensation learning in semantic segmentation. Especially the transformer based \textit{SegFormer} benefits on the small dataset KITTI-STEP and increases the mIoU by $3.3$ percent points.    

Moreover, we induce human-like label noise between similar classes and corrupt the ground truth data during training. Following the approach of Liu \textit{et al.}~\cite{Liu_2022_CVPRa}, we dilate the area of predefined classes with neighboured pixels of similar appearance that have a distance of at most $n$ pixels. Comparable noise patterns can be found in the ground truth data of our datasets. Fig.~\ref{fig:accuracy_noise} shows the accuracy degradation after inducing different label noise levels. While \textit{LogComp} and \textit{s-model} decrease the accuracy of the baseline, our method continuously improves the performance. Note that we restrict this expensive experiment to \textit{DeepLabv3+} with KITTI-STEP and reduce the batchsize to $4$ for ecological reasons. For detailed information about noise induction and visual samples, see supplementary material, Sec.~C.

To show the influence of compensation against overfitting, we measure the average model uncertainty $u_x$ of \textit{DeepLabv3+} trained with and without compensation. Because overfitting manifests in certain predictions for ambiguous or noisy labels, also known as memorization~\cite{Liu_2022_CVPRa}, a robust model should stay uncertain for those uncertain regions even after long training. The comparison of the average model uncertainty $u_x$ is given in Tab.~\ref{tab:uncertainty}. On validation and training data, the model with our proposed method is significantly more uncertain with a factor up to $1.7$. 

Overall, the experiments show that our method is able to learn interpretable guidance for label correction while improving robustness against noise by avoiding memorization.

\subsection{Compensated Inference: Bias Induction}
\label{sec:bias_induced_experiment}

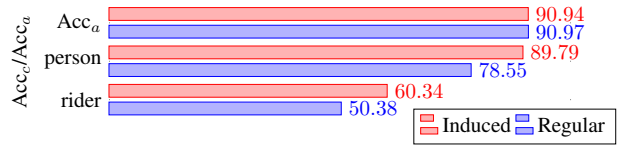
\begin{figure}[t]
\centering
\resizebox{\columnwidth}{!}{
\begin{tikzpicture}
    \begin{groupplot}[
        group style={group size= 1 by 2},
        width=1.1\columnwidth,
        xmin = 0, xmax = 100,
        xlabel = {$aAcc/cAcc$},
        ylabel = {$\text{Acc}_c$/$\text{Acc}_a$},
        xbar=0.08cm,
        y=18pt,
        y axis line style = { opacity = 0 },
        axis x line       = none,
        tickwidth         = 0pt,
        enlarge y limits  = 0.35,
        ytick             = data,
        legend style={at={(1.1,-0.0)},anchor=east},
        legend columns=3, 
        nodes near coords,
        reverse legend,
    ]
        \nextgroupplot[
                    symbolic y coords = {rider, person, $\text{Acc}_a$},
                    bar width=6pt,
                ] 
                \addplot coordinates {
                  (90.97,$\text{Acc}_a$)
                  (78.55,person)
                  (50.38,rider)
                };
                \addplot coordinates {
                  (90.94,$\text{Acc}_a$)
                  (89.79,person)
                  (60.34,rider)
                };
             \vspace{-50pt}
             \legend{Regular, Induced}
    \end{groupplot}
\end{tikzpicture}
}
\vspace{-15pt}
\caption{Induction Experiments on KITTI-STEP.}
\vspace{-15pt}
\label{fig:induction_kitti_short}
\end{figure}

To outline the possibilities of induced compensation during inference as explained in Sec.~\ref{sec:compensated_inference}, we present an exemplary application. In KITTI-STEP, vulnerable classes like \textit{rider} or \textit{person} are expected to be more important for applications in autonomous driving. Therefore, we amplify the segmentation likelihood of those classes during inference by manually defining elements in the compensation matrix $B$. The exact value of the elements is determined experimentally. We set $B_{ii}$ with $i \in \{\textit{person},\textit{rider}\}$ to $30$ and $B_{ij}$ with $j \in \{\textit{sidewalk},\textit{building}\}$ to $-8$. The resulting accuracy of the vulnerable classes is compared against the unmodified model in Fig.~\ref{fig:induction_kitti_short}. Without loosing notable accumulated accuracy, the accuracy of our selected vulnerable classes increases by approx. $10$ percent points. 
This shows the general ability to manipulate inference with compensations to improve annotation for given tasks with prioritized classes.
The application applied on Cityscapes and all class accuracy metrics can be found in the supplementary material, Sec.~F. Note that this experiment briefly outlines future possibilities, but rules to determine exact values in $B$ need to be further investigated.

\section{Conclusion}
\label{sec:conclusion}
In this paper, we present compensation learning in semantic segmentation, a lightweight approach to learn and visualize inter-class relations to tackle ambiguity based label noise during the annotation of new semantic segmentation datasets. Our method creates a ground truth depending bias to compensate the influence of similar classes and ambiguities. 
The experiments demonstrate that our compensation learning method provides global and local guidance in the label correction process and introduces a powerful uncertainty estimation approach.  
Moreover, it improves the robustness against conditional label noise and accurately detects prediction errors of segmentation networks. 
We have presented insights about challenging class pairs in the Cityscapes, KITTI-STEP, ADE20k, and COCO datasets. This contribution helps the community to make semantic segmentation more robust against inter-class ambiguities and subsequent label noise. 
\section{Acknowledgments}

This work was supported by the Federal Ministry of Education and Research (BMBF), Germany under the project LeibnizKILabor (grant no. 01DD20003) and the AI service center KISSKI (grant no. 01IS22093C), the Center for Digital Innovations (ZDIN) and the Deutsche Forschungsgemeinschaft (DFG) under Germany’s Excellence Strategy within the Cluster of Excellence PhoenixD (EXC 2122). 

% textwidth in cm: \printinunitsof{cm}\prntlen{\textwidth}
% textheight in cm: \printinunitsof{cm}\prntlen{\textheight}

\clearpage

%%%%%%%%% REFERENCES
{\small
\bibliographystyle{ieee_fullname}
\bibliography{egbib}
}

% \clearpage
% \input{6_appendix}   % <-- DO NOT FORGET TO REMOVE THIS FOR SUBMISSION!!!

\end{document}


\author{Timo Kaiser, Christoph Reinders, Bodo Rosenhahn\\
Institute for Information Processing (tnt)\\
L3S - Leibniz Universität Hannover, Germany\\
%Institut für Informationsverarbeitung and L3S Research Center\\% Leibniz University Hannover\\
%Hannover, Germany\\
{\tt\small \{kaiser, reinders, rosenhahn\}@tnt.uni-hannover.de}
% \and
% Christoph Reinders
}

\title{Supplementary Material: Compensation Learning in Semantic Segmentation}
\twocolumn[{%
\renewcommand\twocolumn[1][]{#1}%
\maketitle
}]

\newcommand{\todo}[1]{\iftrue\textcolor{blue}{#1}\fi}

\appendix

% \title{Supplementary Material: Compensation Learning in Semantic Segmentation}
% \twocolumn[{%
% \renewcommand\twocolumn[1][]{#1}%
% \maketitle
% }]

% \renewcommand{\thesection}{\Alph{section}}
% \setcounter{section}{0}%
% \setcounter{subsection}{0}%
% % \setcounter{figure}{5}
% % \setcounter{table}{1}
% % \setcounter{equation}{12}
% \setcounter{footnote}{0}%
% \setcounter{page}{1}%

We present supplemental material for our paper \textit{Compensation Learning in Semantic Segmentation} to disclose more details about the method and experiments. In the first section, we present detail of the implementation of our and the referenced methods. The second section shows mathematical details of the metrics used in our experiments. The third section elaborates the process of noise induction used in the main paper. Then, more fine-grained results of the bias induced inference experiments are presented in section four. Finally, the fifth and sixth section provide the full compensation and confusion matrices, mentioned in the main paper.

\section{Experimental Setup}
\label{appendix:experimental_setup}
To simplify the reproduction process, we integrate our method as well as the referenced methods into the well-known framework \textit{MMSegmentation}~\cite{mmseg2020}, if no suitable code was available. \textit{MMSegmentation} provides code for multiple state-of-the-art semantic segmentation frameworks. Also our baseline methods \textit{DeepLabv3+} and \textit{SegFormer} are implemented in \textit{MMSegmentation} with different hyperparameter settings for different datasets. Reference links to the exact configurations for our baseline experiments can be found in Tab.~\ref{tab:apendix_experimental_settings}. The next section elaborates on how the referenced methods and datasets used in the main paper are integrated and configured. 

The code for our method and some mentioned competitors are maintained in a local github fork\footnote{\href{https://github.com/tnt-LUH/compensation\_learning}{https://github.com/tnt-LUH/compensation\_learning}}. We want to note that we endeavour to merge the code into the widely-used public repository\footnote{ \href{https://github.com/open-mmlab/mmsegmentation}{https://github.com/open-mmlab/mmsegmentation}} in the future to enlarge the visibility.

\subsection{Model Architectures}
To compare our method, reference methods are evaluated in the main paper. Except~\textit{Hyperbolic Image Segmentation}, the methods \textit{s-model}, \textit{c-model}, and a \textit{Bayesian Neural Network} methods are integrated on top of the baseline frameworks \textit{DeepLabv3+} and \textit{SegFormer}. The following sections elaborates the implementation of all reference methods.   

\begin{table*}[t]
    \centering
    \caption{Model configurations references to reproduce the training in \textit{MMSegmentation}. Note that learning rate decay as in~\cite{chen2017deeplab} is applied.}
    \vspace{-5pt}
    \resizebox{\linewidth}{!}{%
    \begin{tabular}{c | c | c | c | c| c | c}
    \toprule
     & Dataset & Training Crop-Size & Steps & Learning Rate & Optimizer & Reference Configuration File
    \\\midrule
    \multirow{4}{*}{\rotatebox{90}{\scalebox{0.8}{\textit{DeepLabv3+}}}} & Cityscapes & $512\times1024$& $\num{80000}$ & $1\cdot10^{-2}$& SGD &  \href{https://github.com/open-mmlab/mmsegmentation/blob/master/configs/deeplabv3plus/deeplabv3plus_r50-d8_512x1024_80k_cityscapes.py}{deeplabv3plus\_r50-d8\_512x1024\_80k\_cityscapes.py} \\
    & KITTI-STEP & $368\times368$ & $\num{80000}$ & $1\cdot10^{-2}$ & SGD &\href{https://github.com/tnt-LUH/compensation\_learning/blob/master/configs/deeplabv3plus/deeplabv3plus\_r50-d8\_368x368\_80k\_kittistep.py}{deeplabv3plus\_r50-d8\_368x368\_80k\_kittistep.py} \\
    & COCO-stuff10k & $512\times512$& $\num{80000}$ & $1\cdot10^{-2}$ & SGD & \href{https://github.com/tnt-LUH/compensation\_learning/blob/master/configs/deeplabv3plus/deeplabv3plus\_r50-d8_512x512\_80k\_coco-stuff10k.py}{deeplabv3plus\_r50-d8\_512x512\_80k\_coco-stuff10k.py}\\
    & ADE20k & $512\times512$& $\num{80000}$ & $1\cdot10^{-2}$ & SGD & \href{https://github.com/open-mmlab/mmsegmentation/blob/master/configs/deeplabv3plus/deeplabv3plus_r50-d8_512x512_80k_ade20k.py}{deeplabv3plus\_r50-d8\_512x512\_80k\_ade20k.py}\\
    \midrule    
    \multirow{4}{*}{\rotatebox{90}{\scalebox{0.8}{\textit{SegFormer}}}} & Cityscapes & $1024\times1024$& $\num{160000}$ & $6 \cdot 10^{-5}$ & Adam & \href{https://github.com/open-mmlab/mmsegmentation/blob/master/configs/segformer/segformer_mit-b5_8x1_1024x1024_160k_cityscapes.py}{segformer\_mit-b5\_8x1\_1024x1024\_160k\_cityscapes.py} \\
    & KITTI-STEP & $368\times368$ & $\num{160000}$ & $6 \cdot 10^{-5}$ & Adam & \href{https://github.com/tnt-LUH/compensation\_learning/blob/master/configs/segformer/segformer\_mit-b5\_368x368\_160k\_kittistep.py}{segformer\_mit-b5\_368x368\_160k\_kittistep.py} \\
    & COCO-stuff10k & $512\times512$ & $\num{160000}$ & $6 \cdot 10^{-5}$ & Adam & \href{https://github.com/tnt-LUH/compensation\_learning/blob/master/configs/segformer/segformer\_mit-b5\_512x512\_160k\_coco-stuff10k.py}{segformer\_mit-b5\_512x512\_160k\_coco-stuff10k.py}\\
    & ADE20k & $512\times512$ & $\num{160000}$ & $6 \cdot 10^{-5}$ & Adam & \href{https://github.com/open-mmlab/mmsegmentation/blob/master/configs/segformer/segformer_mit-b5_512x512_160k_ade20k.py}{segformer\_mit-b5\_512x512\_160k\_ade20k.py}\\
    \bottomrule
    \end{tabular}}
\label{tab:apendix_experimental_settings}
\vspace{-10pt}
\end{table*}

% pes~\cite{Cordts2016Cityscapes}, KITTI-STEP~\cite{Weber2021NEURIPSDATA}, ADE20k~\cite{zhou2017scene}, and COCO-stuff10k~\cite{10.1007/978-3-319-10602-1_48}

\subsubsection*{{s-model} and {c-model}}
The methods \textit{s-model} and \textit{c-model} are referenced in the robust learning experiments in Sec.~4.4 in the main paper. Both methods are introduced in~\cite{Goldberger2017TrainingDN} 
and implement a trainable transition matrix into a neural network. The transition matrix $T$ encodes the global noise distribution and can be used to estimate the clean probability $P(Y=i|x)$ for class $i$ from a noisy probability prediction $\overline{P}(Y=i|x)$ of a pixel $x$. For the pixel $x$ and a set of classes $C$, this estimation can be formulated as 
\begin{equation}
\label{equ:s-model}
    \forall x\in I: \quad \Vec{\overline{p}}_x=T\Vec{p}_x,\quad T\in [0,1]^{C\times C} 
\end{equation}
with the additional condition
\begin{equation}
\label{equ:s-model-condition}
    \forall j\in C: \quad \sum_i T_{ij} = 1 
\end{equation}
to conserve a probability distribution. During training, the matrix $T$ is optimized alongside the model parameters to fit on the noisy labels, and during inference of unseen data, Eq.~\eqref{equ:s-model} is removed from the framework to obtain the clean probabilities $\Vec{p}_x$. 

To implement the simple \textit{s-model} from~\cite{Goldberger2017TrainingDN}, we transfer Eq.~\eqref{equ:s-model} into the two-dimensional case by modeling $T$ as a convolutional layer with kernel size 1 (\textit{a.k.a.} pointwise CNN), and input/output-channels of $C$. This convolution is applied to the final two-dimensional probability map of the underlying baseline framework. 
To satisfy Eq.~\eqref{equ:s-model-condition}, the weights of the convolutional layer ($=T$) are normalized using the softmax function (see Eq. 1 in the main paper) on each column before convolving the
two-dimensional probability map containing $\Vec{p}_x$ for all pixels $x$.

To implement the complex \textit{c-model} from~\cite{Goldberger2017TrainingDN}, we extend \textit{s-model}. In \textit{s-model}, a shared transition matrix $T$ is used for all pixels in all images globally. Compared to the global approach, \textit{c-model} estimates the transition matrix individually for every pixel based on the high-level image features. To implement this into the segmentation framework, we employed an additional branch parallel to the segmentation head similar to the implementation of our local uncertainty branch. The branch contains two pointwise two-dimensional convolutional layers. The first layer has input channels depending on the dimensions of the high-level image features and 32 output channels. This layer is used to reduce the complexity to a tractable size. The second layer has an input size of 32 and an output size of $|C|^2$. The $|C|^2$-dimensional output for pixel $x$ is reshaped to a $C\times C$ matrix, normalized with softmax, and used as individual weight for the final convolution described above for \textit{s-model}. We want to note that the computational effort for large-scale datasets with a large amount of classes like ADE20k or COCO-stuff10k is not computable with current state-of-the-art graphic cards and segmentation frameworks. Therefore the experiments for \textit{c-model} and the large-scale datasets cannot be evaluated.

The code integration of \textit{s-model} and \textit{c-model} into \textit{MMSegmentation} can be found \href{https://github.com/tnt-LUH/compensation\_learning\#NAL}{here}\footnote{\href{https://github.com/tnt-LUH/compensation\_learning\#NAL}{https://github.com/tnt-LUH/compensation\_learning\#NAL}}.

\subsubsection*{Bayesian Neural Network}
Mukhoti \textit{et al.}~\cite{mukhoti2018evaluating} present a \textit{Bayesian Neural Network} (BNN) that implements \textit{Monte-Carlo Dropout} into \textit{DeepLabv3+} with a comparable backbone as used in our experiment.  \textit{Monte-Carlo Dropout} consists of basically two modifications of the baseline framework: Dropout layers with high dropout probability $d$ are applied to high-level feature maps during training and during test time, the inference is repeated multiple times with active dropout layers to generate sample predictions from a Bernoulli-distributed weight distribution. The mean of the generated samples is the most likely correct prediction and the variance can be interpreted as the prediction uncertainty.

The implementation of a BNN into \textit{DeepLabv3+} is described in~\cite{mukhoti2018evaluating} for the backbone~\textit{Xception}. To be comparable, we implemented their approach into the \textit{ResNet50} backbone~\cite{He_2016_CVPR} used in the main paper using the same design choices.
Following the design choice of only modifying the \textit{Middle Flow} of \textit{Xception}, we add dropout layers to the third stage of the \textit{ResNet50}. We add the dropout layers after each residual block as described in the Appendix of~\cite{mukhoti2018evaluating}.
Furthermore, Mukhoti \textit{et al.} remove the \textit{Atrous Spatial Pyramid Pooling} (ASPP) modules from \textit{DeepLabv3+} to reduce the complexity. Because ASPP modules increase the accuracy, we do not remove them. 

We optimized the dropout rate and noticed, that $d=0.5$ leads to significant drop in the accuracy, while $d=0.25$ conserves the accuracy and also predicts satisfying uncertainty. Thus, we set $d$ to $0.25$ for all dropout layers. The mean and variance during test time is obtained with 20 samples.  

The code integration of the Bayesian ResNet into \textit{MMSegmentation} can be found \href{https://github.com/tnt-LUH/compensation\_learning\#BayesianNN}{here}\footnote{\href{https://github.com/tnt-LUH/compensation\_learning\#BayesianNN}{https://github.com/tnt-LUH/compensation\_learning\#BayesianNN}}.

\subsubsection*{Hyperbolic Image Segmentation}
The hyperbolic image segmentation framework (HIS) of Atigh~\textit{et al.}~\cite{Atigh_2022_CVPR} is used in Sec.~4.3 in the main paper to compare uncertainty estimation of our and other methods with respect to the label prediction correction task. The approach HIS\footnote{\href{https://github.com/MinaGhadimiAtigh/HyperbolicImageSegmentation}{https://github.com/MinaGhadimiAtigh/HyperbolicImageSegmentat}} provides code for training and inference for the datasets ADE20k and COCO-stuff10k. We trained the models for the datasets with original hyperparameters and created normalized uncertainty maps for every image of both datasets. We noticed that there are high-valued fragments at the border of the uncertainty prediction, which distort the subsequent normalization process. Thus, we set the uncertainty of the 10 pixels closest to the border to zero, to allow reasonable evaluation w.r.t. the prediction error correction task. We do not evaluate HIS for Cityscapes and KITTI-STEP, because optimized hyperparameters and evaluations are not given in~\cite{Atigh_2022_CVPR} and we want to avoid unfair comparisons.

\subsection{Datasets}
\textit{MMSegmentation} provides a standard setting including augmentations and data pre-processing methods. In the following, the configuration for the non-preconfigured dataset KITTI-STEP is explained. 

% \subsubsection*{COCO-stuff10k}
% The dataset COCO-stuff10k is a sub-set of the large-scale dataset COCO-stuff164k~\cite{10.1007/978-3-319-10602-1_48}. 
% An augmentation and pre-processing pipeline for the large-scale dataset COCO-stuff164k is defined in \href{https://github.com/open-mmlab/mmsegmentation/blob/master/configs/_base_/datasets/coco-stuff164k.py}{coco-stuff164k.py}. We derive the configuration for COCO-stuff10k by using the identical setup as for COCO-stuff164k.

% \textcolor{blue}{The link to the configuration file in \textit{MMSegmentation} will be published after review to satisfy the blind review procedure.}

\subsubsection*{KITTI-STEP}
The KITTI-STEP~\cite{Weber2021NEURIPSDATA} dataset shares the same classes and domain as the dataset Cityscapes~\cite{Cordts2016Cityscapes}. Augmentation and pre-processing for Cityscapes is defined in \href{https://github.com/open-mmlab/mmsegmentation/blob/master/configs/_base_/datasets/cityscapes.py}{cityscapes.py}. We use similar configurations than in Cityscapes, but we changed crop size in the pre-processing to $368\times368$. Images in KITTI-STEP are significantly smaller than in Cityscapes. 
The configuration file in \textit{MMSegmentation} can be found in {\href{https://github.com/tnt-LUH/compensation\_learning/blob/master/configs/\_base\_/datasets/kittistep.py}{kittistep.py}}.

\section{Metrics}
\label{appendix:metrics}
We evaluate with the widely-used {mIoU} metric~\cite{thoma2016survey}, which is also referenced as \textit{Semantic Quality}~\cite{Long_2015_CVPR, Weber2021NEURIPSDATA}. With the set of pixels $X_c$ that are assigned to class $c$ and ground truth labels $\hat{Y}_c$ annotated as $c$, the mIoU metric is defined as 
\begin{equation}
\label{equ:miou}
\text{mIoU}=\frac{1}{|C|}\sum_{c\in C}\frac{|X_c\cap \hat{Y}_c|}{|X_c\cup \hat{Y}_c|}\ .
\end{equation}
The {mIoU} metric evaluates the assignment of class labels and balances underrepresented classes. We also use the class accuracy ($\text{Acc}_c$) and aggregated accuracy ($\text{Acc}_a$)  
\begin{equation}
\label{equ:acc_i}
\text{Acc}_c=\frac{|X_c\cap \hat{Y}_c|}{|\hat{Y}_c|}\quad \text{and} \quad \text{Acc}_a=\frac{\sum_{c\in C}|X_c\cap \hat{Y}_c|}{\sum_{c\in C}|\hat{Y}_c|}
\end{equation}
to verify segmentation results on pixel level. 

\section{Noise Induction}
\label{appendix:noise_induction}
We superficially describe the process of inducing synthetic label noise during training in the main paper, Sec.~4.4. This section elaborates this process in detail. The work of Liu \textit{et al.}~\cite{Liu_2022_CVPRa} corrupts ground truth annotations for medical images by dilating or eroding the masks of specific instances. We adapt this process for the multi-class datasets used in this paper. 

With the help of the confusion matrices of Cityscapes and KITTI-STEP, we select some ambiguous class pairs that are not underrepresented and are often located in neighboured regions in images. Overall, we selected the class pairs \textit{road}-\textit{sidewalk}, \textit{building}-\textit{wall}, and \textit{vegetation}-\textit{terrain}.

Before we train a method with noise induction, we randomly sample a superior and inferior class for each class pair and each image. In an exemplary configuration, \textit{road} could be superior in image A and \textit{sidewalk} could be superior in image B. During the training process, the sampled class selection is fixed. 

Then, we introduce a hyperparameter $n$, which reflects the noise level in the next step. We dilate regions of the superior classes, by adding pixels of the inferior class that are at most $n$ pixels far away to a pixel of the superior class. This pattern mimes the uncertainty during human annotation. 

Some examples with the RGB image, the original ground truth, and the ground truth after noise induction are shown in Fig.~\ref{fig:noise}. The resulting label noise is close to human-like noise patterns.

\section{Top-$k$ uncertain classes}
\label{appendix:k_uncertain_classes}
\begin{figure}[t]
\resizebox{\columnwidth}{!}{\begin{tikzpicture}
\begin{axis}[
    title=Analysis of Hyperparameter $k$,
    scaled ticks=false, 
    log ticks with fixed point,
    tick label style={/pgf/number format/fixed},
    xmin = 0, xmax = 19,
    ymin = 0.0, ymax = 0.01,
    xtick distance = 5,
    ytick distance = 0.01,
    grid = both,
    minor tick num = 4,
    major grid style = {lightgray},
    minor grid style = {lightgray!25},
    width = 1.15\columnwidth,
    height = 0.4\linewidth,
    xlabel = {$k$},
    ylabel = {$|\Delta e_x|$ },
    legend pos = north east,
    legend style = {
      legend columns=2},
]
\addplot[
    thick,
    red,
] file[skip first] {top_k.txt};
\legend{Ours ($e_x$)}
\end{axis}
\end{tikzpicture}}
\vspace{-20pt}
\caption{Mean difference of $e_x$ with dynamic $k$ against \mbox{$k=|C|$}.}
\label{fig:appendix_plot_k}
\vspace{-15pt}
\end{figure}
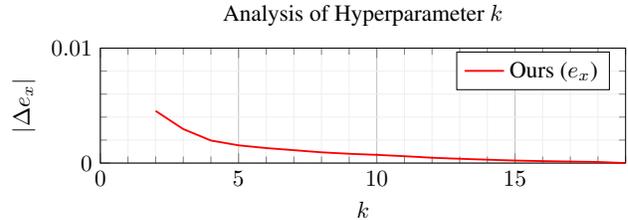

\begin{figure*}[t]
    \centering
    \includegraphics[width=1\textwidth]{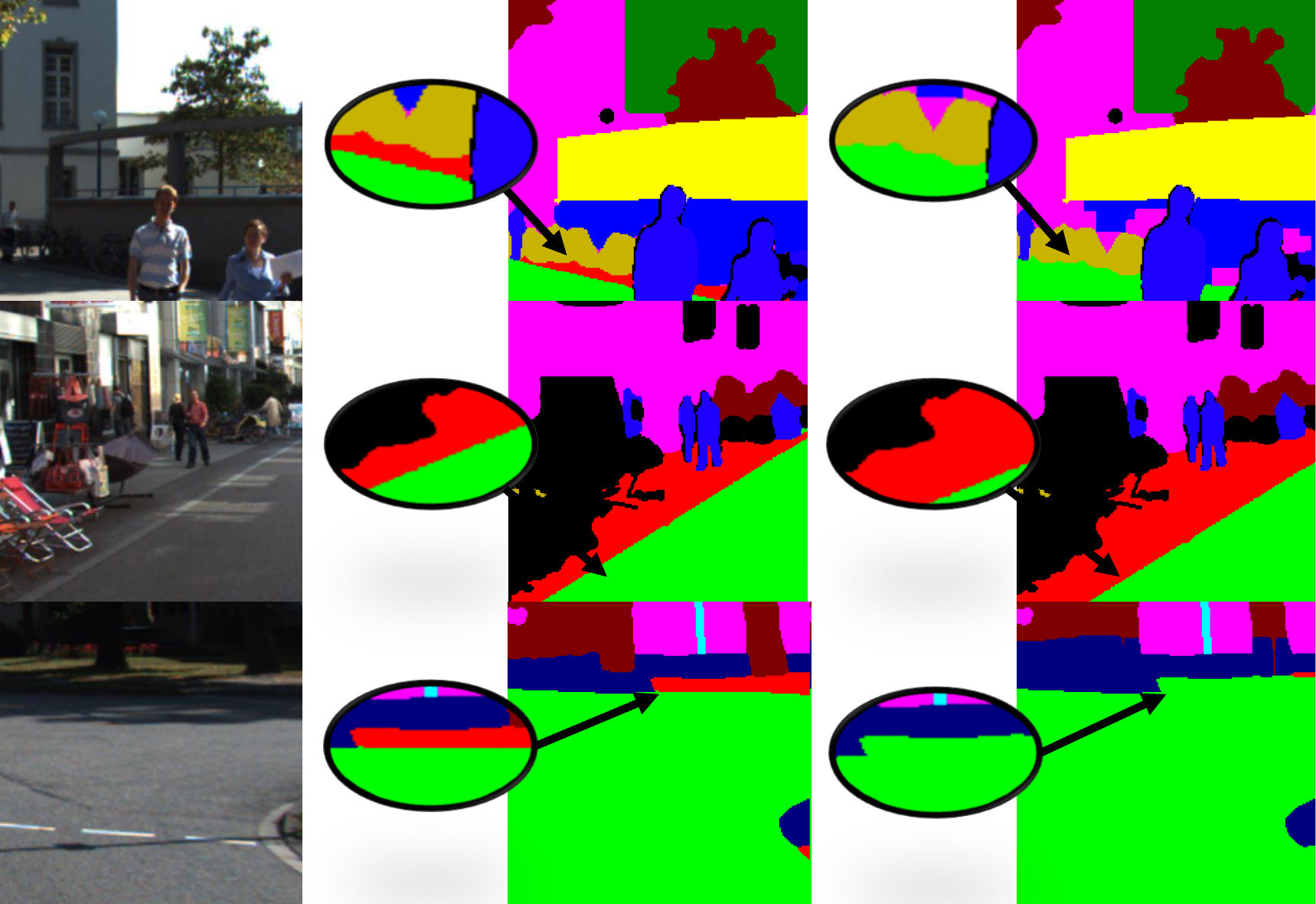}
    \caption{
    Synthetic ground truth degradation. Left image shows a training sample after random cropping. The center image contains ground truth annotations. The degraded ground truth annotations after applying our synthetic noise reduction with $n=20$ are shown in the right image.}
    \label{fig:noise}
    %\vspace{-8pt}
\end{figure*}

We introduced the hyperparameter $k$ in Sec.~3.3 and set it to 5 to reduce the computational effort. 
Because additional experiments showed that $k$ has neglectable impact on $\text{Acc}_a$, we measure the absolute difference of $e_x$ with varying $k$ against a fixed $k=|C|$. Fig.~\ref{fig:appendix_plot_k} presents the mean absolute difference $|\Delta e_x|$ over all pixels on Cityscapes. It shows that small $k$ has most impact on the variance of $e_x$ and additional computation with large $k$ leads to marginal benefits.

\vspace{5pt}

\section{Noise Detection}
\label{appendix:noise_detection}
In the main paper, we analyzed the ability of noise detection in Sec.~4.3. Therefore, the accuracy with respect to the pixel area was reported in Fig.~4. The full plots are shown in larger resolution in Fig.~\ref{fig:noise_city} to~\ref{fig:noise_coco10k}.

\section{Bias Induced Inference}
\label{appendix:inductive_inference}
This section provides more fine-grained results for the bias induced inference experiments presented in Sec.~4.5. In the main paper, we present how the accuracy for specific classes can be boosted without loosing significant accuracy over all classes. The Fig.~\ref{fig:induction_kitti} shows the accuracy of all classes in the experiment for the dataset KITTI-STEP. Most classes are not affected by the induction. The only exceptions are the classes \textit{motorcycle} and \textit{bicycle}, which are closely related to the induced classes. Moreover, we report the same experiment without induction for class \textit{rider}. Since \textit{person} and \textit{rider} are ambiguous, the accuracy of class \textit{rider} diminishes without induction. 

We repeat the experiment for the similar dataset Cityscapes and observe the same behaviour. The results are presented in Fig.~\ref{fig:induction_city}.

\section{Compensation Matrix}
\label{appendix:qualitative_results}
To complete the qualitative results presented in the main paper (see Sec.~4.2, main paper) and to give more dataset specific insights with the learned compensation values in the compensation matrix $B$, this section presents the complete learned compensation matrices for the segmentation frameworks \textit{DeepLabv3+} and \textit{SegFormer} trained with the datasets Cityscapes, KITTI-STEP, ADE20K, and COCO10k-stuff. 
The compensation weights for the small-scale datasets are presented in Tab.~\ref{tab:full_compensation_weights_kitti_deeplab} and~\ref{tab:full_compensation_weights_cityscapes_deeplab} (\textit{DeepLabv3+}) as well as in Tab.~\ref{tab:full_compensation_weights_kitti_segformer} and~\ref{tab:full_compensation_weights_cityscapes_segformer} (\textit{SegFormer}).
The compensation weights for the large-scale datasets are reported in Fig.~\ref{fig:large_scale_compensation}. We recommend to inspect the figure details in the \textit{pdf} version instead of a printed version.

Additional interesting insights are that the matrices for \textit{DeepLabv3+} approximately form symmetrical, which is not happening for \textit{SegFormer}. Moreover, the compensation weights for \textit{SegFormer} have higher values, compared to \textit{DeepLabv3+}. This is an indicator for the higher impact of compensation learning on transformer-based architectures.

\section{Confusion Matrix}
\label{appendix:confusion}
The compensation matrices of Sec.~\ref{appendix:qualitative_results} can be compared to the confusion matrices. The confusion matrices for Cityscapes and KITTI-STEP generated on the validation and training dataset are presented in Tab.~\ref{tab:confusion_matrix_cityscapes_train} to~\ref{tab:confusion_matrix_kitti_val}. It is notable that the confusion matrices on the training data does not behave like the confusion matrices on the unseen validation data. For example, in the Cityscapes and KITTI-STEP validation dataset, the class \textit{building} interferes much with the classes \textit{wall} and \textit{fence}. These correlations are not equally represented in the training dataset. In practice, this behaviour leads to the need of an additional labeled dataset to obtain a suitable confusion matrix for unseen data during inference. Compared to this, our compensation weights are obtained during training without the need of additional data.

\begin{figure}[t]
\centering
\begin{tikzpicture}
\begin{axis}[
    title=Error Detection Rate Cityscapes,
    scaled ticks=false, 
    log ticks with fixed point,
    tick label style={/pgf/number format/fixed},
    xmin = 0, xmax = 1.0,
    ymin = 0.96, ymax = 1.,
    xtick distance = 0.25,
    ytick distance = 0.01,
    grid = both,
    minor tick num = 2,
    major grid style = {lightgray},
    minor grid style = {lightgray!25},
    width = 1\linewidth,
    height = 0.7\linewidth,
    xlabel = {$r_\text{Area}$},
    ylabel = {$\text{Acc}_a$},
    legend style = {
      at={(0.5, -0.4)},
      anchor=north,
      legend columns=4},
    domain=0:1,
]
                \addplot[
                    thick,
                    red,
                ] file[skip first] {detection_city_val_Combined.txt};
                \addplot[
                    black,thick,
                ] file[skip first] {detection_city_val_Alpha.txt};
                \addplot[
                    blue,thick,
                ] file[skip first] {detection_city_val_Bayes.txt};
                \addplot[
                    dashed,
                    black,
                ]{0.96+x};
\legend{$e_x$, $\beta_x$, BNN, Oracle}
\end{axis}
\end{tikzpicture}
\vspace{-15px}
\caption{Noise detection experiment on Cityscapes extending Fig.~4 in the main paper.}
\label{fig:noise_city}
\end{figure}
\begin{figure}[t]
\centering
\begin{tikzpicture}
\begin{axis}[
    title=Error Detection Rate KITTI-STEP,
    scaled ticks=false, 
    log ticks with fixed point,
    tick label style={/pgf/number format/fixed},
    xmin = 0, xmax = 1.0,
    ymin = 0.9, ymax = 1.,
    xtick distance = 0.25,
    ytick distance = 0.025,
    grid = both,
    minor tick num = 2,
    major grid style = {lightgray},
    minor grid style = {lightgray!25},
    width = 1\linewidth,
    height = 0.7\linewidth,
    xlabel = {$r_\text{Area}$},
    ylabel = {$\text{Acc}_a$},
    legend style = {
      at={(0.5, -0.4)},
      anchor=north,
      legend columns=4},
]
                \addplot[
                    thick,
                    red,
                ] file[skip first] {detection_kitti_val_Combined.txt};
                \addplot[
                    black,thick,
                ] file[skip first] {detection_kitti_val_Alpha.txt};
                \addplot[
                    blue,thick,
                ] file[skip first] {detection_kitti_val_Bayes.txt};
                \addplot[
                    dashed,
                    black,
                ]{0.909+x};

\legend{$e_x$, $\beta_x$, BNN, Oracle}
\end{axis}
\end{tikzpicture}
\vspace{-15px}
\caption{Noise detection experiment on KITTI-STEP extending Fig.~4 in the main paper.}
\label{fig:noise_kitti}
\end{figure}
\begin{figure}[t]
\centering
\begin{tikzpicture}
\begin{axis}[
    title=Error Detection Rate ADE20k,
    scaled ticks=false, 
    log ticks with fixed point,
    tick label style={/pgf/number format/fixed},
    xmin = 0, xmax = 1.0,
    ymin = 0.8, ymax = 1.,
    xtick distance = 0.25,
    ytick distance = 0.05,
    grid = both,
    minor tick num = 2,
    major grid style = {lightgray},
    minor grid style = {lightgray!25},
    width = 1\linewidth,
    height = 0.7\linewidth,
    xlabel = {$r_\text{Area}$},
    ylabel = {$\text{Acc}_a$},
    legend style = {
      at={(0.5, -0.4)},
      anchor=north,
      legend columns=5},
]
                \addplot[
                    thick,
                    red,
                ] file[skip first] {detection_ade20k_val_Combined.txt};
                \addplot[
                    black,thick,
                ] file[skip first] {detection_ade20k_val_Alpha.txt};
                \addplot[
                    blue,thick,
                ] file[skip first] {detection_ade20k_val_Bayes.txt};
                \addplot[
                    green,
                    thick,
                ] file[skip first] {detection_ade20k_val_hyperbolic.txt};           \addplot[
                    dashed,
                    black,
                ]{0.8049+x};

\legend{$e_x$, $\beta_x$, BNN, HIS, Oracle}
\end{axis}
\end{tikzpicture}
\vspace{-15px}
\caption{Noise detection experiment on ADE20k extending Fig.~4 in the main paper.}
\label{fig:noise_ade20k}
\end{figure}
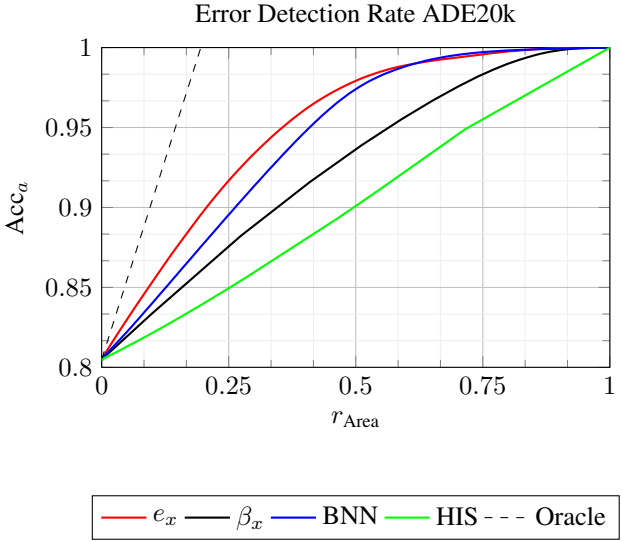
\begin{figure}[t]
\centering
\begin{tikzpicture}
\begin{axis}[
    title=Error Detection Rate COCO,
    scaled ticks=false, 
    log ticks with fixed point,
    tick label style={/pgf/number format/fixed},
    xmin = 0, xmax = 1.0,
    ymin = 0.65, ymax = 1.,
    xtick distance = 0.25,
    ytick distance = 0.1,
    grid = both,
    minor tick num = 2,
    major grid style = {lightgray},
    minor grid style = {lightgray!25},
    width = 1\linewidth,
    height = 0.7\linewidth,
    xlabel = {$r_\text{Area}$},
    ylabel = {$\text{Acc}_a$},
    legend style = {
      at={(0.5, -0.4)},
      anchor=north,
      legend columns=5},
]
                \addplot[
                    thick,
                    red,
                ] file[skip first] {detection_coco10k_val_Combined.txt};
                \addplot[
                    black,thick,
                ] file[skip first] {detection_coco10k_val_Alpha.txt};
                \addplot[
                    blue,thick,
                ] file[skip first] {detection_coco10k_val_Bayes.txt};
                \addplot[
                    green,
                    thick,
                ] file[skip first] {detection_coco10k_val_hyperbolic.txt};
                \addplot[
                    dashed,
                    black,
                ]{0.6626+x};

\legend{$e_x$, $\beta_x$, BNN, HIS, Oracle}
\end{axis}
\end{tikzpicture}
\vspace{-15px}
\caption{Noise detection experiment on COCO extending Fig.~4 in the main paper.}
\label{fig:noise_coco10k}
\end{figure}
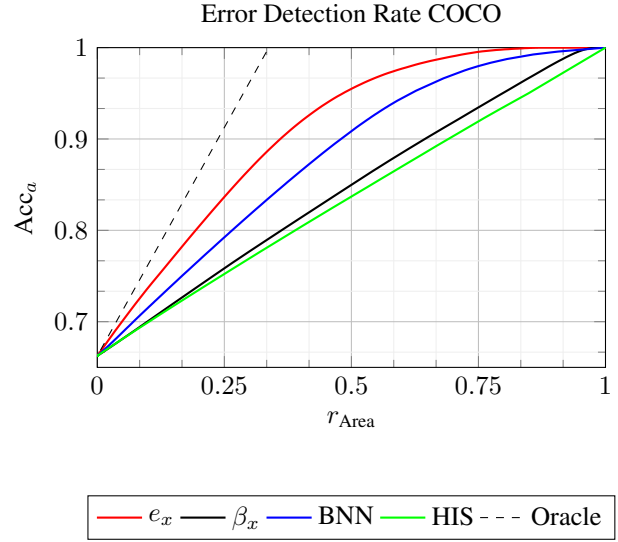
\begin{figure*}
    \centering
    \includegraphics[width=1\linewidth]{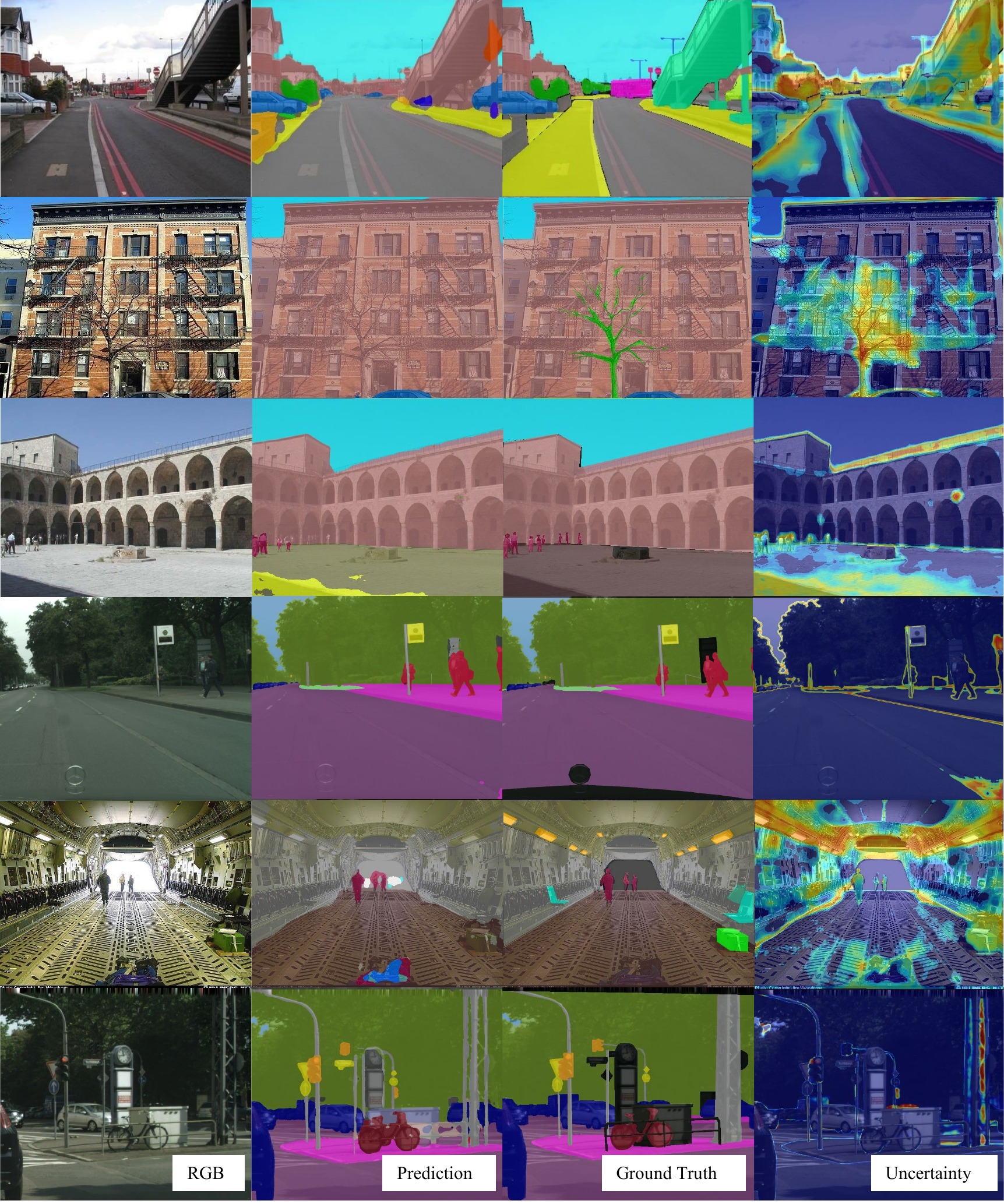}
    \caption{Samples of our uncertainty estimation with \textit{DeepLabv3+}.}
    \label{fig:uncertainty_samples}
\end{figure*}

\begin{figure*}
    \centering
        \begin{subfigure}[b]{0.45\linewidth}
        \centering
        \resizebox{\columnwidth}{!}{
\begin{tikzpicture}
  \begin{axis}[
    title  = KITTI-STEP,
    xbar,
    y axis line style = { opacity = 0 },
    axis x line       = none,
    tickwidth         = 0pt,
    ytick             = data,
    legend style={at={(0.5,0.04)}, anchor=north},
    bar width=6pt,
    height            = 2.8\linewidth,
    width             = \linewidth,
    symbolic y coords = {bicycle, motorcycle, train, bus, truck, car, rider, person, sky, terrain, vegetation, traffic sign, traffic light, pole, fence, wall, building, sidewalk, road, aAcc},
    nodes near coords,
    reverse legend,
  ]

\addplot coordinates {
  (90.97,aAcc)
  (93.98,road)
  (75.49,sidewalk)
  (90.92,building)
  (41.82,wall)
  (38.1,fence)
  (58.77,pole)
  (70.79,traffic light)
  (61.71,traffic sign)
  (96.69,vegetation)
  (88.59,terrain)
  (95.7,sky)
  (78.55,person)
  (50.38,rider)
  (94.33,car)
  (36.02,truck)
  (17.51,bus)
  (1.66,train)
  (57.49,motorcycle)
  (81.96,bicycle)
};

\addplot coordinates {
  (90.93,aAcc)
  (93.95,road)
  (75.22,sidewalk)
  (90.55,building)
  (41.78,wall)
  (38.1,fence)
  (58.62,pole)
  (70.79,traffic light)
  (61.5,traffic sign)
  (96.64,vegetation)
  (88.58,terrain)
  (95.7,sky)
  (90.53,person)
  (28.8,rider)
  (94.25,car)
  (35.82,truck)
  (17.49,bus)
  (1.66,train)
  (51.06,motorcycle)
  (79.56,bicycle)
};

\addplot coordinates {
  (90.94,aAcc)
  (93.94,road)
  (75.22,sidewalk)
  (90.54,building)
  (41.78,wall)
  (38.1,fence)
  (58.62,pole)
  (70.79,traffic light)
  (61.5,traffic sign)
  (96.63,vegetation)
  (88.58,terrain)
  (95.7,sky)
  (89.79,person)
  (60.34,rider)
  (94.24,car)
  (35.82,truck)
  (17.49,bus)
  (1.66,train)
  (50.72,motorcycle)
  (76.76,bicycle)
};

  \legend{Regular , Induced (\textit{person}), Induced (\textit{person}/\textit{rider})}
  \end{axis}
\end{tikzpicture}}
        \caption{Induction Experiments on KITTI-STEP}
        \label{fig:induction_kitti}
        \end{subfigure}
        \hfill
        \begin{subfigure}[b]{0.45\linewidth}
        \centering
        \resizebox{\columnwidth}{!}{
\begin{tikzpicture}
  \begin{axis}[
    title  = Cityscapes,
    xbar,
    y axis line style = { opacity = 0 },
    axis x line       = none,
    tickwidth         = 0pt,
    %enlarge y limits  = 0.0,
    %enlarge y limits  = 1.35,
    ytick             = data,
    legend style={at={(0.5,0.04)}, anchor=north},
    reverse legend,
    bar width=6pt,
    height            = 2.8\linewidth,
    width             = \linewidth,
    symbolic y coords = {bicycle, motorcycle, train, bus, truck, car, rider, person, sky, terrain, vegetation, traffic sign, traffic light, pole, fence, wall, building, sidewalk, road, aAcc},
    nodes near coords,
  ]

\addplot coordinates {
  (96.37,aAcc)
  (99.24,road)
  (92.1,sidewalk)
  (97.5,building)
  (56.42,wall)
  (69.49,fence)
  (76.62,pole)
  (84.23,traffic light)
  (87.49,traffic sign)
  (96.48,vegetation)
  (74.11,terrain)
  (98.21,sky)
  (91.8,person)
  (74.27,rider)
  (97.9,car)
  (88.0,truck)
  (94.26,bus)
  (90.28,train)
  (80.67,motorcycle)
  (89.85,bicycle)
};

\addplot coordinates {
  (96.23,aAcc)
  (99.17,road)
  (91.52,sidewalk)
  (97.14,building)
  (56.2,wall)
  (69.11,fence)
  (75.76,pole)
  (84.23,traffic light)
  (87.35,traffic sign)
  (96.37,vegetation)
  (74.07,terrain)
  (98.21,sky)
  (98.07,person)
  (66.15,rider)
  (97.67,car)
  (87.9,truck)
  (94.16,bus)
  (90.24,train)
  (78.91,motorcycle)
  (87.69,bicycle)
};

\addplot coordinates {
  (96.21,aAcc)
  (99.16,road)
  (91.5,sidewalk)
  (97.12,building)
  (56.19,wall)
  (69.09,fence)
  (75.74,pole)
  (84.22,traffic light)
  (87.34,traffic sign)
  (96.35,vegetation)
  (74.05,terrain)
  (98.21,sky)
  (97.67,person)
  (81.45,rider)
  (97.63,car)
  (87.86,truck)
  (94.09,bus)
  (90.24,train)
  (75.93,motorcycle)
  (84.18,bicycle)
};

  \legend{Regular , Induced (\textit{person}), Induced (\textit{person}/\textit{rider})}
  \end{axis}
\end{tikzpicture}}
        \caption{Induction Experiments on Cityscapes}
        \label{fig:induction_city}
        \end{subfigure}
    \caption{Induction experiments on Cityscapes and KITTI-STEP providing information about all class accuracy metrics to complete Fig.~6 in the main paper. Additionally, the figure presents the metrics, if only one class is induced during inference.}
    \label{fig:induction_long}
\end{figure*}
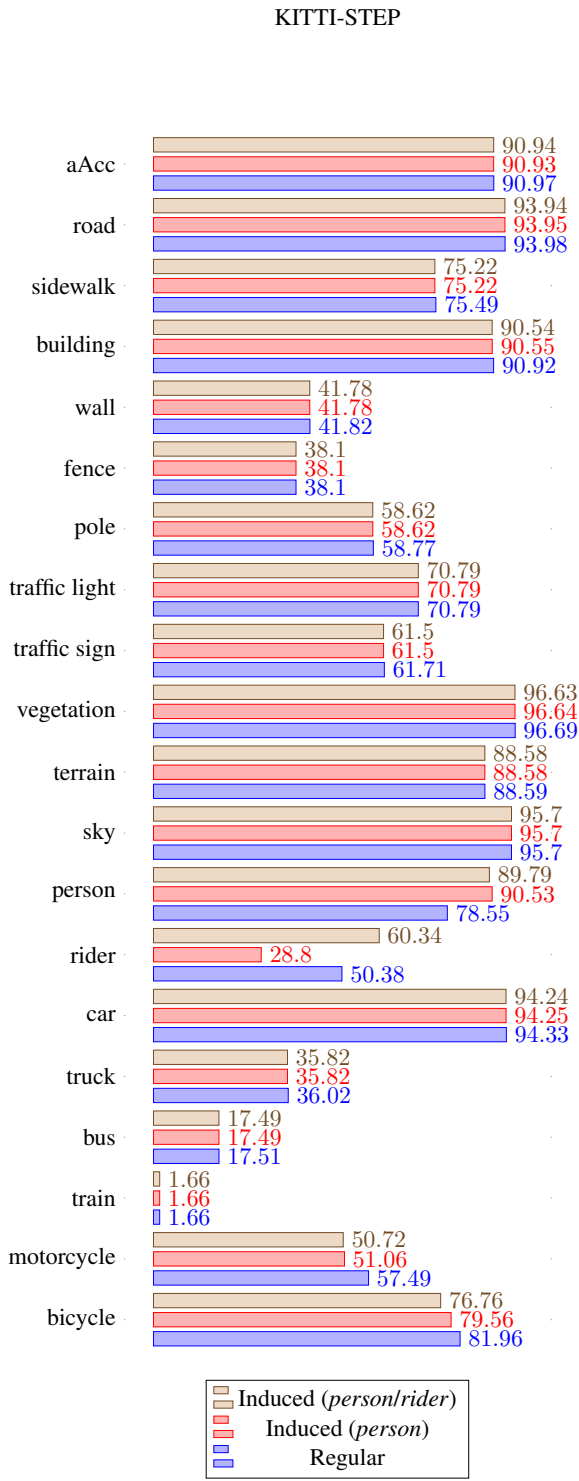
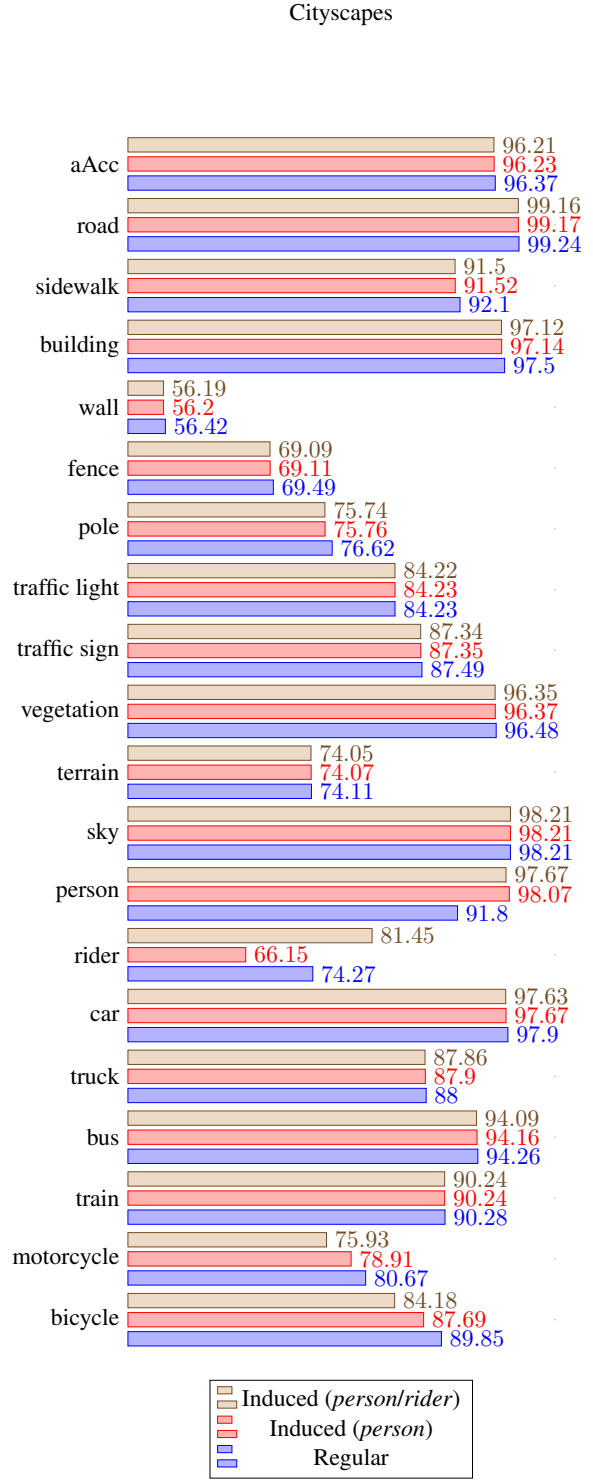

\begin{table*}[t]
    \centering
    \newcommand{\pst}[1]{{\cellcolor{red!\fpeval{100*(#1+0.18)/3.0}}}\scalebox{0.6}{#1}}
    \newcommand{\ngt}[1]{{\cellcolor{blue!\fpeval{100*(#1+0.18)/3.0}}}\scalebox{0.6}{-#1}}
    \newcommand{\zero}[0]{\scalebox{0.4}{0}}
    \newcommand{\labelname}[1]{\scalebox{0.85}{\textit{#1}}}
    \newcolumntype{"}{@{\hskip\tabcolsep\vrule width 1pt \hspace{0.1pt}|}}
    \newcolumntype{x}[0]{>{\centering\arraybackslash\hspace{0pt}}p{0.022\textwidth}}
    \caption{Full compensation weight matrix $B$ for \textit{DeepLabv3} trained on KITTI-STEP.}
    \vspace{-8pt}
    \resizebox{\linewidth}{!}{
    \begin{tabular}{c|x|x|x|x|x|x|x|x|x|x|x|x|x|x|x|x|x|x|x}
    \toprule
    
\multirow{5}{*}{$B_{ij}$} & \multicolumn{11}{c|}{\textit{stuff}} & \multicolumn{8}{c}{\textit{thing}} \\ \cline{2-20}

& \rotatebox{90}{\labelname{road}} 
& \rotatebox{90}{\labelname{sidewalk}} 
& \rotatebox{90}{\labelname{building}} 
& \rotatebox{90}{\labelname{wall}} 
& \rotatebox{90}{\labelname{fence}} 
& \rotatebox{90}{\labelname{pole}} 
& \rotatebox{90}{\labelname{tr. light}}
& \rotatebox{90}{\labelname{tr. sign}}
& \rotatebox{90}{\labelname{vegetation\ }} 
& \rotatebox{90}{\labelname{terrain}} 
& \rotatebox{90}{\labelname{sky}} 
& \rotatebox{90}{\labelname{person}} 
& \rotatebox{90}{\labelname{rider}} 
& \rotatebox{90}{\labelname{car}} 
& \rotatebox{90}{\labelname{truck}}
& \rotatebox{90}{\labelname{bus}}
& \rotatebox{90}{\labelname{train}}
& \rotatebox{90}{\labelname{motorcycle\ }}
& \rotatebox{90}{\labelname{bicycle }} \\\midrule % \hline\hline
    
%%%%% LogComp

\labelname{road} &\zero&\ngt{1.7}&\zero&\zero&\ngt{0.1}&\ngt{0.2}&\zero&\ngt{0.1}&\ngt{0.3}&\ngt{1.1}&\zero&\ngt{0.2}&\zero&\ngt{0.6}&\ngt{0.1}&\ngt{0.1}&\zero&\zero&\ngt{0.1}\\\cline{1-1}

\labelname{sidewalk} &\ngt{1.6}&\zero&\ngt{0.4}&\ngt{0.1}&\ngt{0.1}&\ngt{0.2}&\zero&\zero&\ngt{0.4}&\ngt{0.8}&\zero&\ngt{0.2}&\zero&\ngt{0.3}&\zero&\zero&\zero&\zero&\ngt{0.2}\\\cline{1-1}

\labelname{building} &\zero&\ngt{0.4}&\zero&\ngt{0.1}&\ngt{0.3}&\ngt{0.8}&\ngt{0.1}&\ngt{0.2}&\ngt{1.7}&\ngt{0.1}&\ngt{0.5}&\ngt{0.6}&\zero&\ngt{0.4}&\ngt{0.1}&\zero&\ngt{0.1}&\zero&\ngt{0.2}\\\cline{1-1}

\labelname{wall} &\zero&\zero&\ngt{0.1}&\zero&\ngt{0.2}&\zero&\zero&\zero&\ngt{0.1}&\zero&\zero&\zero&\zero&\zero&\zero&\zero&\zero&\zero&\zero\\\cline{1-1}

\labelname{fence} &\ngt{0.1}&\ngt{0.1}&\ngt{0.2}&\ngt{0.2}&\zero&\ngt{0.1}&\zero&\zero&\ngt{0.7}&\ngt{0.3}&\zero&\zero&\zero&\zero&\zero&\zero&\ngt{0.1}&\zero&\zero\\\cline{1-1}

\labelname{pole} &\ngt{0.1}&\ngt{0.2}&\ngt{0.8}&\zero&\ngt{0.1}&\zero&\ngt{0.1}&\ngt{0.2}&\ngt{1.1}&\ngt{0.2}&\ngt{0.3}&\zero&\zero&\ngt{0.1}&\zero&\zero&\zero&\zero&\ngt{0.1}\\\cline{1-1}

\labelname{traffic light} &\zero&\zero&\zero&\zero&\zero&\ngt{0.1}&\zero&\zero&\ngt{0.1}&\zero&\zero&\zero&\zero&\zero&\zero&\zero&\zero&\zero&\zero\\\cline{1-1}

\labelname{traffic sign} &\zero&\zero&\ngt{0.2}&\zero&\zero&\ngt{0.2}&\zero&\zero&\ngt{0.4}&\zero&\zero&\zero&\zero&\zero&\zero&\zero&\zero&\zero&\zero\\\cline{1-1}

\labelname{vegetation} &\ngt{0.3}&\ngt{0.5}&\ngt{1.8}&\ngt{0.2}&\ngt{0.8}&\ngt{1.2}&\ngt{0.2}&\ngt{0.5}&\zero&\ngt{1.3}&\ngt{1.2}&\ngt{0.2}&\ngt{0.1}&\ngt{0.5}&\ngt{0.1}&\ngt{0.1}&\zero&\zero&\ngt{0.1}\\\cline{1-1}

\labelname{terrain} &\ngt{1.1}&\ngt{0.7}&\zero&\zero&\ngt{0.3}&\ngt{0.2}&\zero&\zero&\ngt{1.2}&\zero&\zero&\zero&\zero&\ngt{0.1}&\zero&\zero&\zero&\zero&\zero\\\cline{1-1}

\labelname{sky} &\zero&\zero&\ngt{0.4}&\zero&\zero&\ngt{0.3}&\zero&\ngt{0.1}&\ngt{1.1}&\zero&\zero&\zero&\zero&\zero&\zero&\ngt{0.1}&\zero&\zero&\zero\\\cline{1-1}

\labelname{person} &\ngt{0.1}&\ngt{0.2}&\ngt{0.5}&\zero&\zero&\zero&\zero&\zero&\ngt{0.1}&\zero&\zero&\zero&\ngt{0.2}&\ngt{0.1}&\zero&\zero&\zero&\zero&\ngt{0.2}\\\cline{1-1}

\labelname{rider} &\zero&\zero&\zero&\zero&\zero&\zero&\zero&\zero&\zero&\zero&\zero&\ngt{0.1}&\zero&\zero&\zero&\zero&\zero&\zero&\ngt{0.1}\\\cline{1-1}

\labelname{car} &\ngt{0.6}&\ngt{0.3}&\ngt{0.4}&\zero&\zero&\ngt{0.1}&\zero&\zero&\ngt{0.4}&\ngt{0.1}&\zero&\ngt{0.1}&\zero&\zero&\ngt{0.2}&\ngt{0.4}&\zero&\zero&\zero\\\cline{1-1}

\labelname{truck} &\zero&\zero&\zero&\zero&\zero&\zero&\zero&\zero&\zero&\zero&\zero&\zero&\zero&\ngt{0.1}&\zero&\ngt{0.7}&\zero&\zero&\zero\\\cline{1-1}

\labelname{bus} &\zero&\zero&\zero&\zero&\zero&\zero&\zero&\zero&\zero&\zero&\ngt{0.1}&\zero&\zero&\ngt{0.3}&\ngt{0.7}&\zero&\zero&\zero&\ngt{0.1}\\\cline{1-1}

\labelname{train} &\zero&\zero&\zero&\zero&\ngt{0.1}&\zero&\zero&\zero&\zero&\zero&\zero&\zero&\zero&\zero&\zero&\zero&\zero&\zero&\zero\\\cline{1-1}

\labelname{motorcycle} &\zero&\zero&\zero&\zero&\zero&\zero&\zero&\zero&\zero&\zero&\zero&\zero&\zero&\zero&\zero&\zero&\zero&\zero&\zero\\\cline{1-1}

\labelname{bicycle} &\zero&\ngt{0.2}&\ngt{0.1}&\zero&\zero&\zero&\zero&\zero&\zero&\zero&\zero&\ngt{0.2}&\ngt{0.1}&\zero&\zero&\ngt{0.1}&\zero&\zero&\zero\\\bottomrule

\end{tabular}}
\label{tab:full_compensation_weights_kitti_deeplab}
\end{table*}

\begin{table*}[t]
    \centering
    \newcommand{\pst}[1]{{\cellcolor{red!\fpeval{100*(#1+0.18)/3.0}}}\scalebox{0.6}{#1}}
    \newcommand{\ngt}[1]{{\cellcolor{blue!\fpeval{100*(#1+0.18)/3.0}}}\scalebox{0.6}{-#1}}
    \newcommand{\labelname}[1]{\scalebox{0.85}{\textit{#1}}}
    \newcommand{\zero}[0]{\scalebox{0.4}{0}}
    \newcolumntype{"}{@{\hskip\tabcolsep\vrule width 1pt \hspace{0.1pt}|}}
    \newcolumntype{x}[0]{>{\centering\arraybackslash\hspace{0pt}}p{0.022\textwidth}}
    \caption{Full compensation weight matrix $B$ for \textit{DeepLabv3} trained on Cityscapes.}
    \vspace{-8pt}
    \resizebox{\linewidth}{!}{
    \begin{tabular}{c|x|x|x|x|x|x|x|x|x|x|x|x|x|x|x|x|x|x|x}
    \toprule
    
\multirow{5}{*}{$B_{ij}$} & \multicolumn{11}{c|}{\textit{stuff}} & \multicolumn{8}{c}{\textit{thing}} \\ \cline{2-20}
    
& \rotatebox{90}{\labelname{road}} 
& \rotatebox{90}{\labelname{sidewalk}} 
& \rotatebox{90}{\labelname{building}} 
& \rotatebox{90}{\labelname{wall}} 
& \rotatebox{90}{\labelname{fence}} 
& \rotatebox{90}{\labelname{pole}} 
& \rotatebox{90}{\labelname{tr. light}}
& \rotatebox{90}{\labelname{tr. sign}}
& \rotatebox{90}{\labelname{vegetation\ }} 
& \rotatebox{90}{\labelname{terrain}} 
& \rotatebox{90}{\labelname{sky}} 
& \rotatebox{90}{\labelname{person}} 
& \rotatebox{90}{\labelname{rider}} 
& \rotatebox{90}{\labelname{car}} 
& \rotatebox{90}{\labelname{truck}}
& \rotatebox{90}{\labelname{bus}}
& \rotatebox{90}{\labelname{train}}
& \rotatebox{90}{\labelname{motorcycle\ }}
& \rotatebox{90}{\labelname{bicycle }} \\\midrule % \hline\hline
    
\labelname{road} &\zero&\ngt{2.1}&\ngt{0.1}&\ngt{0.1}&\ngt{0.1}&\ngt{0.1}&\zero&\zero&\ngt{0.2}&\ngt{0.6}&\zero&\ngt{0.3}&\zero&\ngt{1.0}&\ngt{0.1}&\ngt{0.1}&\zero&\zero&\ngt{0.1}\\\cline{1-1}

\labelname{sidewalk} &\ngt{1.9}&\zero&\ngt{0.8}&\ngt{0.3}&\ngt{0.3}&\ngt{0.5}&\zero&\zero&\ngt{0.4}&\ngt{1.1}&\zero&\ngt{0.4}&\zero&\ngt{0.4}&\zero&\zero&\zero&\zero&\ngt{0.2}\\\cline{1-1}

\labelname{building} &\ngt{0.1}&\ngt{0.8}&\zero&\ngt{0.9}&\ngt{0.9}&\ngt{1.9}&\ngt{0.4}&\ngt{0.8}&\ngt{2.4}&\ngt{0.2}&\ngt{0.8}&\ngt{0.9}&\ngt{0.1}&\ngt{0.9}&\ngt{0.3}&\ngt{0.2}&\ngt{0.2}&\ngt{0.1}&\ngt{0.3}\\\cline{1-1}

\labelname{wall} &\zero&\ngt{0.3}&\ngt{0.8}&\zero&\ngt{0.6}&\ngt{0.1}&\zero&\zero&\ngt{0.5}&\ngt{0.1}&\zero&\ngt{0.1}&\zero&\ngt{0.1}&\zero&\zero&\zero&\zero&\zero\\\cline{1-1}

\labelname{fence} &\zero&\ngt{0.2}&\ngt{0.7}&\ngt{0.6}&\zero&\ngt{0.3}&\zero&\zero&\ngt{0.6}&\ngt{0.2}&\zero&\ngt{0.1}&\zero&\ngt{0.1}&\zero&\zero&\zero&\zero&\ngt{0.1}\\\cline{1-1}

\labelname{pole} &\zero&\ngt{0.4}&\ngt{1.7}&\ngt{0.1}&\ngt{0.3}&\zero&\ngt{0.2}&\ngt{0.3}&\ngt{1.3}&\ngt{0.2}&\ngt{0.2}&\ngt{0.2}&\zero&\ngt{0.3}&\zero&\zero&\zero&\zero&\ngt{0.2}\\\cline{1-1}

\labelname{traffic light} &\zero&\zero&\ngt{0.3}&\zero&\zero&\ngt{0.2}&\zero&\zero&\ngt{0.2}&\zero&\zero&\zero&\zero&\zero&\zero&\zero&\zero&\zero&\zero\\\cline{1-1}

\labelname{traffic sign} &\zero&\zero&\ngt{0.7}&\zero&\zero&\ngt{0.3}&\ngt{0.1}&\zero&\ngt{0.3}&\zero&\zero&\zero&\zero&\zero&\zero&\zero&\zero&\zero&\zero\\\cline{1-1}

\labelname{vegetation} &\ngt{0.1}&\ngt{0.5}&\ngt{2.4}&\ngt{0.6}&\ngt{0.7}&\ngt{1.4}&\ngt{0.3}&\ngt{0.4}&\zero&\ngt{1.1}&\ngt{0.8}&\ngt{0.4}&\ngt{0.1}&\ngt{0.7}&\ngt{0.1}&\ngt{0.1}&\zero&\zero&\ngt{0.1}\\\cline{1-1}

\labelname{terrain} &\ngt{0.4}&\ngt{1.0}&\ngt{0.1}&\ngt{0.1}&\ngt{0.2}&\ngt{0.2}&\zero&\zero&\ngt{1.0}&\zero&\zero&\zero&\zero&\ngt{0.1}&\zero&\zero&\zero&\zero&\ngt{0.1}\\\cline{1-1}

\labelname{sky} &\zero&\zero&\ngt{0.7}&\zero&\zero&\ngt{0.2}&\zero&\zero&\ngt{0.8}&\zero&\zero&\zero&\zero&\zero&\zero&\zero&\zero&\zero&\zero\\\cline{1-1}

\labelname{person} &\ngt{0.2}&\ngt{0.4}&\ngt{0.8}&\ngt{0.1}&\ngt{0.1}&\ngt{0.2}&\zero&\zero&\ngt{0.3}&\zero&\zero&\zero&\ngt{0.4}&\ngt{0.3}&\zero&\zero&\zero&\ngt{0.1}&\ngt{0.2}\\\cline{1-1}

\labelname{rider} &\zero&\zero&\zero&\zero&\zero&\zero&\zero&\zero&\zero&\zero&\zero&\ngt{0.3}&\zero&\zero&\zero&\zero&\zero&\ngt{0.1}&\ngt{0.3}\\\cline{1-1}

\labelname{car} &\ngt{0.9}&\ngt{0.4}&\ngt{0.8}&\ngt{0.1}&\ngt{0.1}&\ngt{0.3}&\zero&\ngt{0.1}&\ngt{0.6}&\ngt{0.1}&\zero&\ngt{0.4}&\ngt{0.1}&\zero&\ngt{0.3}&\ngt{0.2}&\ngt{0.1}&\ngt{0.1}&\ngt{0.2}\\\cline{1-1}

\labelname{truck} &\zero&\zero&\ngt{0.2}&\zero&\zero&\zero&\zero&\zero&\zero&\zero&\zero&\zero&\zero&\ngt{0.3}&\zero&\ngt{0.1}&\zero&\zero&\zero\\\cline{1-1}

\labelname{bus} &\zero&\zero&\zero&\zero&\zero&\zero&\zero&\zero&\zero&\zero&\zero&\zero&\zero&\ngt{0.1}&\ngt{0.1}&\zero&\ngt{0.1}&\zero&\zero\\\cline{1-1}

\labelname{train} &\zero&\zero&\ngt{0.1}&\zero&\zero&\zero&\zero&\zero&\zero&\zero&\zero&\zero&\zero&\zero&\zero&\ngt{0.1}&\zero&\zero&\zero\\\cline{1-1}

\labelname{motorcycle} &\zero&\zero&\zero&\zero&\zero&\zero&\zero&\zero&\zero&\zero&\zero&\zero&\ngt{0.1}&\ngt{0.1}&\zero&\zero&\zero&\zero&\ngt{0.1}\\\cline{1-1}

\labelname{bicycle} &\zero&\ngt{0.2}&\ngt{0.2}&\zero&\ngt{0.1}&\ngt{0.2}&\zero&\zero&\ngt{0.1}&\zero&\zero&\ngt{0.2}&\ngt{0.3}&\ngt{0.2}&\zero&\zero&\zero&\ngt{0.1}&\zero\\\bottomrule

\end{tabular}}
\label{tab:full_compensation_weights_cityscapes_deeplab}
\end{table*}

\begin{table*}[t]
    \centering
    \newcommand{\pst}[1]{{\cellcolor{red!\fpeval{100*(#1+0.5)/9.0}}}\scalebox{0.6}{#1}}
    \newcommand{\ngt}[1]{{\cellcolor{blue!\fpeval{100*(#1+0.5)/9.0}}}\scalebox{0.6}{-#1}}
    \newcommand{\zero}[0]{\scalebox{0.4}{0}}
    \newcommand{\labelname}[1]{\scalebox{0.85}{\textit{#1}}}
    \newcolumntype{"}{@{\hskip\tabcolsep\vrule width 1pt \hspace{0.1pt}|}}
    \newcolumntype{x}[0]{>{\centering\arraybackslash\hspace{0pt}}p{0.022\textwidth}}
    \caption{Full compensation weight matrix $B$ for \textit{SegFormer} trained on KITTI-STEP.}
    \vspace{-8pt}
    \resizebox{\linewidth}{!}{
    \begin{tabular}{c|x|x|x|x|x|x|x|x|x|x|x|x|x|x|x|x|x|x|x}
    \toprule
    
\multirow{5}{*}{$B_{ij}$} & \multicolumn{11}{c|}{\textit{stuff}} & \multicolumn{8}{c}{\textit{thing}} \\ \cline{2-20}

& \rotatebox{90}{\labelname{road}} 
& \rotatebox{90}{\labelname{sidewalk}} 
& \rotatebox{90}{\labelname{building}} 
& \rotatebox{90}{\labelname{wall}} 
& \rotatebox{90}{\labelname{fence}} 
& \rotatebox{90}{\labelname{pole}} 
& \rotatebox{90}{\labelname{tr. light}}
& \rotatebox{90}{\labelname{tr. sign}}
& \rotatebox{90}{\labelname{vegetation\ }} 
& \rotatebox{90}{\labelname{terrain}} 
& \rotatebox{90}{\labelname{sky}} 
& \rotatebox{90}{\labelname{person}} 
& \rotatebox{90}{\labelname{rider}} 
& \rotatebox{90}{\labelname{car}} 
& \rotatebox{90}{\labelname{truck}}
& \rotatebox{90}{\labelname{bus}}
& \rotatebox{90}{\labelname{train}}
& \rotatebox{90}{\labelname{motorcycle\ }}
& \rotatebox{90}{\labelname{bicycle }} \\\midrule % \hline\hline
    
%%%%% LogComp

\labelname{road} &\zero&\ngt{7.5}&\zero&\zero&\zero&\zero&\zero&\zero&\zero&\zero&\zero&\zero&\zero&\zero&\zero&\zero&\zero&\zero&\zero\\\cline{1-1}

\labelname{sidewalk} &\zero&\zero&\zero&\zero&\zero&\zero&\zero&\zero&\zero&\zero&\zero&\zero&\zero&\zero&\zero&\zero&\zero&\zero&\zero\\\cline{1-1}

\labelname{building} &\zero&\zero&\zero&\ngt{1.1}&\zero&\ngt{3.3}&\zero&\zero&\zero&\zero&\zero&\zero&\zero&\zero&\zero&\zero&\ngt{3.7}&\ngt{2.2}&\zero\\\cline{1-1}

\labelname{wall} &\zero&\zero&\zero&\zero&\zero&\zero&\zero&\zero&\zero&\zero&\zero&\zero&\zero&\zero&\zero&\zero&\zero&\zero&\zero\\\cline{1-1}

\labelname{fence} &\zero&\zero&\zero&\ngt{0.6}&\zero&\zero&\zero&\zero&\zero&\zero&\zero&\zero&\zero&\zero&\zero&\zero&\zero&\zero&\zero\\\cline{1-1}

\labelname{pole} &\zero&\zero&\zero&\zero&\zero&\zero&\zero&\zero&\zero&\zero&\zero&\zero&\zero&\zero&\zero&\zero&\zero&\zero&\zero\\\cline{1-1}

\labelname{traffic light} &\zero&\zero&\zero&\zero&\zero&\zero&\zero&\zero&\zero&\zero&\zero&\zero&\zero&\zero&\zero&\zero&\zero&\zero&\zero\\\cline{1-1}

\labelname{traffic sign} &\zero&\zero&\zero&\zero&\zero&\zero&\zero&\zero&\zero&\zero&\zero&\zero&\zero&\zero&\zero&\zero&\zero&\zero&\zero\\\cline{1-1}

\labelname{vegetation} &\zero&\zero&\zero&\ngt{4.0}&\ngt{5.1}&\ngt{6.2}&\ngt{5.9}&\ngt{5.3}&\zero&\ngt{6.2}&\zero&\zero&\zero&\zero&\zero&\zero&\zero&\ngt{0.6}&\zero\\\cline{1-1}

\labelname{terrain} &\zero&\zero&\zero&\zero&\zero&\zero&\zero&\zero&\zero&\zero&\zero&\zero&\zero&\zero&\zero&\zero&\zero&\zero&\zero\\\cline{1-1}

\labelname{sky} &\zero&\zero&\zero&\zero&\zero&\zero&\zero&\zero&\zero&\zero&\zero&\zero&\zero&\zero&\zero&\zero&\zero&\zero&\zero\\\cline{1-1}

\labelname{person} &\zero&\zero&\zero&\zero&\zero&\zero&\zero&\zero&\zero&\zero&\zero&\zero&\ngt{2.0}&\zero&\zero&\zero&\zero&\zero&\zero\\\cline{1-1}

\labelname{rider} &\zero&\zero&\zero&\zero&\zero&\zero&\zero&\zero&\zero&\zero&\zero&\zero&\zero&\zero&\zero&\zero&\zero&\zero&\zero\\\cline{1-1}

\labelname{car} &\zero&\zero&\zero&\zero&\zero&\zero&\zero&\zero&\zero&\zero&\zero&\zero&\zero&\zero&\zero&\zero&\zero&\zero&\zero\\\cline{1-1}

\labelname{truck} &\zero&\zero&\zero&\zero&\zero&\zero&\zero&\zero&\zero&\zero&\zero&\zero&\zero&\zero&\zero&\ngt{1.6}&\zero&\zero&\zero\\\cline{1-1}

\labelname{bus} &\zero&\zero&\zero&\zero&\zero&\zero&\zero&\zero&\zero&\zero&\zero&\zero&\zero&\zero&\ngt{3.4}&\zero&\zero&\zero&\zero\\\cline{1-1}

\labelname{train} &\zero&\zero&\zero&\zero&\zero&\zero&\zero&\zero&\zero&\zero&\zero&\zero&\zero&\zero&\zero&\zero&\zero&\zero&\zero\\\cline{1-1}

\labelname{motorcycle} &\zero&\zero&\zero&\zero&\zero&\zero&\zero&\zero&\zero&\zero&\zero&\zero&\zero&\zero&\zero&\zero&\zero&\zero&\zero\\\cline{1-1}

\labelname{bicycle} &\zero&\zero&\zero&\zero&\zero&\zero&\zero&\zero&\zero&\zero&\zero&\zero&\ngt{4.2}&\zero&\zero&\zero&\zero&\zero&\zero\\\bottomrule

\end{tabular}}
\label{tab:full_compensation_weights_kitti_segformer}
\end{table*}

\begin{table*}[t]
    \centering
    \newcommand{\pst}[1]{{\cellcolor{red!\fpeval{100*(#1+0.5)/9.0}}}\scalebox{0.6}{#1}}
    \newcommand{\ngt}[1]{{\cellcolor{blue!\fpeval{100*(#1+0.5)/9.0}}}\scalebox{0.6}{-#1}}
    \newcommand{\zero}[0]{\scalebox{0.4}{0}}
    \newcommand{\labelname}[1]{\scalebox{0.85}{\textit{#1}}}
    \newcolumntype{"}{@{\hskip\tabcolsep\vrule width 1pt \hspace{0.1pt}|}}
    \newcolumntype{x}[0]{>{\centering\arraybackslash\hspace{0pt}}p{0.022\textwidth}}
\caption{Full compensation weight matrix $B$ for \textit{SegFormer} trained on Cityscapes.}
    \vspace{-8pt}
    \resizebox{\linewidth}{!}{
    \begin{tabular}{c|x|x|x|x|x|x|x|x|x|x|x|x|x|x|x|x|x|x|x}
    \toprule
    
\multirow{5}{*}{$B_{ij}$} & \multicolumn{11}{c|}{\textit{stuff}} & \multicolumn{8}{c}{\textit{thing}} \\ \cline{2-20}
    
& \rotatebox{90}{\labelname{road}} 
& \rotatebox{90}{\labelname{sidewalk}} 
& \rotatebox{90}{\labelname{building}} 
& \rotatebox{90}{\labelname{wall}} 
& \rotatebox{90}{\labelname{fence}} 
& \rotatebox{90}{\labelname{pole}} 
& \rotatebox{90}{\labelname{tr. light}}
& \rotatebox{90}{\labelname{tr. sign}}
& \rotatebox{90}{\labelname{vegetation\ }} 
& \rotatebox{90}{\labelname{terrain}} 
& \rotatebox{90}{\labelname{sky}} 
& \rotatebox{90}{\labelname{person}} 
& \rotatebox{90}{\labelname{rider}} 
& \rotatebox{90}{\labelname{car}} 
& \rotatebox{90}{\labelname{truck}}
& \rotatebox{90}{\labelname{bus}}
& \rotatebox{90}{\labelname{train}}
& \rotatebox{90}{\labelname{motorcycle\ }}
& \rotatebox{90}{\labelname{bicycle }} \\\midrule % \hline\hline

\labelname{road} &\zero&\ngt{7.1}&\zero&\zero&\zero&\zero&\zero&\zero&\zero&\ngt{1.0}&\zero&\zero&\zero&\zero&\zero&\zero&\zero&\zero&\zero\\\cline{1-1}

\labelname{sidewalk} &\zero&\zero&\zero&\zero&\zero&\zero&\zero&\zero&\zero&\ngt{0.2}&\zero&\zero&\zero&\zero&\zero&\zero&\zero&\zero&\zero\\\cline{1-1}

\labelname{building} &\zero&\zero&\zero&\ngt{3.3}&\ngt{4.2}&\ngt{7.3}&\ngt{7.5}&\ngt{6.8}&\zero&\zero&\zero&\ngt{5.7}&\zero&\zero&\ngt{2.0}&\ngt{2.2}&\ngt{2.9}&\ngt{3.3}&\zero\\\cline{1-1}

\labelname{wall} &\zero&\zero&\zero&\zero&\zero&\zero&\zero&\zero&\zero&\zero&\zero&\zero&\zero&\zero&\zero&\zero&\zero&\zero&\zero\\\cline{1-1}

\labelname{fence} &\zero&\zero&\zero&\zero&\zero&\zero&\zero&\zero&\zero&\zero&\zero&\zero&\zero&\zero&\zero&\zero&\zero&\zero&\zero\\\cline{1-1}

\labelname{pole} &\zero&\zero&\zero&\zero&\zero&\zero&\ngt{2.6}&\zero&\zero&\zero&\zero&\zero&\zero&\zero&\zero&\zero&\zero&\zero&\zero\\\cline{1-1}

\labelname{traffic light} &\zero&\zero&\zero&\zero&\zero&\zero&\zero&\zero&\zero&\zero&\zero&\zero&\zero&\zero&\zero&\zero&\zero&\zero&\zero\\\cline{1-1}

\labelname{traffic sign} &\zero&\zero&\zero&\zero&\zero&\zero&\zero&\zero&\zero&\zero&\zero&\zero&\zero&\zero&\zero&\zero&\zero&\zero&\zero\\\cline{1-1}

\labelname{vegetation} &\zero&\zero&\zero&\ngt{4.0}&\ngt{5.1}&\ngt{4.8}&\ngt{5.5}&\ngt{2.5}&\zero&\ngt{5.7}&\ngt{7.9}&\zero&\zero&\zero&\zero&\zero&\zero&\zero&\zero\\\cline{1-1}

\labelname{terrain} &\zero&\zero&\zero&\zero&\zero&\zero&\zero&\zero&\zero&\zero&\zero&\zero&\zero&\zero&\zero&\zero&\zero&\zero&\zero\\\cline{1-1}

\labelname{sky} &\zero&\zero&\zero&\zero&\zero&\zero&\zero&\zero&\zero&\zero&\zero&\zero&\zero&\zero&\zero&\zero&\zero&\zero&\zero\\\cline{1-1}

\labelname{person} &\zero&\zero&\zero&\zero&\zero&\zero&\zero&\zero&\zero&\zero&\zero&\zero&\zero&\zero&\zero&\zero&\zero&\zero&\zero\\\cline{1-1}

\labelname{rider} &\zero&\zero&\zero&\zero&\zero&\zero&\zero&\zero&\zero&\zero&\zero&\zero&\zero&\zero&\zero&\zero&\zero&\zero&\zero\\\cline{1-1}

\labelname{car} &\zero&\zero&\zero&\zero&\zero&\zero&\zero&\zero&\zero&\zero&\zero&\zero&\zero&\zero&\ngt{2.5}&\zero&\zero&\ngt{0.4}&\zero\\\cline{1-1}

\labelname{truck} &\zero&\zero&\zero&\zero&\zero&\zero&\zero&\zero&\zero&\zero&\zero&\zero&\zero&\zero&\zero&\zero&\zero&\zero&\zero\\\cline{1-1}

\labelname{bus} &\zero&\zero&\zero&\zero&\zero&\zero&\zero&\zero&\zero&\zero&\zero&\zero&\zero&\zero&\zero&\zero&\zero&\zero&\zero\\\cline{1-1}

\labelname{train} &\zero&\zero&\zero&\zero&\zero&\zero&\zero&\zero&\zero&\zero&\zero&\zero&\zero&\zero&\zero&\zero&\zero&\zero&\zero\\\cline{1-1}

\labelname{motorcycle} &\zero&\zero&\zero&\zero&\zero&\zero&\zero&\zero&\zero&\zero&\zero&\zero&\zero&\zero&\zero&\zero&\zero&\zero&\zero\\\cline{1-1}

\labelname{bicycle} &\zero&\zero&\zero&\zero&\zero&\zero&\zero&\zero&\zero&\zero&\zero&\zero&\ngt{4.6}&\zero&\zero&\zero&\zero&\ngt{1.7}&\zero\\\bottomrule

\end{tabular}}
\label{tab:full_compensation_weights_cityscapes_segformer}
\end{table*}

\begin{figure*}
    \begin{subfigure}[b]{0.5\linewidth}
         \centering
         \includegraphics[width=1\linewidth]{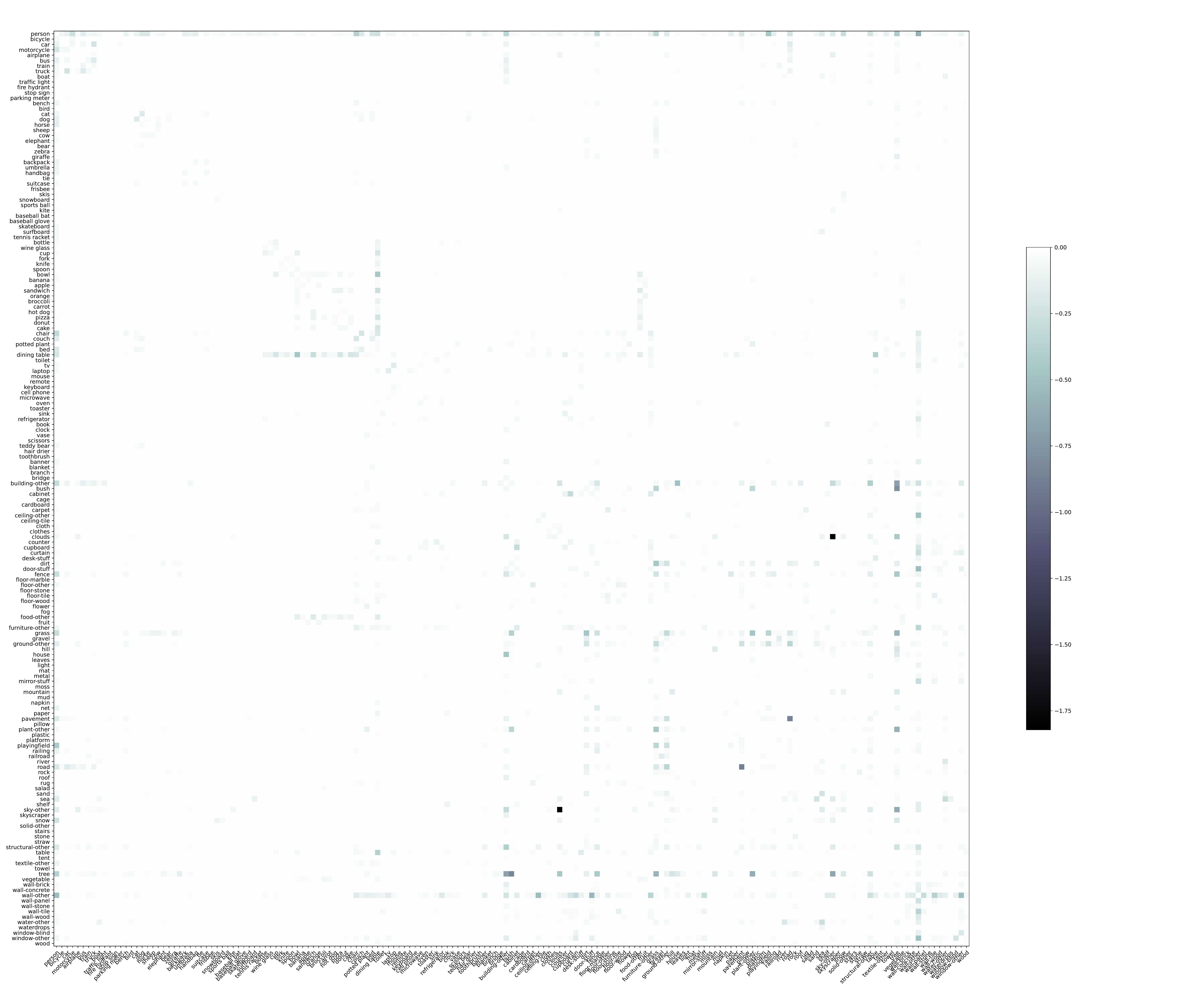}
         \caption{COCO-stuff10k with \textit{DeepLabv3+}.}
         \label{fig:compensations_deeplab_coco}
    \end{subfigure}
    \hfill
    \begin{subfigure}[b]{0.5\linewidth}
         \centering
         \includegraphics[width=1\linewidth]{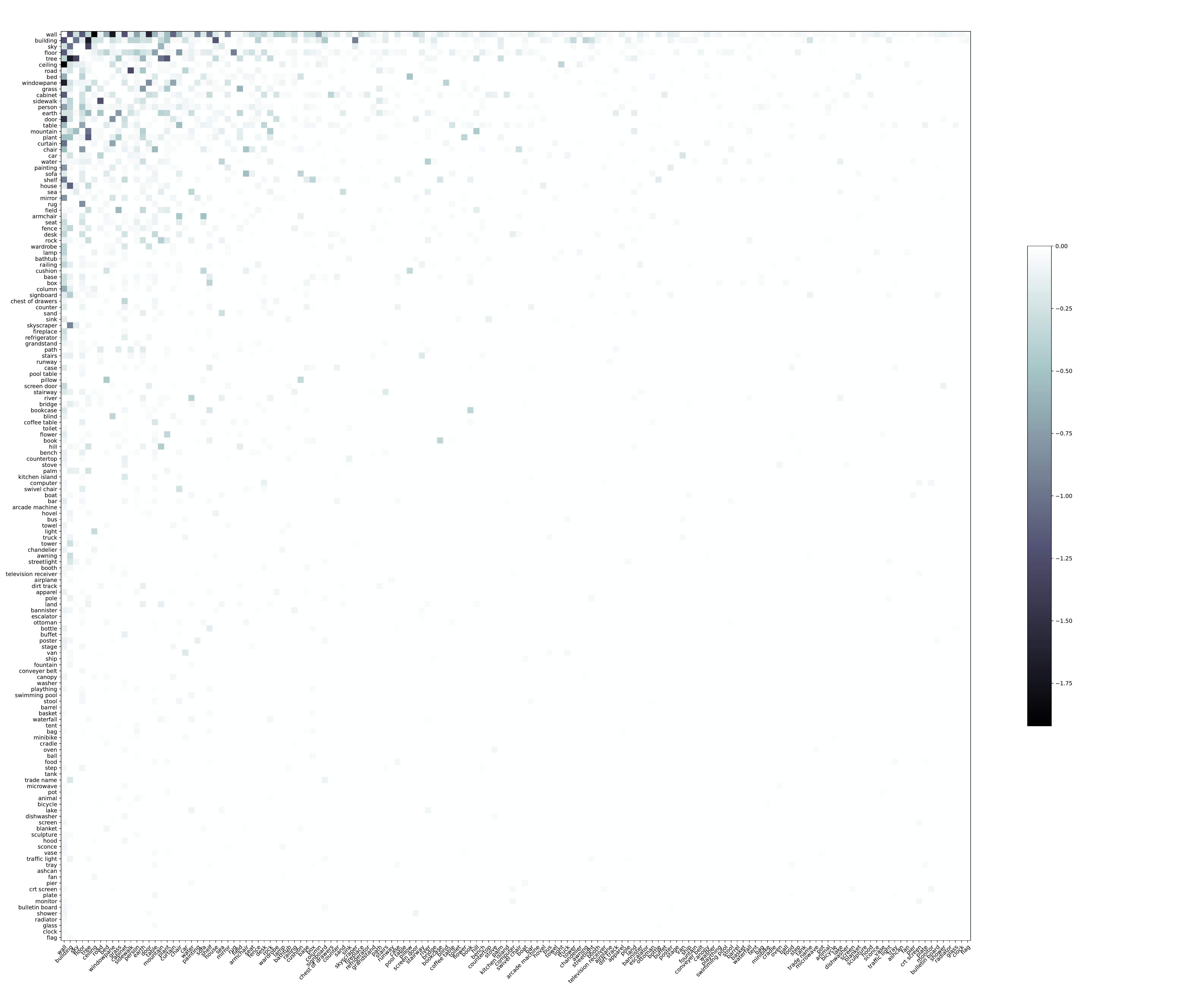}
         \caption{ADE20k with \textit{DeepLabv3+}.}
         \label{fig:compensations_deeplab_ade20k}
    \end{subfigure}
    \hfill
    \vskip\baselineskip
    \begin{subfigure}[b]{0.5\linewidth}
         \centering
         \includegraphics[width=1\linewidth]{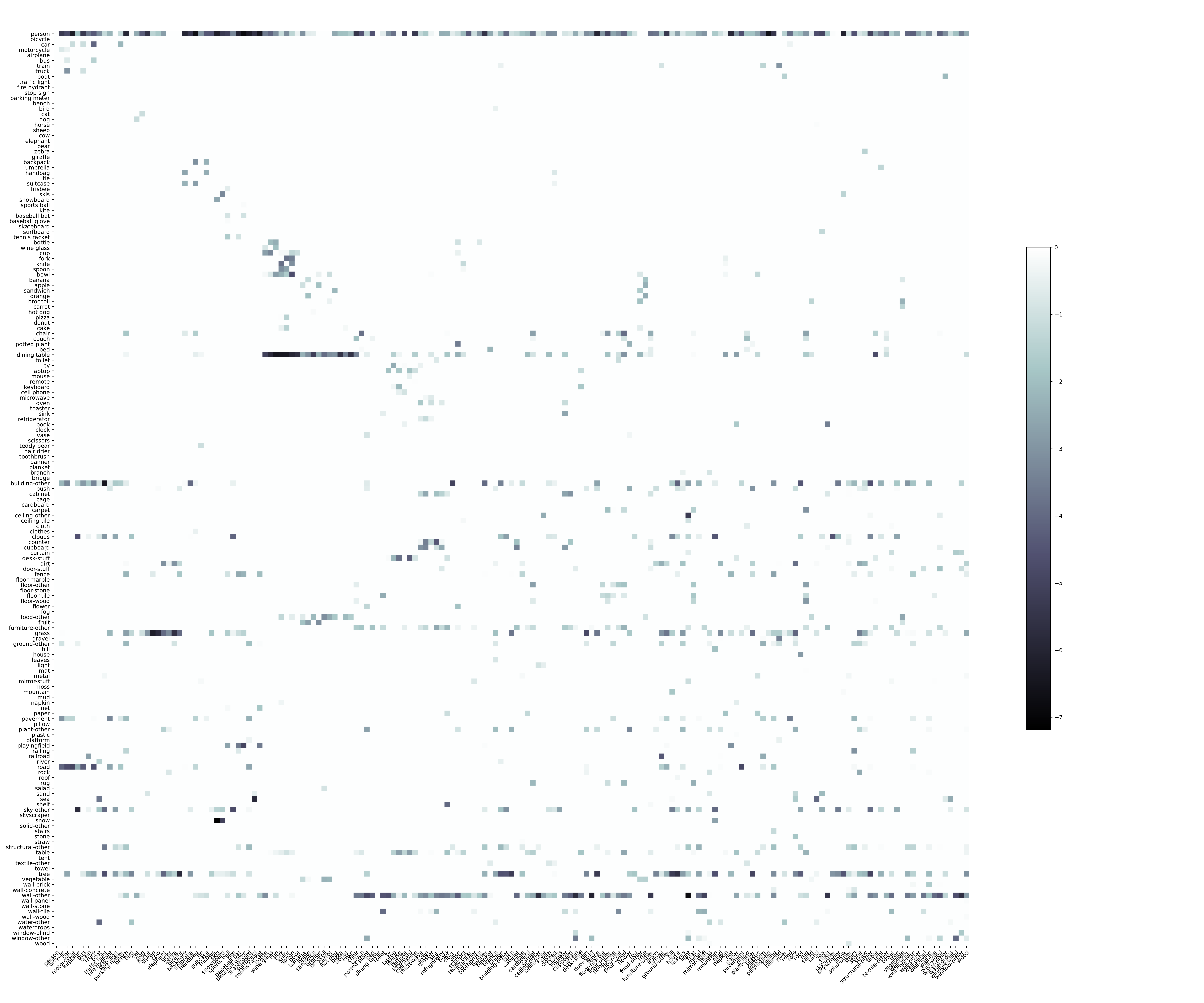}
         \caption{COCO-stuff10k with \textit{SegFormer}.}
         \label{fig:compensations_segformer_coco}
    \end{subfigure}
    \begin{subfigure}[b]{0.5\linewidth}
         \centering
         \includegraphics[width=1\linewidth]{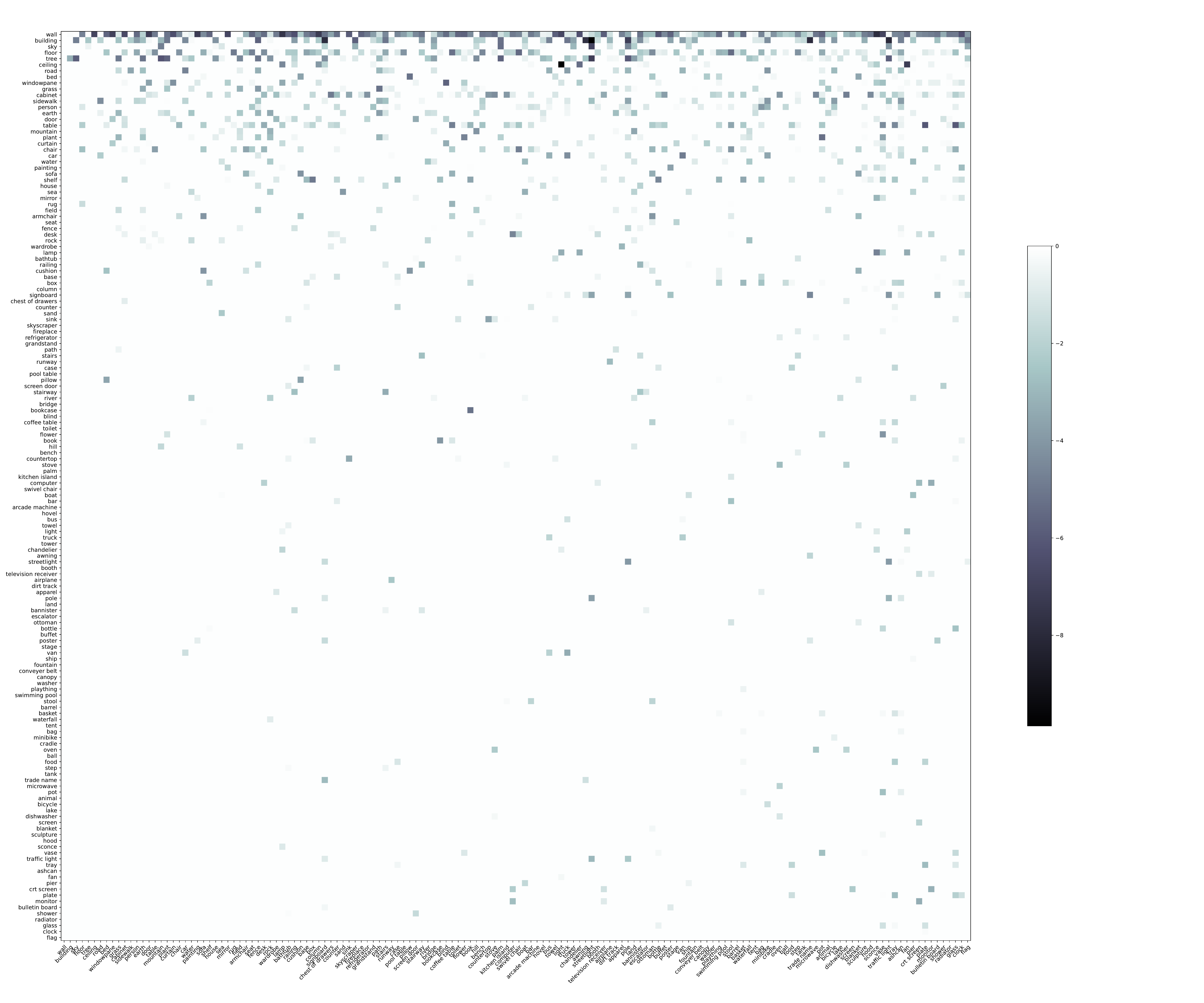}
         \caption{ADE20k with \textit{SegFormer}.}
         \label{fig:compensations_segformer_ade20k}
    \end{subfigure}
     
    \caption{Complete compensation weight matrices $B$ for the large-scale datasets ADE20k and COCO-stuff10k for our baseline methods \textit{SegFormer} and \textit{DeepLabv3+}. We recommend to view the plots digital in the PDF version and zooming in to discover details such as class names and value ranges.}
    \label{fig:large_scale_compensation}
\end{figure*}

\begin{table*}[t]
    \centering
    \newcommand{\pst}[1]{{\cellcolor{red!\fpeval{(#1 + 500*#1/(50+#1*#1))/1.3}}}\scalebox{0.6}{#1}}
    \newcommand{\ngt}[1]{{\cellcolor{blue!\fpeval{100*(#1+0.2)/1.8}}}\scalebox{0.6}{-#1}}
    \newcommand{\labelname}[1]{\scalebox{0.85}{\textit{#1}}}
    \newcommand{\zero}[0]{\scalebox{0.6}{$\approx0$}}
    \newcolumntype{"}{@{\hskip\tabcolsep\vrule width 1pt \hspace{0.1pt}|}}
    \newcolumntype{x}[0]{>{\centering\arraybackslash\hspace{0pt}}p{0.022\textwidth}}
    \caption{Confusion Matrix on the training set of Cityscapes after training with \textit{DeepLabv3+}.}
    \vspace{-8pt}
    \resizebox{\linewidth}{!}{
    \begin{tabular}{c|x|x|x|x|x|x|x|x|x|x|x|x|x|x|x|x|x|x|x}
    \toprule
    
\multirow{5}{*}{$\%$} & \multicolumn{11}{c|}{\textit{stuff}} & \multicolumn{8}{c}{\textit{thing}} \\ \cline{2-20}

& \rotatebox{90}{\labelname{road}} 
& \rotatebox{90}{\labelname{sidewalk}} 
& \rotatebox{90}{\labelname{building}} 
& \rotatebox{90}{\labelname{wall}} 
& \rotatebox{90}{\labelname{fence}} 
& \rotatebox{90}{\labelname{pole}} 
& \rotatebox{90}{\labelname{tr. light}}
& \rotatebox{90}{\labelname{tr. sign}}
& \rotatebox{90}{\labelname{vegetation\ }} 
& \rotatebox{90}{\labelname{terrain}} 
& \rotatebox{90}{\labelname{sky}} 
& \rotatebox{90}{\labelname{person}} 
& \rotatebox{90}{\labelname{rider}} 
& \rotatebox{90}{\labelname{car}} 
& \rotatebox{90}{\labelname{truck}}
& \rotatebox{90}{\labelname{bus}}
& \rotatebox{90}{\labelname{train}}
& \rotatebox{90}{\labelname{motorcycle\ }}
& \rotatebox{90}{\labelname{bicycle }} \\\midrule % \hline\hline
    
%%%%% LogComp
\labelname{road} &\pst{99.4}&\pst{0.3}&\pst{0.0}&\pst{0.0}&\pst{0.0}&\pst{0.0}&\pst{0.0}&\pst{0.0}&\pst{0.0}&\pst{0.0}&\pst{0.0}&\pst{0.0}&\pst{0.0}&\pst{0.1}&\pst{0.0}&\pst{0.0}&\pst{0.0}&\pst{0.0}&\pst{0.0}\\\cline{1-1}

\labelname{sidewalk} &\pst{1.7}&\pst{95.8}&\pst{0.5}&\pst{0.1}&\pst{0.1}&\pst{0.2}&\pst{0.0}&\pst{0.0}&\pst{0.3}&\pst{0.6}&\pst{0.0}&\pst{0.2}&\pst{0.0}&\pst{0.2}&\pst{0.0}&\pst{0.0}&\pst{0.0}&\pst{0.0}&\pst{0.1}\\\cline{1-1}

\labelname{building} &\pst{0.0}&\pst{0.1}&\pst{97.6}&\pst{0.0}&\pst{0.0}&\pst{0.3}&\pst{0.0}&\pst{0.0}&\pst{1.2}&\pst{0.0}&\pst{0.2}&\pst{0.1}&\pst{0.0}&\pst{0.0}&\pst{0.0}&\pst{0.0}&\pst{0.0}&\pst{0.0}&\pst{0.0}\\\cline{1-1}

\labelname{wall} &\pst{0.1}&\pst{0.8}&\pst{3.5}&\pst{89.8}&\pst{2.0}&\pst{0.4}&\pst{0.0}&\pst{0.0}&\pst{2.3}&\pst{0.3}&\pst{0.0}&\pst{0.2}&\pst{0.0}&\pst{0.2}&\pst{0.0}&\pst{0.0}&\pst{0.0}&\pst{0.0}&\pst{0.0}\\\cline{1-1}

\labelname{fence} &\pst{0.2}&\pst{0.9}&\pst{2.0}&\pst{1.5}&\pst{90.4}&\pst{0.7}&\pst{0.0}&\pst{0.0}&\pst{2.7}&\pst{0.4}&\pst{0.0}&\pst{0.3}&\pst{0.0}&\pst{0.3}&\pst{0.0}&\pst{0.0}&\pst{0.0}&\pst{0.0}&\pst{0.3}\\\cline{1-1}

\labelname{pole} &\pst{0.2}&\pst{1.0}&\pst{8.6}&\pst{0.2}&\pst{0.4}&\pst{81.2}&\pst{0.6}&\pst{0.5}&\pst{5.0}&\pst{0.3}&\pst{0.8}&\pst{0.3}&\pst{0.0}&\pst{0.5}&\pst{0.0}&\pst{0.0}&\pst{0.0}&\pst{0.0}&\pst{0.3}\\\cline{1-1}

\labelname{traffic light} &\pst{0.0}&\pst{0.0}&\pst{7.0}&\pst{0.0}&\pst{0.0}&\pst{3.3}&\pst{84.3}&\pst{0.3}&\pst{4.2}&\pst{0.0}&\pst{0.6}&\pst{0.0}&\pst{0.0}&\pst{0.0}&\pst{0.0}&\pst{0.0}&\pst{0.0}&\pst{0.0}&\pst{0.0}\\\cline{1-1}

\labelname{traffic sign} &\pst{0.1}&\pst{0.0}&\pst{5.3}&\pst{0.0}&\pst{0.2}&\pst{1.2}&\pst{0.2}&\pst{89.8}&\pst{2.3}&\pst{0.0}&\pst{0.4}&\pst{0.0}&\pst{0.0}&\pst{0.2}&\pst{0.0}&\pst{0.0}&\pst{0.0}&\pst{0.0}&\pst{0.0}\\\cline{1-1}

\labelname{vegetation} &\pst{0.0}&\pst{0.0}&\pst{1.4}&\pst{0.0}&\pst{0.1}&\pst{0.3}&\pst{0.0}&\pst{0.0}&\pst{97.3}&\pst{0.2}&\pst{0.3}&\pst{0.0}&\pst{0.0}&\pst{0.0}&\pst{0.0}&\pst{0.0}&\pst{0.0}&\pst{0.0}&\pst{0.0}\\\cline{1-1}

\labelname{terrain} &\pst{1.2}&\pst{3.3}&\pst{0.3}&\pst{0.2}&\pst{0.3}&\pst{0.3}&\pst{0.0}&\pst{0.0}&\pst{3.8}&\pst{90.0}&\pst{0.0}&\pst{0.0}&\pst{0.0}&\pst{0.3}&\pst{0.0}&\pst{0.0}&\pst{0.0}&\pst{0.0}&\pst{0.1}\\\cline{1-1}

\labelname{sky} &\pst{0.0}&\pst{0.0}&\pst{0.5}&\pst{0.0}&\pst{0.0}&\pst{0.1}&\pst{0.0}&\pst{0.0}&\pst{1.3}&\pst{0.0}&\pst{98.1}&\pst{0.0}&\pst{0.0}&\pst{0.0}&\pst{0.0}&\pst{0.0}&\pst{0.0}&\pst{0.0}&\pst{0.0}\\\cline{1-1}

\labelname{person} &\pst{0.7}&\pst{0.7}&\pst{2.4}&\pst{0.0}&\pst{0.2}&\pst{0.3}&\pst{0.0}&\pst{0.0}&\pst{0.6}&\pst{0.0}&\pst{0.0}&\pst{93.6}&\pst{0.2}&\pst{0.5}&\pst{0.0}&\pst{0.0}&\pst{0.0}&\pst{0.0}&\pst{0.3}\\\cline{1-1}

\labelname{rider} &\pst{0.5}&\pst{0.3}&\pst{1.6}&\pst{0.0}&\pst{0.1}&\pst{0.2}&\pst{0.0}&\pst{0.0}&\pst{0.8}&\pst{0.0}&\pst{0.0}&\pst{3.4}&\pst{86.1}&\pst{0.8}&\pst{0.0}&\pst{0.0}&\pst{0.0}&\pst{1.5}&\pst{4.5}\\\cline{1-1}

\labelname{car} &\pst{0.5}&\pst{0.0}&\pst{0.5}&\pst{0.0}&\pst{0.0}&\pst{0.0}&\pst{0.0}&\pst{0.0}&\pst{0.3}&\pst{0.0}&\pst{0.0}&\pst{0.0}&\pst{0.0}&\pst{98.2}&\pst{0.0}&\pst{0.0}&\pst{0.0}&\pst{0.0}&\pst{0.0}\\\cline{1-1}

\labelname{truck} &\pst{0.2}&\pst{0.0}&\pst{1.1}&\pst{0.0}&\pst{0.0}&\pst{0.1}&\pst{0.0}&\pst{0.0}&\pst{0.4}&\pst{0.0}&\pst{0.0}&\pst{0.1}&\pst{0.0}&\pst{1.3}&\pst{96.3}&\pst{0.0}&\pst{0.0}&\pst{0.0}&\pst{0.0}\\\cline{1-1}

\labelname{bus} &\pst{0.2}&\pst{0.0}&\pst{0.7}&\pst{0.0}&\pst{0.0}&\pst{0.2}&\pst{0.0}&\pst{0.0}&\pst{0.5}&\pst{0.0}&\pst{0.0}&\pst{0.2}&\pst{0.0}&\pst{0.3}&\pst{0.2}&\pst{97.4}&\pst{0.0}&\pst{0.0}&\pst{0.0}\\\cline{1-1}

\labelname{train} &\pst{0.1}&\pst{0.0}&\pst{1.1}&\pst{0.0}&\pst{0.0}&\pst{0.2}&\pst{0.0}&\pst{0.0}&\pst{0.3}&\pst{0.0}&\pst{0.0}&\pst{0.1}&\pst{0.0}&\pst{0.3}&\pst{0.0}&\pst{0.0}&\pst{97.6}&\pst{0.0}&\pst{0.0}\\\cline{1-1}

\labelname{motorcycle} &\pst{0.4}&\pst{0.6}&\pst{1.7}&\pst{0.0}&\pst{0.1}&\pst{0.4}&\pst{0.0}&\pst{0.0}&\pst{0.6}&\pst{0.1}&\pst{0.0}&\pst{0.6}&\pst{1.3}&\pst{1.0}&\pst{0.0}&\pst{0.0}&\pst{0.0}&\pst{91.6}&\pst{1.5}\\\cline{1-1}

\labelname{bicycle} &\pst{0.5}&\pst{1.1}&\pst{2.1}&\pst{0.0}&\pst{0.5}&\pst{1.0}&\pst{0.0}&\pst{0.0}&\pst{0.8}&\pst{0.2}&\pst{0.0}&\pst{1.0}&\pst{1.6}&\pst{1.0}&\pst{0.0}&\pst{0.0}&\pst{0.0}&\pst{0.6}&\pst{89.5}\\\bottomrule

\end{tabular}}
\label{tab:confusion_matrix_cityscapes_train}
\end{table*}

\begin{table*}[t]
    \centering
    \newcommand{\pst}[1]{{\cellcolor{red!\fpeval{(#1 + 500*#1/(50+#1*#1))/1.3}}}\scalebox{0.6}{#1}}
    \newcommand{\ngt}[1]{{\cellcolor{blue!\fpeval{100*(#1+0.2)/1.8}}}\scalebox{0.6}{-#1}}
    \newcommand{\labelname}[1]{\scalebox{0.85}{\textit{#1}}}
    \newcommand{\zero}[0]{\scalebox{0.6}{$\approx0$}}
    \newcolumntype{"}{@{\hskip\tabcolsep\vrule width 1pt \hspace{0.1pt}|}}
    \newcolumntype{x}[0]{>{\centering\arraybackslash\hspace{0pt}}p{0.022\textwidth}}
    \caption{Confusion Matrix on the validation set of Cityscapes after training with \textit{DeepLabv3+}.}
    \vspace{-8pt}
    \resizebox{\linewidth}{!}{
    \begin{tabular}{c|x|x|x|x|x|x|x|x|x|x|x|x|x|x|x|x|x|x|x}
    \toprule
    
\multirow{5}{*}{$\%$} & \multicolumn{11}{c|}{\textit{stuff}} & \multicolumn{8}{c}{\textit{thing}} \\ \cline{2-20}
    
& \rotatebox{90}{\labelname{road}} 
& \rotatebox{90}{\labelname{sidewalk}} 
& \rotatebox{90}{\labelname{building}} 
& \rotatebox{90}{\labelname{wall}} 
& \rotatebox{90}{\labelname{fence}} 
& \rotatebox{90}{\labelname{pole}} 
& \rotatebox{90}{\labelname{tr. light}}
& \rotatebox{90}{\labelname{tr. sign}}
& \rotatebox{90}{\labelname{vegetation\ }} 
& \rotatebox{90}{\labelname{terrain}} 
& \rotatebox{90}{\labelname{sky}} 
& \rotatebox{90}{\labelname{person}} 
& \rotatebox{90}{\labelname{rider}} 
& \rotatebox{90}{\labelname{car}} 
& \rotatebox{90}{\labelname{truck}}
& \rotatebox{90}{\labelname{bus}}
& \rotatebox{90}{\labelname{train}}
& \rotatebox{90}{\labelname{motorcycle\ }}
& \rotatebox{90}{\labelname{bicycle }} \\\midrule % \hline\hline
    
\labelname{road} &\pst{99.1}&\pst{0.6}&\pst{0.0}&\pst{0.0}&\pst{0.0}&\pst{0.0}&\pst{0.0}&\pst{0.0}&\pst{0.0}&\pst{0.0}&\pst{0.0}&\pst{0.0}&\pst{0.0}&\pst{0.2}&\pst{0.0}&\pst{0.0}&\pst{0.0}&\pst{0.0}&\pst{0.0}\\\cline{1-1}

\labelname{sidewalk} &\pst{4.4}&\pst{92.7}&\pst{0.6}&\pst{0.1}&\pst{0.2}&\pst{0.4}&\pst{0.0}&\pst{0.0}&\pst{0.4}&\pst{0.5}&\pst{0.0}&\pst{0.3}&\pst{0.0}&\pst{0.3}&\pst{0.0}&\pst{0.0}&\pst{0.0}&\pst{0.0}&\pst{0.2}\\\cline{1-1}

\labelname{building} &\pst{0.0}&\pst{0.1}&\pst{96.8}&\pst{0.0}&\pst{0.0}&\pst{0.5}&\pst{0.0}&\pst{0.1}&\pst{1.7}&\pst{0.0}&\pst{0.2}&\pst{0.1}&\pst{0.0}&\pst{0.0}&\pst{0.0}&\pst{0.0}&\pst{0.0}&\pst{0.0}&\pst{0.0}\\\cline{1-1}

\labelname{wall} &\pst{0.2}&\pst{3.8}&\pst{19.2}&\pst{63.9}&\pst{4.5}&\pst{0.7}&\pst{0.0}&\pst{0.0}&\pst{5.1}&\pst{1.1}&\pst{0.0}&\pst{0.6}&\pst{0.0}&\pst{0.5}&\pst{0.0}&\pst{0.0}&\pst{0.0}&\pst{0.0}&\pst{0.3}\\\cline{1-1}

\labelname{fence} &\pst{0.5}&\pst{1.1}&\pst{13.5}&\pst{3.9}&\pst{73.1}&\pst{1.3}&\pst{0.0}&\pst{0.4}&\pst{3.4}&\pst{0.4}&\pst{0.0}&\pst{0.5}&\pst{0.0}&\pst{0.5}&\pst{0.3}&\pst{0.0}&\pst{0.0}&\pst{0.1}&\pst{0.8}\\\cline{1-1}

\labelname{pole} &\pst{0.2}&\pst{1.2}&\pst{8.9}&\pst{0.1}&\pst{0.7}&\pst{80.3}&\pst{0.6}&\pst{0.5}&\pst{5.2}&\pst{0.3}&\pst{0.5}&\pst{0.4}&\pst{0.0}&\pst{0.6}&\pst{0.0}&\pst{0.0}&\pst{0.0}&\pst{0.0}&\pst{0.4}\\\cline{1-1}

\labelname{traffic light} &\pst{0.0}&\pst{0.0}&\pst{6.2}&\pst{0.0}&\pst{0.0}&\pst{3.3}&\pst{84.1}&\pst{0.2}&\pst{5.5}&\pst{0.0}&\pst{0.6}&\pst{0.0}&\pst{0.0}&\pst{0.0}&\pst{0.0}&\pst{0.0}&\pst{0.0}&\pst{0.0}&\pst{0.0}\\\cline{1-1}

\labelname{traffic sign} &\pst{0.3}&\pst{0.1}&\pst{5.4}&\pst{0.0}&\pst{0.9}&\pst{1.6}&\pst{0.2}&\pst{87.8}&\pst{2.7}&\pst{0.0}&\pst{0.1}&\pst{0.2}&\pst{0.0}&\pst{0.2}&\pst{0.1}&\pst{0.0}&\pst{0.0}&\pst{0.0}&\pst{0.2}\\\cline{1-1}

\labelname{vegetation} &\pst{0.0}&\pst{0.1}&\pst{1.5}&\pst{0.0}&\pst{0.1}&\pst{0.4}&\pst{0.0}&\pst{0.0}&\pst{96.9}&\pst{0.4}&\pst{0.3}&\pst{0.0}&\pst{0.0}&\pst{0.0}&\pst{0.0}&\pst{0.0}&\pst{0.0}&\pst{0.0}&\pst{0.0}\\\cline{1-1}

\labelname{terrain} &\pst{1.1}&\pst{8.0}&\pst{0.5}&\pst{0.5}&\pst{0.6}&\pst{0.5}&\pst{0.0}&\pst{0.0}&\pst{13.4}&\pst{74.8}&\pst{0.0}&\pst{0.0}&\pst{0.0}&\pst{0.3}&\pst{0.0}&\pst{0.0}&\pst{0.0}&\pst{0.0}&\pst{0.2}\\\cline{1-1}

\labelname{sky} &\pst{0.0}&\pst{0.0}&\pst{0.6}&\pst{0.0}&\pst{0.0}&\pst{0.0}&\pst{0.0}&\pst{0.0}&\pst{1.0}&\pst{0.0}&\pst{98.3}&\pst{0.0}&\pst{0.0}&\pst{0.0}&\pst{0.0}&\pst{0.0}&\pst{0.0}&\pst{0.0}&\pst{0.0}\\\cline{1-1}

\labelname{person} &\pst{0.5}&\pst{0.6}&\pst{2.8}&\pst{0.0}&\pst{0.1}&\pst{0.6}&\pst{0.0}&\pst{0.0}&\pst{0.7}&\pst{0.0}&\pst{0.0}&\pst{92.2}&\pst{0.7}&\pst{0.7}&\pst{0.0}&\pst{0.0}&\pst{0.0}&\pst{0.0}&\pst{0.7}\\\cline{1-1}

\labelname{rider} &\pst{0.3}&\pst{0.3}&\pst{1.6}&\pst{0.0}&\pst{0.0}&\pst{0.2}&\pst{0.0}&\pst{0.0}&\pst{1.2}&\pst{0.0}&\pst{0.0}&\pst{13.4}&\pst{75.4}&\pst{0.9}&\pst{0.0}&\pst{0.1}&\pst{0.0}&\pst{0.8}&\pst{5.6}\\\cline{1-1}

\labelname{car} &\pst{0.4}&\pst{0.0}&\pst{0.5}&\pst{0.0}&\pst{0.0}&\pst{0.1}&\pst{0.0}&\pst{0.0}&\pst{0.3}&\pst{0.0}&\pst{0.0}&\pst{0.1}&\pst{0.0}&\pst{98.0}&\pst{0.2}&\pst{0.0}&\pst{0.0}&\pst{0.0}&\pst{0.1}\\\cline{1-1}

\labelname{truck} &\pst{0.3}&\pst{0.1}&\pst{2.7}&\pst{0.0}&\pst{0.0}&\pst{0.2}&\pst{0.0}&\pst{0.0}&\pst{0.9}&\pst{0.0}&\pst{0.3}&\pst{0.0}&\pst{0.0}&\pst{4.6}&\pst{89.1}&\pst{1.5}&\pst{0.0}&\pst{0.0}&\pst{0.0}\\\cline{1-1}

\labelname{bus} &\pst{0.3}&\pst{0.0}&\pst{1.3}&\pst{0.0}&\pst{0.4}&\pst{0.2}&\pst{0.0}&\pst{0.0}&\pst{1.3}&\pst{0.0}&\pst{0.0}&\pst{0.1}&\pst{0.0}&\pst{0.7}&\pst{0.3}&\pst{95.1}&\pst{0.0}&\pst{0.0}&\pst{0.0}\\\cline{1-1}

\labelname{train} &\pst{0.3}&\pst{0.0}&\pst{4.3}&\pst{0.0}&\pst{0.2}&\pst{0.7}&\pst{0.1}&\pst{0.2}&\pst{2.3}&\pst{0.0}&\pst{0.1}&\pst{0.0}&\pst{0.0}&\pst{0.8}&\pst{0.2}&\pst{1.0}&\pst{89.4}&\pst{0.0}&\pst{0.0}\\\cline{1-1}

\labelname{motorcycle} &\pst{0.3}&\pst{1.0}&\pst{2.1}&\pst{0.0}&\pst{1.2}&\pst{0.9}&\pst{0.0}&\pst{0.0}&\pst{0.6}&\pst{0.0}&\pst{0.0}&\pst{3.5}&\pst{2.8}&\pst{1.7}&\pst{0.0}&\pst{0.0}&\pst{0.4}&\pst{80.2}&\pst{5.3}\\\cline{1-1}

\labelname{bicycle} &\pst{0.4}&\pst{0.8}&\pst{2.2}&\pst{0.0}&\pst{0.5}&\pst{0.8}&\pst{0.0}&\pst{0.0}&\pst{1.0}&\pst{0.2}&\pst{0.0}&\pst{1.5}&\pst{2.3}&\pst{0.9}&\pst{0.0}&\pst{0.0}&\pst{0.0}&\pst{0.5}&\pst{88.8}\\\bottomrule

\end{tabular}}
\label{tab:confusion_matrix_cityscapes_val}
\end{table*}

\begin{table*}[t]
    \centering
    \newcommand{\pst}[1]{{\cellcolor{red!\fpeval{(#1 + 500*#1/(50+#1*#1))/1.3}}}\scalebox{0.6}{#1}}
    \newcommand{\ngt}[1]{{\cellcolor{blue!\fpeval{100*(#1+0.2)/1.8}}}\scalebox{0.6}{-#1}}
    \newcommand{\labelname}[1]{\scalebox{0.85}{\textit{#1}}}
    \newcommand{\zero}[0]{\scalebox{0.6}{$\approx0$}}
    \newcolumntype{"}{@{\hskip\tabcolsep\vrule width 1pt \hspace{0.1pt}|}}
    \newcolumntype{x}[0]{>{\centering\arraybackslash\hspace{0pt}}p{0.022\textwidth}}
    \caption{Confusion Matrix on the training set of KITTI-STEP after training with \textit{DeepLabv3+}.}
    \vspace{-8pt}
    \resizebox{\linewidth}{!}{
    \begin{tabular}{c|x|x|x|x|x|x|x|x|x|x|x|x|x|x|x|x|x|x|x}
    \toprule
    
\multirow{5}{*}{$\%$} & \multicolumn{11}{c|}{\textit{stuff}} & \multicolumn{8}{c}{\textit{thing}} \\ \cline{2-20}

& \rotatebox{90}{\labelname{road}} 
& \rotatebox{90}{\labelname{sidewalk}} 
& \rotatebox{90}{\labelname{building}} 
& \rotatebox{90}{\labelname{wall}} 
& \rotatebox{90}{\labelname{fence}} 
& \rotatebox{90}{\labelname{pole}} 
& \rotatebox{90}{\labelname{tr. light}}
& \rotatebox{90}{\labelname{tr. sign}}
& \rotatebox{90}{\labelname{vegetation\ }} 
& \rotatebox{90}{\labelname{terrain}} 
& \rotatebox{90}{\labelname{sky}} 
& \rotatebox{90}{\labelname{person}} 
& \rotatebox{90}{\labelname{rider}} 
& \rotatebox{90}{\labelname{car}} 
& \rotatebox{90}{\labelname{truck}}
& \rotatebox{90}{\labelname{bus}}
& \rotatebox{90}{\labelname{train}}
& \rotatebox{90}{\labelname{motorcycle\ }}
& \rotatebox{90}{\labelname{bicycle }} \\\midrule % \hline\hline
    
%%%%% LogComp
\labelname{road} &\pst{98.2}&\pst{0.8}&\pst{0.0}&\pst{0.0}&\pst{0.0}&\pst{0.1}&\pst{0.0}&\pst{0.0}&\pst{0.0}&\pst{0.5}&\pst{0.0}&\pst{0.0}&\pst{0.0}&\pst{0.2}&\pst{0.0}&\pst{0.0}&\pst{0.0}&\pst{0.0}&\pst{0.0}\\\cline{1-1}

\labelname{sidewalk} &\pst{2.2}&\pst{95.2}&\pst{0.3}&\pst{0.0}&\pst{0.0}&\pst{0.7}&\pst{0.0}&\pst{0.0}&\pst{0.4}&\pst{0.5}&\pst{0.0}&\pst{0.2}&\pst{0.0}&\pst{0.2}&\pst{0.0}&\pst{0.0}&\pst{0.0}&\pst{0.0}&\pst{0.2}\\\cline{1-1}

\labelname{building} &\pst{0.0}&\pst{0.2}&\pst{97.4}&\pst{0.0}&\pst{0.0}&\pst{0.6}&\pst{0.0}&\pst{0.0}&\pst{1.0}&\pst{0.0}&\pst{0.2}&\pst{0.2}&\pst{0.0}&\pst{0.1}&\pst{0.0}&\pst{0.0}&\pst{0.0}&\pst{0.0}&\pst{0.0}\\\cline{1-1}

\labelname{wall} &\pst{0.1}&\pst{2.0}&\pst{1.0}&\pst{92.3}&\pst{0.6}&\pst{1.0}&\pst{0.0}&\pst{0.0}&\pst{1.8}&\pst{0.0}&\pst{0.0}&\pst{0.6}&\pst{0.0}&\pst{0.0}&\pst{0.0}&\pst{0.0}&\pst{0.0}&\pst{0.0}&\pst{0.2}\\\cline{1-1}

\labelname{fence} &\pst{0.3}&\pst{0.7}&\pst{0.3}&\pst{0.4}&\pst{94.7}&\pst{0.6}&\pst{0.0}&\pst{0.0}&\pst{1.9}&\pst{0.6}&\pst{0.0}&\pst{0.0}&\pst{0.0}&\pst{0.1}&\pst{0.0}&\pst{0.0}&\pst{0.0}&\pst{0.0}&\pst{0.0}\\\cline{1-1}

\labelname{pole} &\pst{1.8}&\pst{2.4}&\pst{8.0}&\pst{0.1}&\pst{0.7}&\pst{67.6}&\pst{0.8}&\pst{0.9}&\pst{12.3}&\pst{1.2}&\pst{2.3}&\pst{0.2}&\pst{0.0}&\pst{0.9}&\pst{0.2}&\pst{0.0}&\pst{0.1}&\pst{0.0}&\pst{0.2}\\\cline{1-1}

\labelname{traffic light} &\pst{0.0}&\pst{0.0}&\pst{1.0}&\pst{0.0}&\pst{0.0}&\pst{4.2}&\pst{86.2}&\pst{0.3}&\pst{7.7}&\pst{0.0}&\pst{0.5}&\pst{0.0}&\pst{0.0}&\pst{0.0}&\pst{0.0}&\pst{0.0}&\pst{0.0}&\pst{0.0}&\pst{0.0}\\\cline{1-1}

\labelname{traffic sign} &\pst{0.9}&\pst{0.1}&\pst{1.8}&\pst{0.0}&\pst{0.0}&\pst{1.5}&\pst{0.0}&\pst{89.6}&\pst{4.9}&\pst{0.1}&\pst{0.4}&\pst{0.0}&\pst{0.0}&\pst{0.2}&\pst{0.1}&\pst{0.0}&\pst{0.0}&\pst{0.0}&\pst{0.0}\\\cline{1-1}

\labelname{vegetation} &\pst{0.0}&\pst{0.1}&\pst{0.6}&\pst{0.0}&\pst{0.1}&\pst{0.6}&\pst{0.0}&\pst{0.3}&\pst{97.4}&\pst{0.3}&\pst{0.3}&\pst{0.0}&\pst{0.0}&\pst{0.0}&\pst{0.0}&\pst{0.0}&\pst{0.0}&\pst{0.0}&\pst{0.0}\\\cline{1-1}

\labelname{terrain} &\pst{1.4}&\pst{0.7}&\pst{0.0}&\pst{0.0}&\pst{0.4}&\pst{0.6}&\pst{0.0}&\pst{0.0}&\pst{2.4}&\pst{94.3}&\pst{0.0}&\pst{0.0}&\pst{0.0}&\pst{0.0}&\pst{0.0}&\pst{0.0}&\pst{0.0}&\pst{0.0}&\pst{0.0}\\\cline{1-1}

\labelname{sky} &\pst{0.0}&\pst{0.0}&\pst{0.2}&\pst{0.0}&\pst{0.0}&\pst{0.3}&\pst{0.0}&\pst{0.0}&\pst{1.3}&\pst{0.0}&\pst{98.1}&\pst{0.0}&\pst{0.0}&\pst{0.0}&\pst{0.0}&\pst{0.0}&\pst{0.0}&\pst{0.0}&\pst{0.0}\\\cline{1-1}

\labelname{person} &\pst{1.7}&\pst{2.0}&\pst{3.2}&\pst{0.2}&\pst{0.0}&\pst{0.3}&\pst{0.0}&\pst{0.0}&\pst{0.7}&\pst{0.0}&\pst{0.0}&\pst{90.5}&\pst{0.2}&\pst{0.3}&\pst{0.0}&\pst{0.0}&\pst{0.0}&\pst{0.0}&\pst{0.7}\\\cline{1-1}

\labelname{rider} &\pst{1.2}&\pst{0.9}&\pst{1.2}&\pst{0.0}&\pst{0.0}&\pst{1.5}&\pst{0.0}&\pst{0.0}&\pst{1.7}&\pst{0.0}&\pst{0.0}&\pst{0.8}&\pst{85.3}&\pst{0.7}&\pst{0.3}&\pst{0.0}&\pst{0.0}&\pst{0.0}&\pst{6.2}\\\cline{1-1}

\labelname{car} &\pst{0.9}&\pst{0.3}&\pst{0.3}&\pst{0.0}&\pst{0.0}&\pst{0.2}&\pst{0.0}&\pst{0.0}&\pst{0.4}&\pst{0.1}&\pst{0.0}&\pst{0.0}&\pst{0.0}&\pst{97.4}&\pst{0.0}&\pst{0.2}&\pst{0.0}&\pst{0.0}&\pst{0.0}\\\cline{1-1}

\labelname{truck} &\pst{0.4}&\pst{0.3}&\pst{0.3}&\pst{0.0}&\pst{0.0}&\pst{0.3}&\pst{0.0}&\pst{0.0}&\pst{0.7}&\pst{0.0}&\pst{0.1}&\pst{0.0}&\pst{0.0}&\pst{0.4}&\pst{88.9}&\pst{8.2}&\pst{0.0}&\pst{0.0}&\pst{0.2}\\\cline{1-1}

\labelname{bus} &\pst{0.4}&\pst{0.0}&\pst{0.2}&\pst{0.0}&\pst{0.0}&\pst{0.0}&\pst{0.0}&\pst{0.0}&\pst{0.4}&\pst{0.0}&\pst{0.6}&\pst{0.0}&\pst{0.0}&\pst{0.5}&\pst{1.2}&\pst{95.9}&\pst{0.1}&\pst{0.0}&\pst{0.5}\\\cline{1-1}

\labelname{train} &\pst{0.0}&\pst{0.0}&\pst{1.3}&\pst{0.0}&\pst{1.2}&\pst{0.8}&\pst{0.1}&\pst{0.1}&\pst{0.1}&\pst{0.0}&\pst{0.0}&\pst{0.1}&\pst{0.0}&\pst{0.0}&\pst{0.0}&\pst{0.0}&\pst{96.1}&\pst{0.0}&\pst{0.0}\\\cline{1-1}

\labelname{motorcycle} &\pst{0.5}&\pst{1.5}&\pst{1.9}&\pst{0.0}&\pst{0.0}&\pst{0.8}&\pst{0.0}&\pst{0.0}&\pst{1.9}&\pst{0.0}&\pst{0.0}&\pst{0.2}&\pst{0.0}&\pst{0.6}&\pst{0.0}&\pst{0.0}&\pst{0.0}&\pst{91.7}&\pst{0.7}\\\cline{1-1}

\labelname{bicycle} &\pst{0.7}&\pst{3.1}&\pst{1.3}&\pst{0.1}&\pst{0.0}&\pst{0.6}&\pst{0.0}&\pst{0.0}&\pst{0.7}&\pst{0.1}&\pst{0.0}&\pst{0.6}&\pst{0.8}&\pst{0.0}&\pst{0.0}&\pst{0.1}&\pst{0.0}&\pst{0.0}&\pst{91.6}\\\bottomrule

\end{tabular}}
\label{tab:confusion_matrix_kitti_train}
\end{table*}

\begin{table*}[t]
    \centering
    \newcommand{\pst}[1]{{\cellcolor{red!\fpeval{(#1 + 500*#1/(50+#1*#1))/1.3}}}\scalebox{0.6}{#1}}
    \newcommand{\ngt}[1]{{\cellcolor{blue!\fpeval{100*(#1+0.2)/1.8}}}\scalebox{0.6}{-#1}}
    \newcommand{\labelname}[1]{\scalebox{0.85}{\textit{#1}}}
    \newcommand{\zero}[0]{\scalebox{0.6}{$\approx0$}}
    \newcolumntype{"}{@{\hskip\tabcolsep\vrule width 1pt \hspace{0.1pt}|}}
    \newcolumntype{x}[0]{>{\centering\arraybackslash\hspace{0pt}}p{0.022\textwidth}}
    \caption{Confusion Matrix on the validation set of KITTI-STEP after training with \textit{DeepLabv3+}.}
    \vspace{-8pt}
    \resizebox{\linewidth}{!}{
    \begin{tabular}{c|x|x|x|x|x|x|x|x|x|x|x|x|x|x|x|x|x|x|x}
    \toprule
    
\multirow{5}{*}{$\%$} & \multicolumn{11}{c|}{\textit{stuff}} & \multicolumn{8}{c}{\textit{thing}} \\ \cline{2-20}
    
& \rotatebox{90}{\labelname{road}} 
& \rotatebox{90}{\labelname{sidewalk}} 
& \rotatebox{90}{\labelname{building}} 
& \rotatebox{90}{\labelname{wall}} 
& \rotatebox{90}{\labelname{fence}} 
& \rotatebox{90}{\labelname{pole}} 
& \rotatebox{90}{\labelname{tr. light}}
& \rotatebox{90}{\labelname{tr. sign}}
& \rotatebox{90}{\labelname{vegetation\ }} 
& \rotatebox{90}{\labelname{terrain}} 
& \rotatebox{90}{\labelname{sky}} 
& \rotatebox{90}{\labelname{person}} 
& \rotatebox{90}{\labelname{rider}} 
& \rotatebox{90}{\labelname{car}} 
& \rotatebox{90}{\labelname{truck}}
& \rotatebox{90}{\labelname{bus}}
& \rotatebox{90}{\labelname{train}}
& \rotatebox{90}{\labelname{motorcycle\ }}
& \rotatebox{90}{\labelname{bicycle }} \\\midrule % \hline\hline
    
\labelname{road} &\pst{94.7}&\pst{3.8}&\pst{0.0}&\pst{0.0}&\pst{0.0}&\pst{0.2}&\pst{0.0}&\pst{0.0}&\pst{0.0}&\pst{0.8}&\pst{0.0}&\pst{0.0}&\pst{0.0}&\pst{0.2}&\pst{0.0}&\pst{0.0}&\pst{0.0}&\pst{0.0}&\pst{0.0}\\\cline{1-1}

\labelname{sidewalk} &\pst{16.4}&\pst{75.5}&\pst{0.5}&\pst{0.0}&\pst{0.2}&\pst{0.8}&\pst{0.0}&\pst{0.0}&\pst{1.0}&\pst{5.1}&\pst{0.0}&\pst{0.2}&\pst{0.0}&\pst{0.3}&\pst{0.0}&\pst{0.0}&\pst{0.0}&\pst{0.0}&\pst{0.0}\\\cline{1-1}

\labelname{building} &\pst{0.0}&\pst{0.7}&\pst{90.4}&\pst{0.6}&\pst{2.1}&\pst{0.7}&\pst{0.0}&\pst{0.0}&\pst{3.4}&\pst{0.0}&\pst{1.3}&\pst{0.2}&\pst{0.0}&\pst{0.2}&\pst{0.2}&\pst{0.0}&\pst{0.0}&\pst{0.0}&\pst{0.1}\\\cline{1-1}

\labelname{wall} &\pst{0.2}&\pst{5.1}&\pst{23.7}&\pst{35.7}&\pst{17.3}&\pst{1.9}&\pst{0.0}&\pst{0.0}&\pst{14.5}&\pst{0.6}&\pst{0.0}&\pst{0.0}&\pst{0.0}&\pst{0.8}&\pst{0.0}&\pst{0.0}&\pst{0.0}&\pst{0.0}&\pst{0.0}\\\cline{1-1}

\labelname{fence} &\pst{1.7}&\pst{9.5}&\pst{7.9}&\pst{0.6}&\pst{36.5}&\pst{2.3}&\pst{0.0}&\pst{0.1}&\pst{26.6}&\pst{12.8}&\pst{0.0}&\pst{0.2}&\pst{0.0}&\pst{1.3}&\pst{0.1}&\pst{0.0}&\pst{0.0}&\pst{0.0}&\pst{0.2}\\\cline{1-1}

\labelname{pole} &\pst{2.1}&\pst{2.6}&\pst{6.0}&\pst{0.0}&\pst{0.3}&\pst{58.0}&\pst{0.7}&\pst{0.7}&\pst{20.4}&\pst{2.8}&\pst{5.2}&\pst{0.3}&\pst{0.0}&\pst{0.4}&\pst{0.0}&\pst{0.0}&\pst{0.0}&\pst{0.0}&\pst{0.3}\\\cline{1-1}

\labelname{traffic light} &\pst{0.0}&\pst{0.0}&\pst{2.9}&\pst{0.0}&\pst{0.0}&\pst{5.4}&\pst{71.7}&\pst{3.2}&\pst{14.6}&\pst{0.0}&\pst{2.1}&\pst{0.0}&\pst{0.0}&\pst{0.0}&\pst{0.0}&\pst{0.0}&\pst{0.0}&\pst{0.0}&\pst{0.0}\\\cline{1-1}

\labelname{traffic sign} &\pst{2.4}&\pst{0.2}&\pst{7.0}&\pst{0.0}&\pst{0.0}&\pst{6.2}&\pst{1.6}&\pst{63.8}&\pst{12.5}&\pst{1.2}&\pst{3.2}&\pst{0.8}&\pst{0.0}&\pst{0.7}&\pst{0.1}&\pst{0.0}&\pst{0.0}&\pst{0.0}&\pst{0.0}\\\cline{1-1}

\labelname{vegetation} &\pst{0.0}&\pst{0.0}&\pst{0.6}&\pst{0.0}&\pst{0.1}&\pst{0.7}&\pst{0.0}&\pst{0.0}&\pst{96.7}&\pst{1.0}&\pst{0.5}&\pst{0.0}&\pst{0.0}&\pst{0.0}&\pst{0.0}&\pst{0.0}&\pst{0.0}&\pst{0.0}&\pst{0.0}\\\cline{1-1}

\labelname{terrain} &\pst{2.1}&\pst{1.7}&\pst{0.0}&\pst{0.0}&\pst{0.4}&\pst{1.1}&\pst{0.0}&\pst{0.0}&\pst{5.1}&\pst{89.4}&\pst{0.0}&\pst{0.0}&\pst{0.0}&\pst{0.1}&\pst{0.0}&\pst{0.0}&\pst{0.0}&\pst{0.0}&\pst{0.0}\\\cline{1-1}

\labelname{sky} &\pst{0.0}&\pst{0.0}&\pst{0.7}&\pst{0.0}&\pst{0.0}&\pst{0.7}&\pst{0.0}&\pst{0.0}&\pst{2.2}&\pst{0.0}&\pst{96.3}&\pst{0.0}&\pst{0.0}&\pst{0.0}&\pst{0.0}&\pst{0.0}&\pst{0.0}&\pst{0.0}&\pst{0.0}\\\cline{1-1}

\labelname{person} &\pst{2.3}&\pst{3.8}&\pst{7.2}&\pst{0.0}&\pst{0.0}&\pst{0.8}&\pst{0.0}&\pst{0.0}&\pst{2.3}&\pst{0.4}&\pst{0.0}&\pst{78.9}&\pst{1.1}&\pst{0.8}&\pst{0.0}&\pst{0.0}&\pst{0.0}&\pst{0.0}&\pst{2.2}\\\cline{1-1}

\labelname{rider} &\pst{0.9}&\pst{2.0}&\pst{4.8}&\pst{0.0}&\pst{0.0}&\pst{0.9}&\pst{0.0}&\pst{0.0}&\pst{3.4}&\pst{0.0}&\pst{0.0}&\pst{27.7}&\pst{47.2}&\pst{1.4}&\pst{0.0}&\pst{0.0}&\pst{0.0}&\pst{0.0}&\pst{11.6}\\\cline{1-1}

\labelname{car} &\pst{1.1}&\pst{0.4}&\pst{1.0}&\pst{0.0}&\pst{0.0}&\pst{0.2}&\pst{0.0}&\pst{0.0}&\pst{0.7}&\pst{0.1}&\pst{0.0}&\pst{0.0}&\pst{0.0}&\pst{94.7}&\pst{1.0}&\pst{0.3}&\pst{0.0}&\pst{0.0}&\pst{0.0}\\\cline{1-1}

\labelname{truck} &\pst{1.0}&\pst{0.6}&\pst{12.3}&\pst{0.0}&\pst{0.0}&\pst{0.7}&\pst{0.0}&\pst{2.7}&\pst{9.6}&\pst{0.3}&\pst{0.2}&\pst{0.0}&\pst{0.0}&\pst{37.8}&\pst{33.3}&\pst{1.2}&\pst{0.0}&\pst{0.0}&\pst{0.0}\\\cline{1-1}

\labelname{bus} &\pst{1.4}&\pst{0.1}&\pst{20.2}&\pst{0.0}&\pst{0.0}&\pst{1.5}&\pst{0.2}&\pst{0.3}&\pst{5.5}&\pst{0.0}&\pst{0.0}&\pst{0.3}&\pst{0.0}&\pst{20.1}&\pst{8.1}&\pst{29.8}&\pst{12.4}&\pst{0.0}&\pst{0.0}\\\cline{1-1}

\labelname{train} &\pst{0.2}&\pst{1.9}&\pst{61.8}&\pst{0.0}&\pst{0.4}&\pst{0.8}&\pst{0.0}&\pst{1.5}&\pst{0.4}&\pst{0.0}&\pst{0.0}&\pst{0.0}&\pst{0.0}&\pst{0.5}&\pst{24.2}&\pst{7.1}&\pst{1.1}&\pst{0.0}&\pst{0.0}\\\cline{1-1}

\labelname{motorcycle} &\pst{0.0}&\pst{6.1}&\pst{3.2}&\pst{0.0}&\pst{0.0}&\pst{1.9}&\pst{0.0}&\pst{0.0}&\pst{0.0}&\pst{0.0}&\pst{0.0}&\pst{5.9}&\pst{2.4}&\pst{0.2}&\pst{0.0}&\pst{0.0}&\pst{0.0}&\pst{72.4}&\pst{7.8}\\\cline{1-1}

\labelname{bicycle} &\pst{1.3}&\pst{4.6}&\pst{4.6}&\pst{0.0}&\pst{0.0}&\pst{0.7}&\pst{0.0}&\pst{0.0}&\pst{2.8}&\pst{0.3}&\pst{0.0}&\pst{2.3}&\pst{1.1}&\pst{1.3}&\pst{0.2}&\pst{0.0}&\pst{0.0}&\pst{0.1}&\pst{80.6}\\\bottomrule

\end{tabular}}
\label{tab:confusion_matrix_kitti_val}
\end{table*}
   % <-- DO NOT FORGET TO REMOVE THIS FOR SUBMISSION!!!

\clearpage

%%%%%%%%% REFERENCES
{\small
%\balance
\bibliographystyle{ieee_fullname}
\bibliography{egbib}
}